\documentclass{article}

\usepackage{microtype}
\usepackage{graphicx}
\usepackage{subfigure}
\usepackage{booktabs} %
\usepackage[dvipsnames]{xcolor}

\usepackage{hyperref}

\usepackage[accepted]{icml2024}

\usepackage{amsmath}
\usepackage{amssymb}
\usepackage{mathtools}
\usepackage{amsthm}
\usepackage{xspace}
\usepackage{array}
\usepackage{arydshln}

\usepackage[capitalize,noabbrev]{cleveref}

\usepackage{tcolorbox}

\newlength\newl

\newcommand{\alpaca}{Alpaca\xspace}
\newcommand{\alpagasus}{AlpaGasus\xspace}
\newcommand{\lima}{LIMA\xspace}
\newcommand{\longest}{longest\xspace}
\newcommand{\llama}{Llama\xspace}
\newcommand{\palm}{PaLM-2\xspace}
\newcommand{\evol}{Evol-Instruct\xspace}
\newcommand{\neftune}{NEFTune\xspace}
\newcommand{\refined}{Refined\xspace}
\newcommand{\rowindent}{\hspace{2mm}\xspace}

\newcolumntype{C}[1]{>{\centering\arraybackslash}p{#1}}
\newcolumntype{L}[1]{>{\raggedright\arraybackslash}p{#1}}
\newcolumntype{R}[1]{>{\raggedleft\arraybackslash}p{#1}}

\theoremstyle{plain}

\theoremstyle{definition}

\theoremstyle{remark}

\usepackage[textsize=tiny]{todonotes}

\icmltitlerunning{A Simple but Tough-to-Beat Baseline for Instruction Fine-Tuning}

\begin{document}

\twocolumn[
\icmltitle{\textit{Long} Is More for Alignment:\\A Simple but Tough-to-Beat Baseline for Instruction Fine-Tuning}

\icmlsetsymbol{equal}{*}

\begin{icmlauthorlist}
\vspace{-1mm}
\icmlauthor{Hao Zhao}{epfl}
\icmlauthor{Maksym Andriushchenko}{epfl}
\icmlauthor{Francesco Croce}{epfl}
\icmlauthor{Nicolas Flammarion}{epfl}
\vspace{-1mm}
\end{icmlauthorlist}

\icmlaffiliation{epfl}{EPFL, Switzerland}
\icmlcorrespondingauthor{Hao Zhao}{hao.zhao@epfl.ch}

\icmlkeywords{Machine Learning, ICML}

\vskip 0.3in
]

\printAffiliationsAndNotice{}  %

\begin{abstract}
There is a consensus that instruction fine-tuning of LLMs requires \textit{high-quality} data, but what are they?
LIMA (NeurIPS 2023) and AlpaGasus (ICLR 2024) are state-of-the-art methods for selecting such high-quality examples, either via manual curation or using GPT-3.5-Turbo as a quality scorer.
We show that the extremely simple baseline of selecting the \textit{1,000 instructions with longest responses}---that intuitively contain more learnable information and are harder to overfit---from standard datasets can consistently outperform these sophisticated methods according to GPT-4 and \palm as judges, while remaining competitive on the Open LLM benchmarks that test factual knowledge. 
We demonstrate this for several LLMs (Llama-2-7B, Llama-2-13B, Mistral-7B-v0.1) and datasets (Alpaca-52k, Evol-Instruct-70k). 
In addition, a lightweight refinement of such long instructions can further improve the abilities of the fine-tuned LLMs, and allows us to obtain competitive results on MT-Bench and the 2nd highest-ranked Llama-2-7B-based model on AlpacaEval 2.0, while training on only 1,000 examples and no extra preference data. 
We also conduct a thorough analysis of our models to ensure that their enhanced performance is \textit{not} simply due to GPT-4's preference for longer responses. %
Overall, our findings suggest that fine-tuning on the longest responses should be the default baseline for any work on instruction fine-tuning. We provide our code in \href{https://github.com/tml-epfl/long-is-more-for-alignment}{this GitHub repository}.
\end{abstract}

\section{Introduction}

\begin{figure*}[!t]
    \centering
    \subfigure[\small Head-to-head comparisons (in \%) with two different LLM judges]{
        \includegraphics[width=0.58\textwidth]{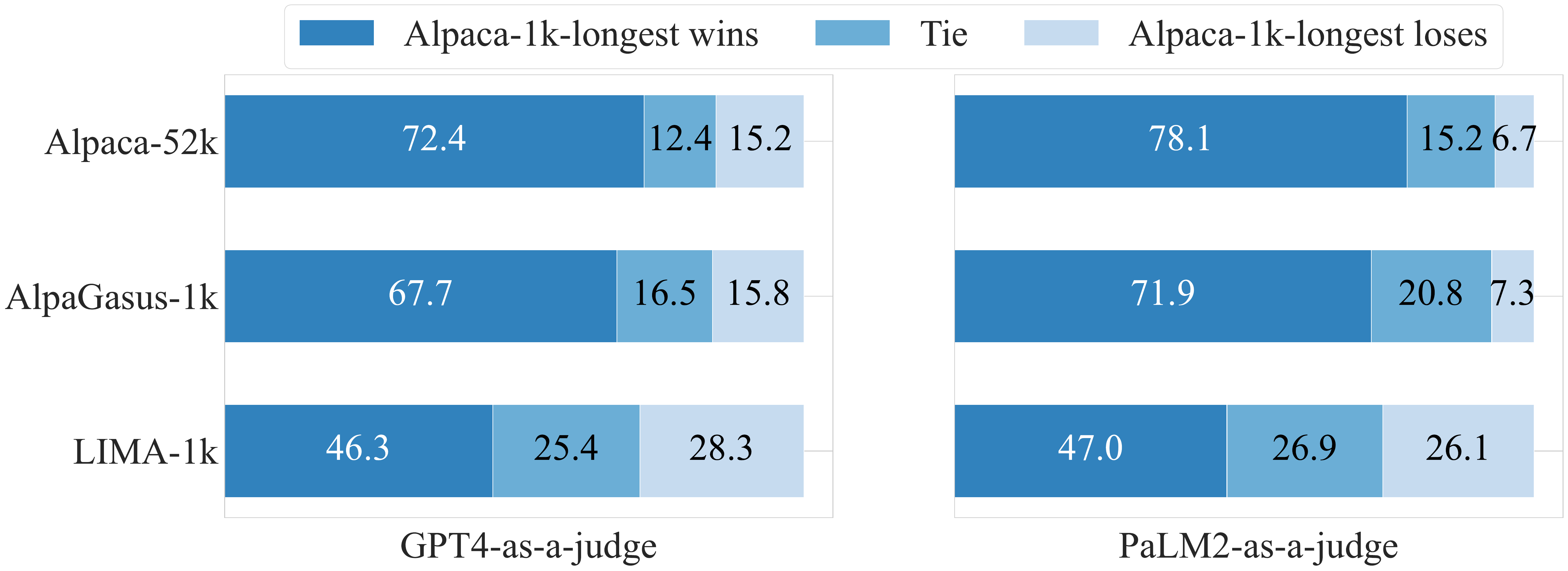}
        \label{fig:llama_2_alpaca_win_rate_multiple_judges}
    }
    \subfigure[\small Average number of tokens in responses]{
        \includegraphics[width=0.38\textwidth]{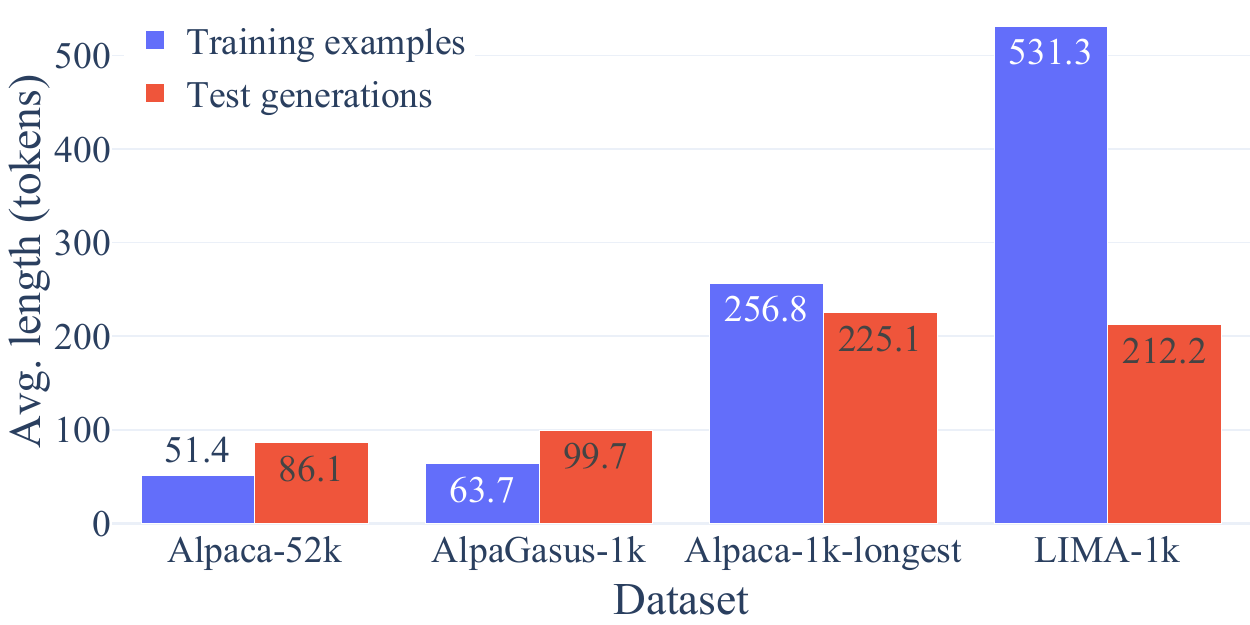}
        \label{fig:num_tokens_dist}
    }
    \vspace{-3mm}
    \caption{\textbf{Selecting the longest responses leads to a strong IFT dataset.} We fine-tune LLaMA-2-7B models on Alpaca-52k~\citep{alpaca}, AlpaGasus-1k~\citep{chen2023alpagasus}, LIMA-1k~\citep{zhou2023lima} and our \alpaca-1k-\longest datasets.
    \textbf{(a)} Alpaca-1k-longest beats three baselines in instruction-following performance according to both GPT-4 and \palm as judges.
    \textbf{(b)} \alpaca-1k-\longest leads to an average response length at test time higher than \alpaca-52k and \alpagasus-1k, but similar to \lima-1k: then its higher win rate cannot be solely attributed to the model having learnt to generate long responses. %
    }
    \label{fig:teaser}
    
\end{figure*}

Pre-trained large language models (LLMs) need to undergo an alignment phase~\citep{askell2021general,bai2022training,ouyang2022training,wang2022self,alpaca} to make them suitable for downstream tasks like user interaction or question answering.
While the details may vary, alignment often relies on supervised fine-tuning (SFT) on a dataset of instruction-response pairs to improve conversational ability, followed by reinforcement learning from either human (RLHF)~\citep{ouyang2022training} or automated (RLAIF)~\citep{bai2022constitutional,lee2023rlaif} feedback to promote the preferred style and content of replies.
It is an active research direction to study whether it is possible to achieve satisfactory results while relying only on SFT, which would avoid the (potentially expensive) process of collecting preference data.
\citet{alpaca} created \alpaca, an open source dataset of 52k instruction-response pairs, and fine-tuned on it a \llama-2-7B model to match the performance of the closed-source text-davinci-003 model.
Then, \citet{chen2023alpagasus} introduced \alpagasus, consisting of the 9k examples of \alpaca which are judged of highest quality by GPT-3.5-Turbo, to further improve the instruction-following abilities of the fine-tuned models.
The intuition that instruction fine-tuning (IFT) might benefit from fewer demonstrations but of higher quality has been further pursued by \citet{zhou2023lima} which manually curated \lima, a dataset of 1k examples, which outperforms \alpagasus.
While the quality of the instructions seems to play a major role for IFT, it remains unclear which are the distinguishing features of high quality demonstrations.
\looseness=-1

In this work, we revisit the significant efforts in constructing instruction-tuning datasets from prior work. 
Inspired by the fact \lima contains much longer examples than \alpaca and the observation of recent works \citep{singhal2023long, yuan2024self} that RLHF and direct preference optimization (DPO) \citep{rafailov2023direct} seem to mostly make the outputs longer, %
we test selecting longest responses as a simple and inexpensive heuristic to curate a small (only 1k examples) and high-quality IFT dataset from a larger one. 
Surprisingly, fine-tuning a \llama-2-7B \citep{touvron2023llama2} base model on the 1k longest elements of \alpaca outperforms both \alpagasus and \lima in one-to-one comparison with different LLMs as judges and on the AlpacaEval 2.0 benchmark (see Fig.~\ref{fig:teaser}).
Moreover, simply improving the quality and the style of the response in \alpaca-1k-\longest with GPT-3.5-Turbo, in combination with \neftune noise augmentation \citep{jain2023neftune}, allows us to obtain \textit{the 2nd highest-ranked Llama-2-7B-based model} on AlpacaEval 2.0.
In this case, our simple method yields models which surpass LLMs with the same base model but fine-tuned with %
orders of magnitude more instructions as well as millions of preference data points.

Next we analyze several aspects of our models to understand the unexpected effectiveness of our approach. First, via several ablation studies, we show that our models do not just exploit the bias to favor longer responses of GPT-4 \citep{Achiam2023GPT4TR} or \palm \citep{anil2023palm}, but provide higher quality replies.
Then, since \citet{jha2023limit, gudibande2023false} suggest that optimizing performance of instruction-following tasks might 
be disconnected from factual knowledge, we additionally test our models on then Open LLM benchmarks. 
On these datasets assessing reasoning and factuality, our models perform similarly or better than the baselines fine-tuned on \alpagasus and \lima from the same base model, i.e. with the same factual knowledge coming from pre-training.
Finally, we confirm our findings with extensive experiments using multiple IFT datasets (\alpaca, \evol) and architectures (\llama-2-7B, \llama-2-13B, Mistral-7B-v0.1 \citep{jiang2023mistral}), and including head-to-head evaluation and on established benchmarks (AlpacaEval 2.0, Open LLM), %
to show the generality of our approach.

In summary, we uncover the surprising effectiveness of fine-tuning only on the longest 1,000 instructions of large datasets to obtain aligned models. 
Moreover, we show that such small datasets, potentially refined via an inexpensive automatic process, constitute a strong and tough-to-beat baseline for any method for instruction fine-tuning.

\begin{figure*}[t]
    \centering
    \subfigure[\scriptsize Alpaca-1k-longest vs. LIMA-1k]{
        \includegraphics[width=0.31\textwidth]{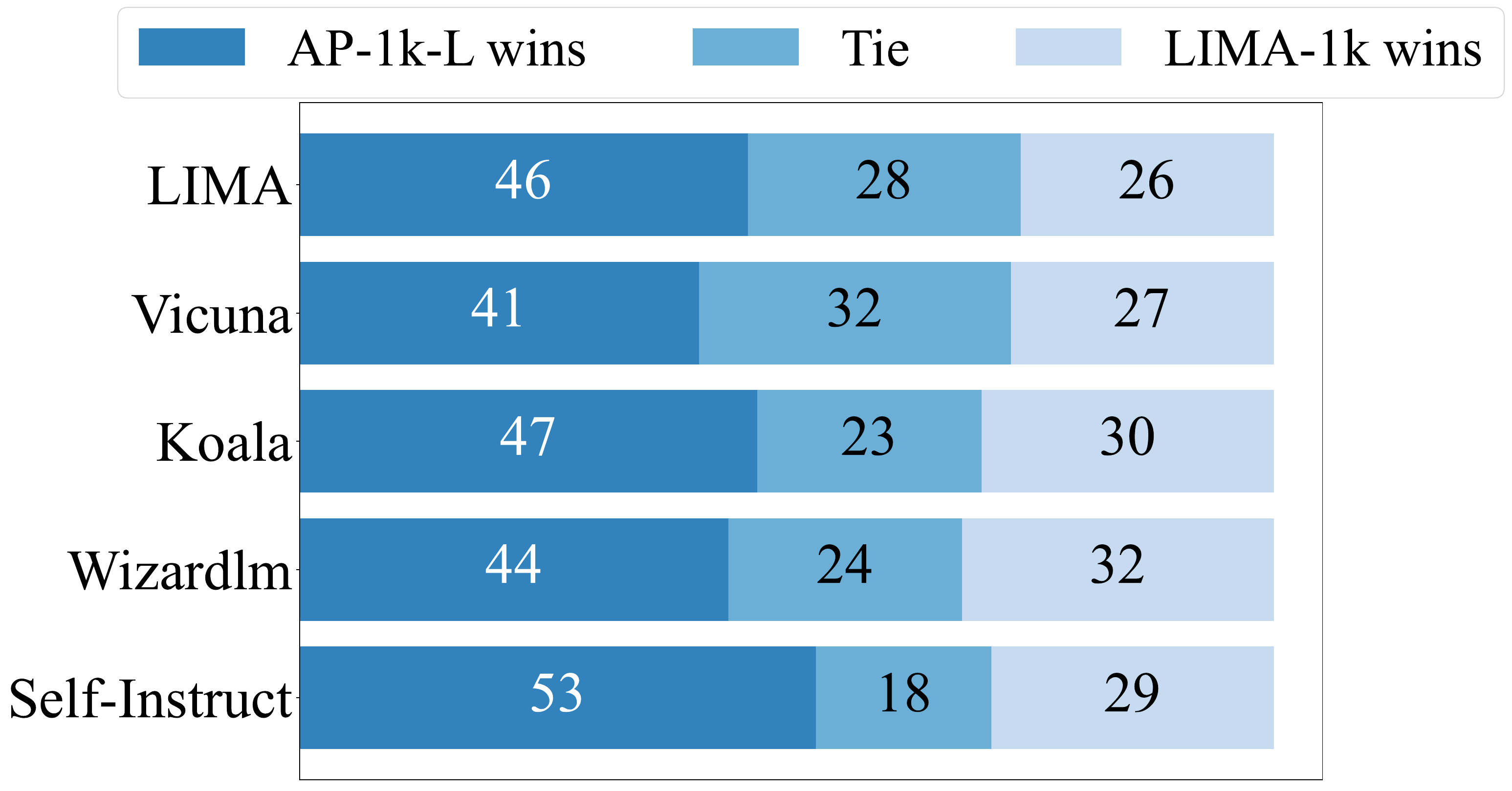}
        \label{fig:llama_2_alpaca_1k_longest_vs_lima_gpt_4}
    }
    \hfill
    \subfigure[\scriptsize Alpaca-1k-longest vs. AlpaGasus-1k]{
        \includegraphics[width=0.31\textwidth]{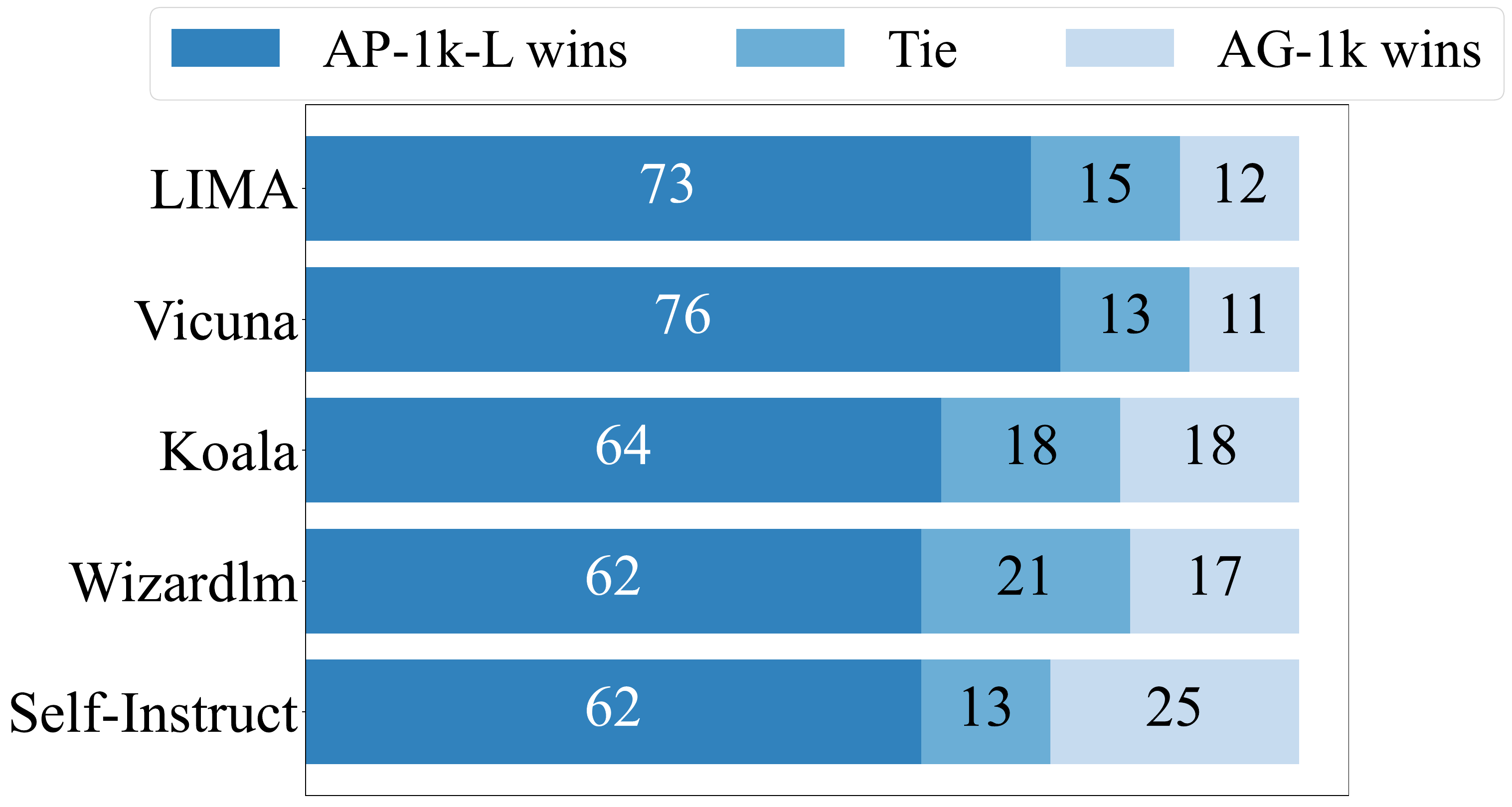}
        \label{fig:llama_2_alpaca_1k_longest_vs_alpagasus_1k_gpt_4}
    }
    \hfill
    \subfigure[\scriptsize Alpaca-1k-longest vs. Alpaca-52k]{
        \includegraphics[width=0.31\textwidth]{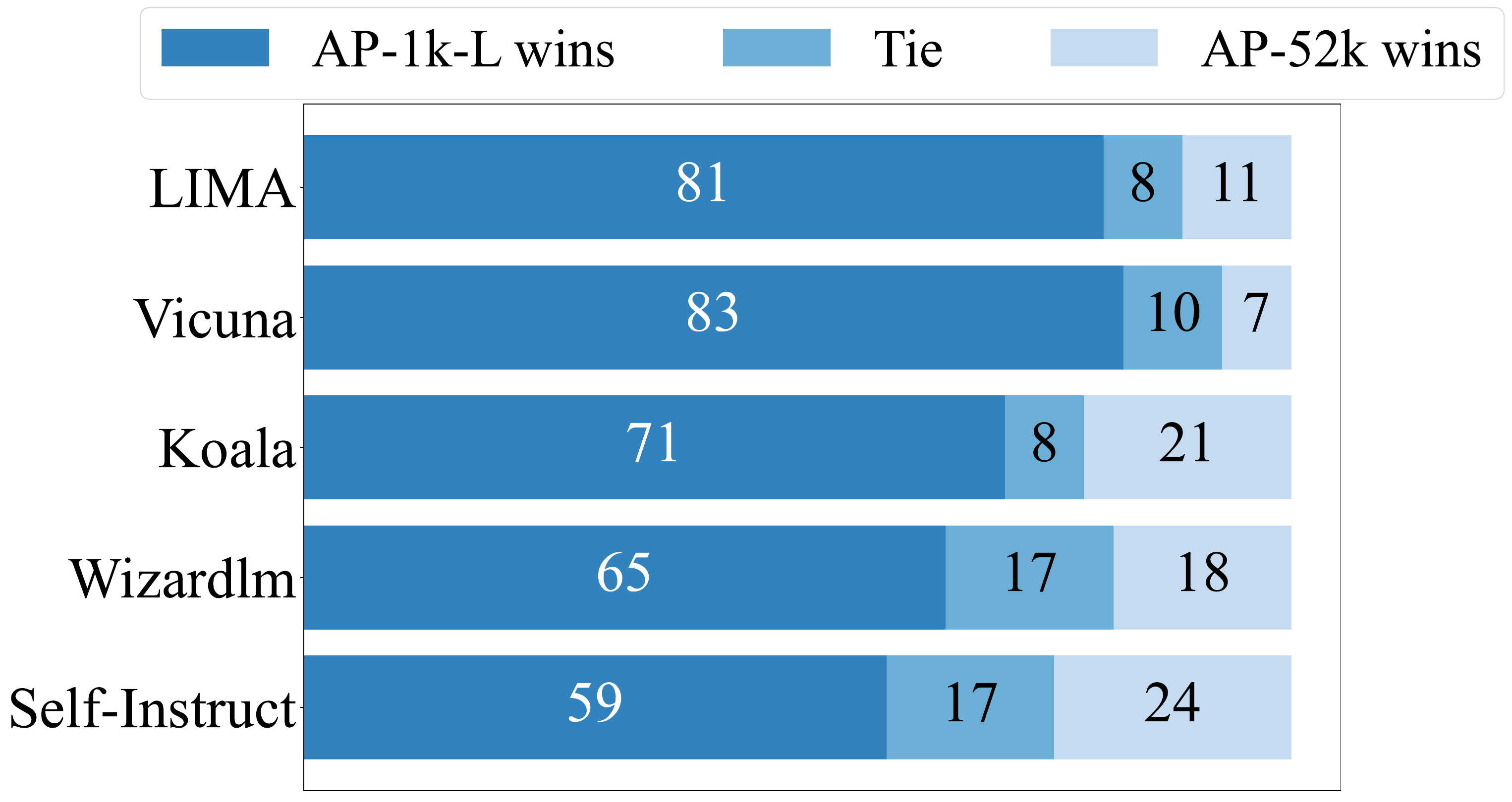}
        \label{fig:llama_2_alpaca_1k_longest_vs_alpaca_52k_gpt_4}
    } \\
    \subfigure[\scriptsize Evol-Instruct-1k-longest vs. LIMA-1k]{
        \includegraphics[width=0.31\textwidth]{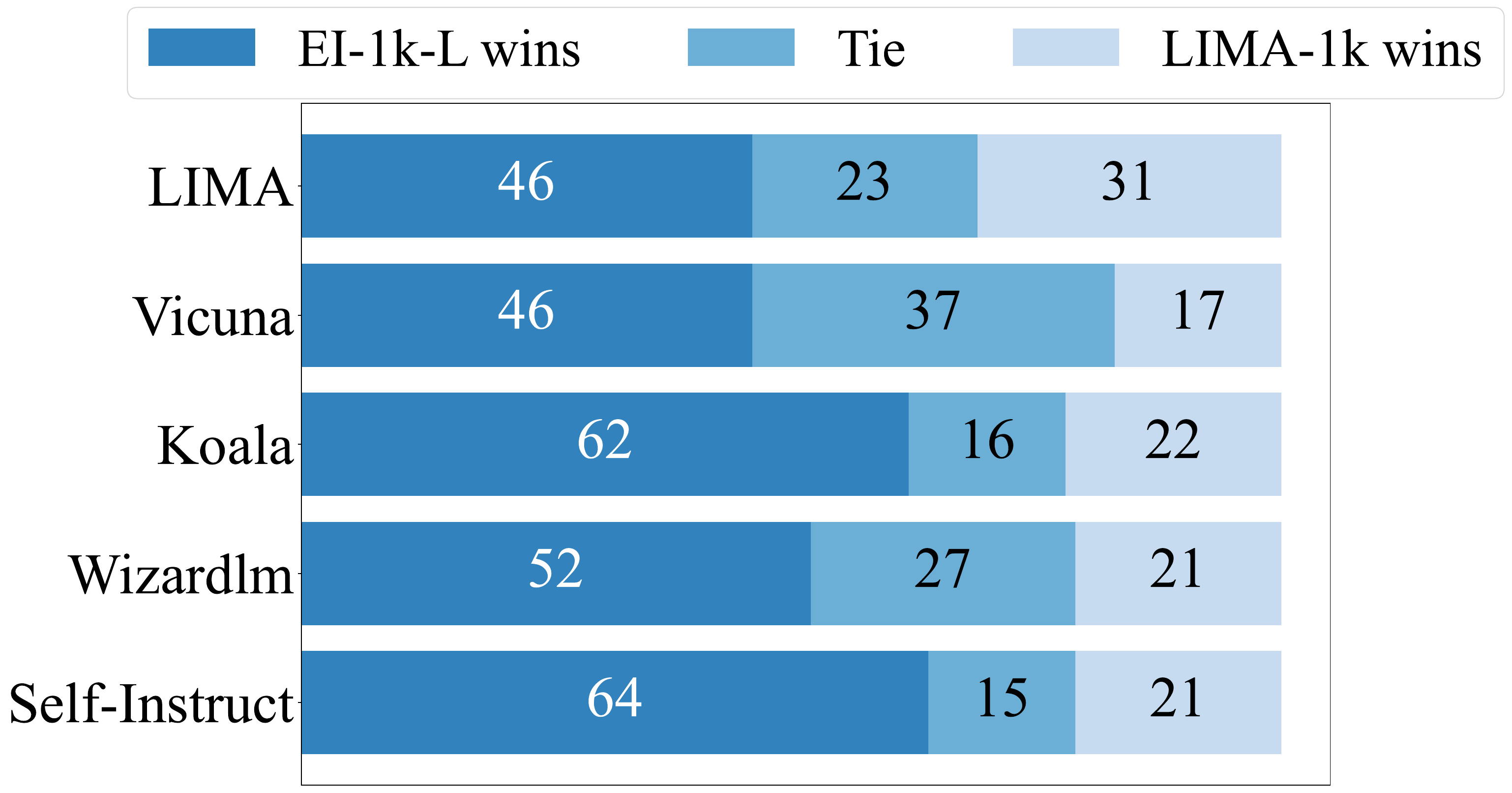}
        \label{fig:llama_2_evol_instruct_1k_longest_vs_lima_gpt_4}
    }
    \hfill
    \subfigure[\scriptsize Evol-Instr.-1k-longest vs. EvolInstr.-AlpaGasus-1k]{
        \includegraphics[width=0.31\textwidth]{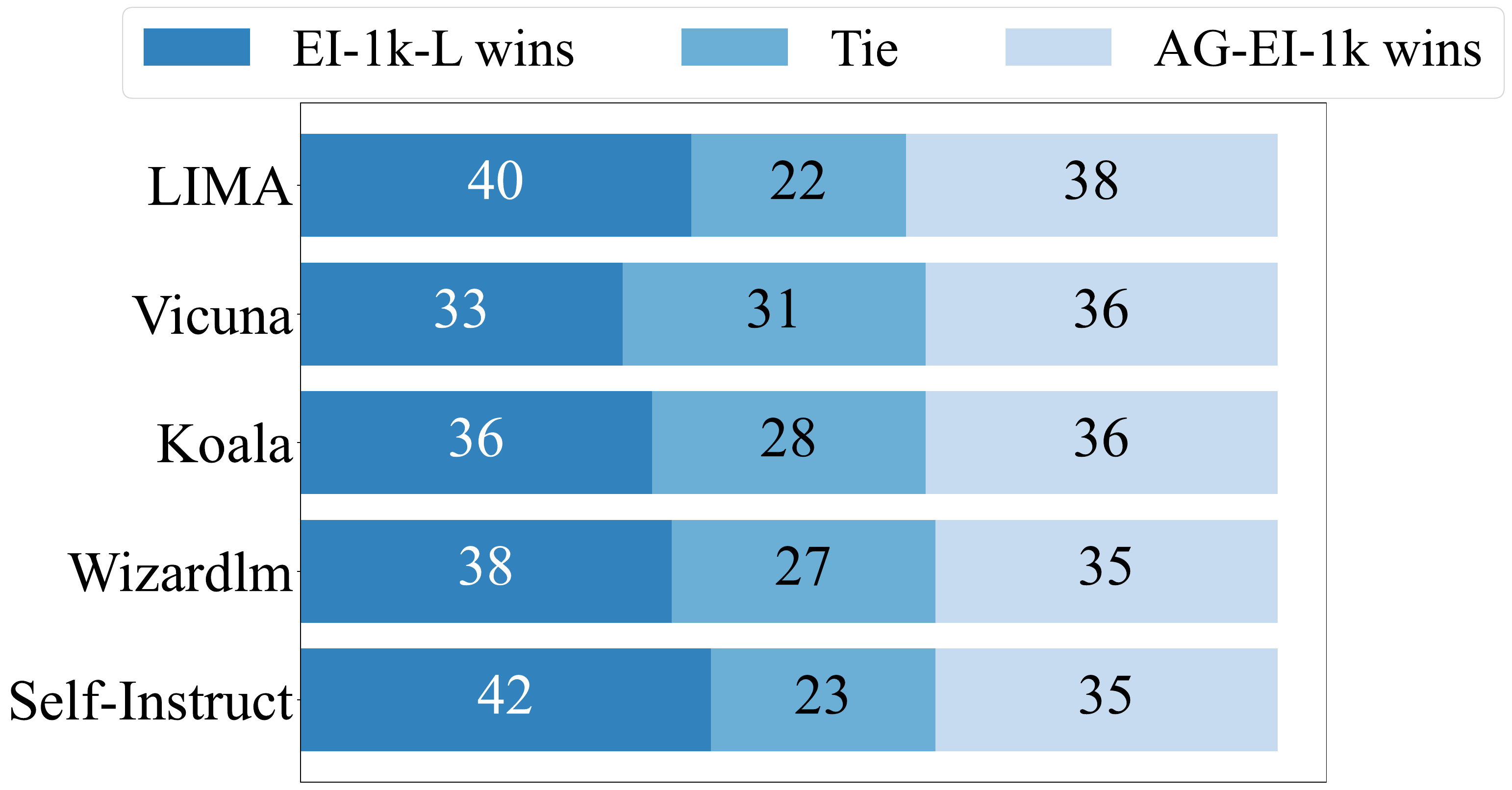}
        \label{fig:llama_2_evol_instruct_1k_longest_vs_alpagasus_evol_instruct_1k_gpt_4}
    }
    \hfill
    \subfigure[\scriptsize Evol-Instruct-1k-longest vs. Evol-Instruct-70k]{
        \includegraphics[width=0.31\textwidth]{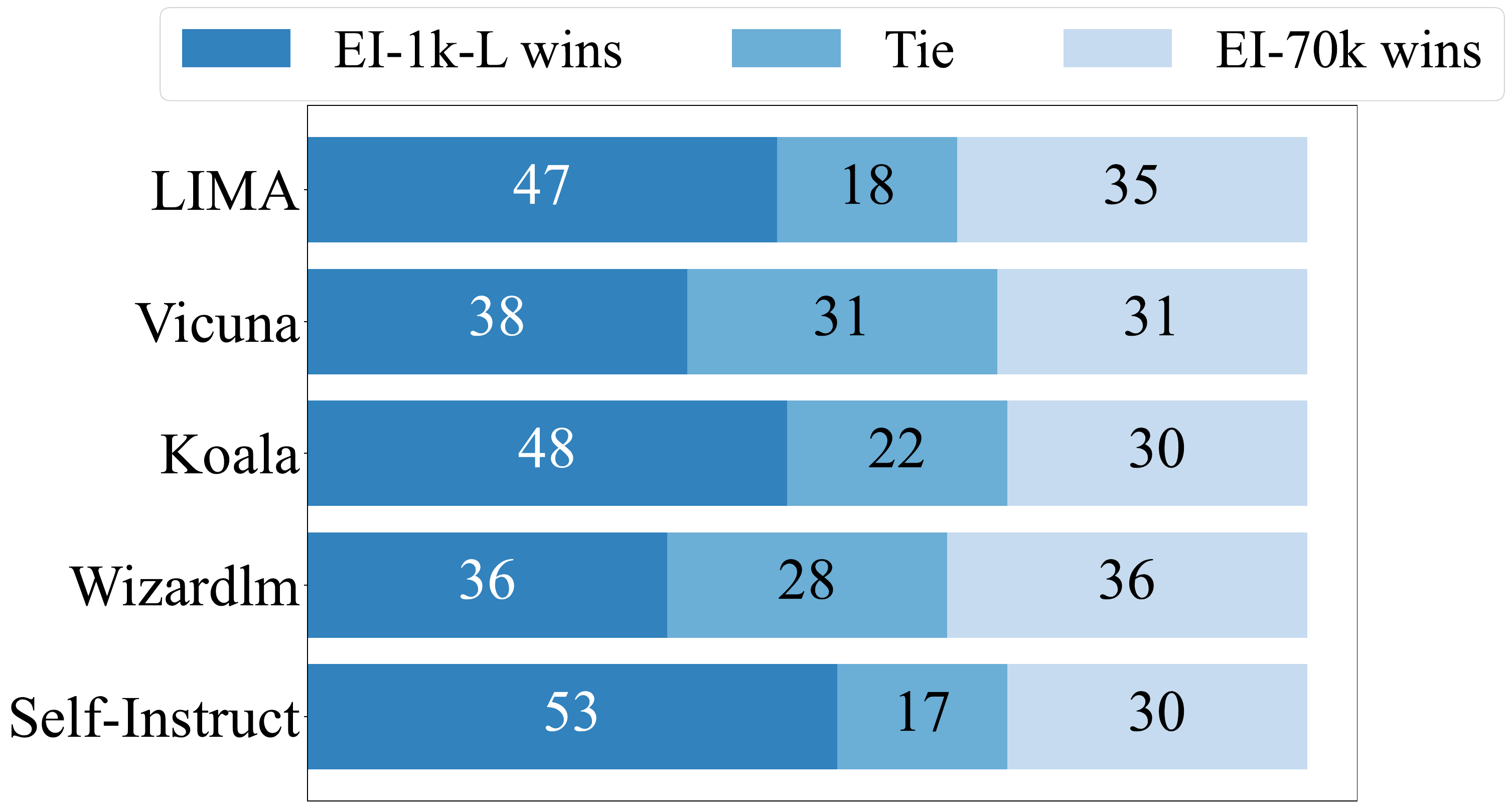}
        \label{fig:llama_2_evol_instruct_1k_longest_vs_evol_instruct_70k_gpt_4}
    }
    \vspace{-3mm}
    \caption{\textbf{Detailed preference evaluation (in \%).}
    For each pair of LLMs we report the win rate on 5 datasets (LIMA, Vicuna, Koala, WizardLM, Self-Instruct) according to GPT-4-as-a-judge. 
    \textbf{Top:} we compare fine-tuning on \alpaca-1k-\longest (AP-1k-L) to \alpaca-52k, \alpagasus-1k, and \lima-1k.
    \textbf{Bottom:} we compare fine-tuning on \evol-1k-\longest (EI-1k-L) to \evol-70k, \evol-\alpagasus-1k (i.e. using the method of \citet{chen2023alpagasus} to subsample \evol-70k), and \lima-1k.
    Our datasets of long responses consistently lead to higher preferences (higher win rate) than the existing methods.
    }
    \label{fig:preference_eval_alpaca_evol}
    
\end{figure*}

\section{Related work}

\textbf{Instruction fine-tuning of LLMs.}
Since pre-trained LLMs usually do not accurately understand user intents and provide coherent and beneficial responses, an instruction fine-tuning stage is necessary \citep{ouyang2022training,bai2022training}. %
Diversity of demonstrations and tasks~\citep{chung2022scaling,xu2022zeroprompt} plays a pivotal role in enhancing the instruction-following performance of LMs. %
InstructGPT~\citep{ouyang2022training} first demonstrated how to achieve impressive performance in handling open-ended queries by fine-tuning GPT-3 models \citep{brown2020language} with RLHF, which led to the release of ChatGPT. %
Subsequently, the community attempted to replicate the exceptional performance of proprietary models %
\citep{wang2023far,xu2023wizardlm,vicuna2023}, %
but \citet{gudibande2023false} show that it might be easy to mimic the style but not the factuality of closed-source LLMs. %
\citet{singhal2023long} identify a strong correlation between response length and reward when doing RLHF, %
implying that optimizing response length might be an implicit goal of RLHF. 
Also, \citet{yuan2024self} show that their self-improved reward model based on DPO encourages more verbose responses.

\textbf{Data selection for IFT.}
The community has focused on creating IFT datasets of high quality~\cite{peng2023instruction}. As one of the pioneering works, Alpaca~\citep{alpaca} collects 52k interactions with the text-davinci-003 model using techniques from Self-Instruct~\citep{wang2022self}. 
However, direct distillation from language models without careful screening inevitably introduces demonstrations with incorrect or ill-favored answers.
To filter these cases out, AlpaGasus~\citep{chen2023alpagasus} measures the quality of each demonstration using a powerful LLM (GPT-3.5-Turbo) as a scorer. %
\citet{touvron2023llama2} note that fewer (in the order of tens of thousands) but higher-quality examples annotated by their own vendors significantly improve their \llama-2-Chat models. The definition of data quality also pertains to other factors, such as the complexity of queries~\cite{xu2023wizardlm}, the difficulty of tasks presented~\cite{mukherjee2023orca} and the diversity of semantics~\cite{lu2023instag}. \citet{zhao2023preliminary} propose to control these factors through an instruction refinement approach, which maintains an instruction semantic tree and yields new instructions by modifying the structure of the semantic tree.
To better reflect human intentions, LIMA~\citep{zhou2023lima} relies on community forums and human labor to curate 1,000 demonstrations with an emphasis on quality and diversity, achieving strong instruction-following ability, surpassing some proprietary LLMs. %
They also formulate the \textit{Superficial Alignment Hypothesis}: the general-purpose capabilities of an LLM mostly come from pre-training, and instruction tuning only guides the LLM to mimic the style, persona, and instruction adherence of desired outputs. 
Finally, while \citet{liu2023makes, cao2023instruction} consider response length as an indicator of example quality, our comprehensive exploration is the first to reveal its reliability and effectiveness in data selection for IFT.

\section{%
Fine-tuning on long instructions is a very strong baseline
} \label{sec:1k_longest_instructions}

We first study the importance of length of the training examples for IFT, and its applicability as a simple and inexpensive heuristic to obtain small but effective IFT datasets. 
Surprisingly, %
we observe that this simple criterion %
can often outperform much more sophisticated existing methods.

\subsection{%
Subsampling high-quality IFT datasets
}

\textbf{Existing methods.}
Recent works have shown that IFT on a small curated dataset of instructions is sufficient to enhance the ability of LLMs to follow instructions and complete tasks. 
In particular, \citet{chen2023alpagasus} adopt %
GPT-3.5-Turbo as the oracle to judge the quality of (instruction, input, output) tuples with grades on a 1-5 scale. Only the highest scoring examples (grade $\geq$ 4.5) from Alpaca-52k (but the same approach can be generalized to other datasets) are used to form the AlpaGasus dataset on 9k instructions.
Later, \citet{zhou2023lima} collect 750 top instruction-response pairs from community forums with some heuristic rules, such as comments and upvotes, and manually write 250 examples to enhance task diversity and quality. These 1,000 examples are optimized for a uniform response style %
to turn the LLM into a useful conversational agent, and constitute the LIMA-1k dataset.

\textbf{Our simple baseline: 1k instructions with the longest responses.}
Though both AlpaGasus and LIMA present promising performance improvements, they require either access to proprietary LLMs or %
very expensive human labor. 
Then, since previous works %
suggest that longer responses naturally arise during alignment \citep{singhal2023long, yuan2024self}, 
we explore response length as the selection criterion to prune IFT datasets. 
\textit{We select the 1,000 longest examples from the popular \alpaca-52k and \evol-70k datasets} to form our IFT datasets that we refer to as \alpaca-1k-\longest and \evol-1k-\longest.
Note that we use the term 1k-\longest examples or responses interchangeably for simplicity in the remaining text, but always refer to the length of the responses.
We restrict ourselves to using 1,000 examples for consistency with \lima and since we are interested in testing how far the instruction following ability of LLMs can be pushed with a minimal SFT dataset. 
Using longer examples can be seen as a natural choice since these are usually more informative and thus contain more features relevant to human intentions. 
Longer responses are also intuitively harder for LLMs to fit, which forces the model to actually learn the response style rather than just memorize the answer. 
In addition, fitting longer responses encourages the model to capture long-distance semantic connections, and stay on-topic when answering complicated instructions. We provide empirical evidence to support our intuition in App.~\ref{app:empirical_proof_intuition_longer_context}.
Interestingly, we observe that the instructions with longest responses minimally overlap with those receiving high score by LLMs: for example, most of the 1k longest examples from \alpaca receive a score of 3.5 from GPT-3.5-Turbo, i.e. signficantly lower than those in \alpagasus (see details in Fig.~\ref{fig:scores_distribution} in App.~\ref{app:scores}).

\subsection{Effectiveness of our approach for open-ended generation} \label{sec:llm_evaluation_longest_response}

\textbf{Setting.}
To test the effectiveness of our approach, 
we compare our 1k-\longest datasets to the full original \alpaca and \evol datasets (52k and 70k examples), %
the 1k examples with highest scores according to GPT-3.5-Turbo as done by \citet{chen2023alpagasus} (hence we refer to these as \alpagasus-1k and \evol-\alpagasus-1k), and \lima-1k.
For each instruction dataset, we fine-tune \llama-2-7B %
base models (complete training configurations in App.~\ref{app:training_hparams}). 
Then, we test their abilities on five evaluation datasets (LIMA, Vicuna, Koala, WizardLM, Self-Instruct, see the description of the datasets in App.~\ref{app:datasets}). We provide head-to-head comparisons in terms of win rate, where GPT-4 judges the preferable response (ties are allowed, details in App.~\ref{app:evaluation_details}). %

\textbf{Results.}
Fig.~\ref{fig:preference_eval_alpaca_evol} shows that the responses of our models fine-tuned on the 1k-longest examples of either \alpaca or \evol  consistently outperform the existing methods across evaluation datasets.
In particular, \alpaca-1k-\longest is largely preferred over all competitors, and has an average win rate of 46.3\% vs. \lima-1k, with only 28.3\% of losses (see Fig.~\ref{fig:teaser}). 
This performance is significant when considering that \lima has been carefully curated \textit{manually} while our instructions come from a simpler dataset and selected only according to their length.
Similarly, \evol-1k-\longest clearly outperforms \lima-1k and the full \evol-72k, while it has a smaller but consistent advantage over \evol-\alpagasus-1k. 
We hypothesize that the advantage is smaller on \evol because \evol contains higher-quality data than \alpaca, thus even selecting examples using GPT-3.5-Turbo scores can find relatively effective training examples. 
Finally, to exclude the possibility of overfitting to GPT-4 preferences, we repeat this evaluation with \palm as judge and even in this case our models are largely preferred (see Fig.~\ref{fig:preference_eval_on_alpaca_evol_palm2} in App.~\ref{app:palm_evaluation}).

\textbf{Role of response length.} 
As frontier LLMs like GPT-4 might be biased to favor longer responses \citep{zheng2023judging}, Fig.~\ref{fig:teaser} additionally illustrates the average length (as number of tokens) of the responses in several datasets described above, as well as the average length of the responses generated by the LLMs fine-tuned on them during evaluation (on 1030 new instructions from the 5 evaluation datasets). %
As expected, both training and generated answers of \alpaca-1k-\longest are longer than those of \alpaca and \alpagasus.
Interestingly, the training examples of \lima-1k are more than two times longer than those of \alpaca-1k-\longest, while the generated responses of the two models are similar.
We conclude that the length of the responses is not the main factor for our model being consistently preferred to \lima-1k.

\begin{figure}[t]
    \centering
    \vspace{-2mm}
    \subfigure{
        \includegraphics[width=0.48\textwidth]{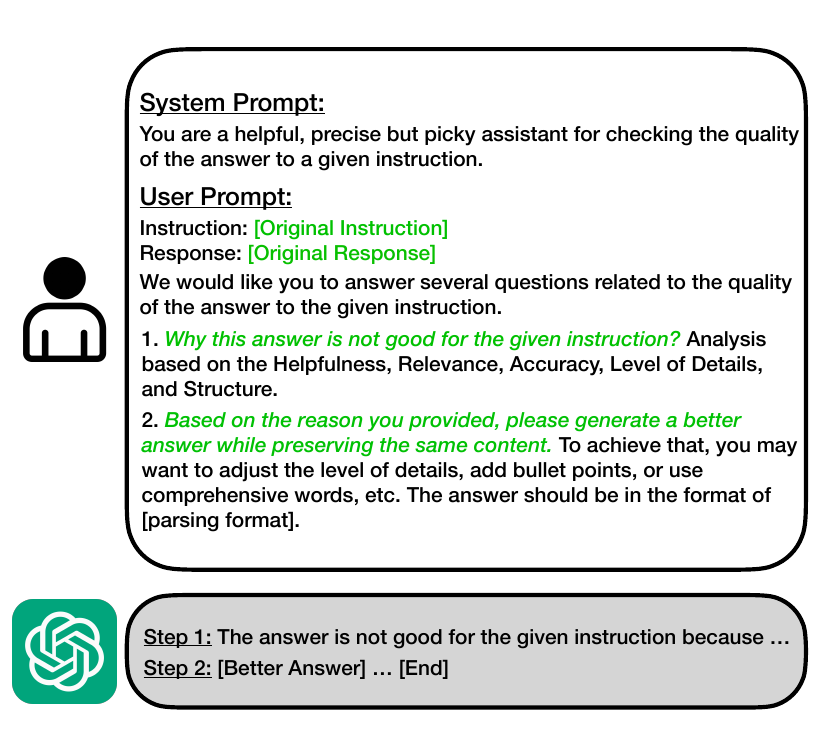}
    }
    \vspace{-10mm}
    \caption{\textbf{The template of introspection prompting} used to refine the responses in terms of style, structure, and the level of details.
    }
    \label{fig:refine_via_introspection_prompting}
    
\end{figure}

\section{How far can we go with 1,000 instructions?}

\begin{figure*}[t]
    \centering
        
    \subfigure{
        \includegraphics[width=\textwidth]{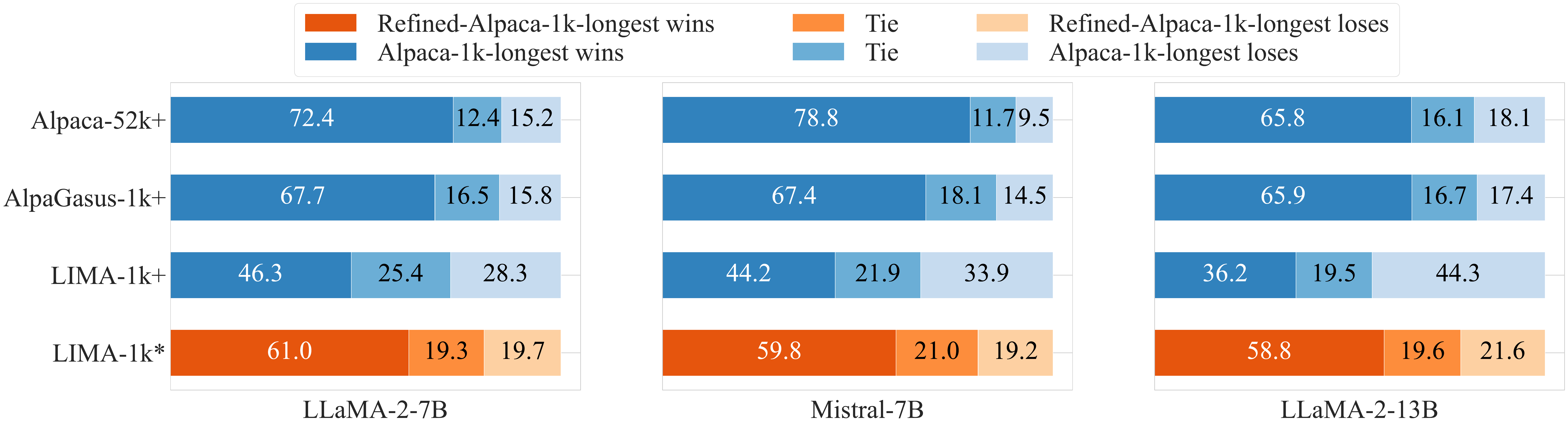}
        
    }

    \vspace{-3mm}
    \caption{\textbf{Refinement via introspection improves instruction-following performance across architectures.} We report the average preference performance (\%) across five evaluation sets using GPT-4 as a judge.
    We show win rate of models with different architectures fine-tuned on \alpaca-1k-\longest against \alpaca-52k, \alpagasus-1k, and \lima-1k in blue (+ symbol). Additionally we illustrate the improvement brought by our \refined-\alpaca-1k-\longest over \lima-1k, the strongest baseline, in red (* symbol).
    }
    \label{fig:avg_win_rate_refined}
\end{figure*}

In Sec.~\ref{sec:1k_longest_instructions} we have shown that length is a strong heuristic to select which instructions to use for IFT. However, the resulting LLMs still fall short compared to those fine-tuned with either more sophisticated (proprietary) pools of instructions or especially preference data e.g. via RLHF.
Then, in the following, we want to explore the limit of the ability that can be achieved from SFT on 1k examples.
For this, we first refine the style of the \longest-1k instructions to be more amenable for IFT.
Second, we show that our dataset and \neftune \citep{jain2023neftune}, a recent algorithm to improve IFT via noise augmentation, can be successfully combined.
Finally, we %
test that the ability of our models in instruction-following evaluations (1) is stable even when forcibly changing the response length, and (2) does not negatively impact their performance on factual knowledge benchmarks.

\subsection{Refining the instructions via introspection}

\begin{table}[h!]
    \caption{\textbf{Preference evaluation results on AlpacaEval 2.0.} The evaluator used to measure instruction-following performance comprehensively considers quality, price, time, variance, and length bias. For our models, if not specified otherwise, we use a limit of 2048 tokens for generation. * denotes results which are directly copied from the AlpacaEval 2.0 leaderboard.
    }
    \vspace{2mm}
    \centering
    \small \tabcolsep=2pt
    \extrarowheight=1.5pt
        \begin{tabular}{L{44mm}  *{4}{C{8mm}}}
            \toprule
             Models                             & \# IFT Data & \# Pref. Data & Win Rate & Avg. Length \\
             \midrule
             \addlinespace[2mm]
             \multicolumn{5}{l}{\textbf{Notable baselines}} \\
             \toprule
             \rowindent GPT-4-Turbo*                                  & ?           & ?             & $50.0$        & $2049$      \\
             \rowindent Alpaca-7B*                        &  52k         & 0             & $2.59$        & $396$       \\
             \rowindent Vicuna-7B*                        &  70k         & 0             & $4.16$        & $1044$      \\
             \midrule
             \addlinespace[2mm]
             \multicolumn{5}{l}{Base model: \textbf{\llama-2-7B}} \\
             \toprule
             
             \rowindent \llama-2-Chat-7B*                 &  27k         & 3M            & $4.96$        & $1479$      \\
             \rowindent %
             \quad\quad  + Evol70k-NEFTune* &  97k   & 3M            & $7.60$        & $1612$      \\
             \rowindent Tulu-2-DPO-7B*                    &  326k        & 64k           & $8.20$        & $1663$      \\
             \hdashline
             \rowindent AlpaGasus-1k          &    1k          & 0             & $2.69$        & $745$      \\
             \rowindent LIMA-1k                 & 1k          & 0             & $2.74$        & $1360$      \\
             \rowindent Alpaca-52k            &  52k         & 0             & $2.74$        & $586$       \\
             
            \rowindent Alpaca-1k-\longest           &  1k          & 0             &   $3.16$        & $1810$    \\
             \rowindent \quad \quad +   max gen. 2048 $\rightarrow$ 4096     &  1k   & 0      &   $3.11$        & $2290$    \\
             \rowindent Evol-Instruct-70k           & 70k          & 0             & $3.44$        & $850$      \\
             
\rowindent Evol-Instruct-1k-\longest           &  1k          & 0             &   $4.09$        & $1866$    \\
              \rowindent \quad \quad + max gen. 2048 $\rightarrow$ 4096     &  1k   & 0      &$4.16$        & $2486$      \\
             
             \rowindent \evol-\alpagasus-1k   &  1k          & 0             & $4.32$        & $1156$      \\
             \rowindent \refined-Evol-Instruct-1k-\longest   &  1k          & 0             & $5.12$        & $1289$      \\
             
             \rowindent Refined-Alpaca-1k-\longest   &  1k          & 0             & $6.00$ & $1732$ \\
             \rowindent \quad \quad + max gen. 2048 $\rightarrow$ 4096 & 1k & 0 &$6.03$        & $2326$ \\
             \rowindent \quad \quad + NEFTune             &  1k          & 0             &  $7.88$        & $1801$\\
             \rowindent \quad \quad + NEFTune  +      2048 $\rightarrow$ 4096       &  1k          & 0             &$7.83$        & $2478$ \\
             \midrule
             \addlinespace[2mm]
             
             \multicolumn{5}{l}{Base model: \textbf{Mistral-7B-v0.1}} \\
             \toprule
             \rowindent \alpaca-52k    &  52k          & 0             & $3.42$ & $450$      \\
             \rowindent \alpagasus-1k    &  1k          & 0             & $4.91$    & $502$      \\
             \rowindent \lima-1k    &  1k          & 0             & $6.76$    & $1197$      \\
             \rowindent \alpaca-1k-\longest    &  1k          & 0             & $7.13$    & $937$      \\
             \rowindent Refined-Alpaca-1k-\longest    &  1k          & 0             &   $11.74$        & $1170$    \\
             \rowindent \quad \quad + max gen. 2048 $\rightarrow$ 4096   &  1k          & 0             & $11.76$ & $1330$ \\
             \rowindent \quad \quad + \neftune & 1k & 0 & $11.94$ & $1199$ \\
             \midrule
             \addlinespace[2mm]
             
             \multicolumn{5}{l}{Base model: \textbf{\llama-2-13B}} \\
             \toprule
             
             \rowindent\alpaca-52k      & 52k          & 0             & $3.90$        & $556$      \\
             \rowindent\alpaca-1k-\longest    &    1k          & 0             & $4.80$        & $1104$      \\
             \rowindent\alpagasus-1k    &  1k          & 0             & $4.87$        & $540$      \\
             \rowindent\lima-1k    &  1k          & 0             & $5.64$        & $1097$      \\
             \rowindent
             \refined-\alpaca-1k-\longest    &  1k          & 0             & $8.44$        & $1646$      \\
             \rowindent \quad \quad + max gen. 2048 $\rightarrow$ 4096   &  1k          & 0             & $8.30$ & $2244$ \\
             \rowindent \quad \quad + \neftune    &  1k          & 0             & $8.76$    & $1582$     \\
             \bottomrule
        \end{tabular}
    
    \label{tab:alpaca_eval2_small}
    \end{table}

As suggested by %
\citet{zhou2023lima}, the goal of IFT is to teach LLMs the format to employ when interacting with the users rather than instilling new knowledge.
We argue that fine-tuning on rich and detailed instructions may improve the ability of the models to capture deeper semantic structure and logic. %
Then, we want to refine our 1k-\longest examples to improve the quality of responses of training examples in terms of style, structure and the level of detail.
In fact, there is no guarantee that the instructions selected by length also have high quality in terms of structure, glossary and logic.

Given that LLMs are surprisingly good at self-improving~\citep{huang2022large,pan2023automatically} and judging~\citep{zheng2023judging,alpaca_eval}, we propose using an Oracle LLM for this task, via encouraging it to introspect.
In particular, inspired by Chain-of-Thought prompting~\citep{wei2022chain}, we prompt the GPT-3.5-Turbo model to produce a brief review of the original response given the instruction, followed by a new response generation process that has access to the original instruction-response pair and the introspection output. The details of the prompt are presented in Fig.~\ref{fig:refine_via_introspection_prompting}.
Applying this procedure to the 1k-\longest examples of \alpaca and \evol we obtain new IFT datasets: \refined-\alpaca-1k-\longest and \refined-\evol-1k-\longest.

\subsection{Instruction-following evaluation}

\textbf{Setup.}
First, we provide a pairwise comparison between fine-tuning different LLMs on our \refined-1k-\longest and baseline datasets, in particular \lima-1k.
Next, to facilitate a unified comparison of all models and position them among existing baselines, we compute their performance on AlpacaEval 2.0 \citep{alpaca_eval} and MT-Bench~\citep{zheng2023judging}. 
This allows us to compare many LLMs, including those reported on the existing leaderboards by previous works, more efficiently than with pairwise analyses.

\textbf{Head-to-head comparisons.} 
We compare fine-tuning on our \refined-\alpaca-1k-\longest to the \alpaca-1k-\longest against \alpaca-52k , \alpagasus-1k  and \lima-1k in a head-to-head fashion: Fig.~\ref{fig:avg_win_rate_refined} reports the average (over the 5 evaluation datasets introduced in Sec.~\ref{sec:llm_evaluation_longest_response}) preference of GPT-4, repeated for three base models, i.e. \llama-2-7B, Mistral-7B-v0.1%
, \llama-2-13B (the corresponding results with \palm as judge are shown in Fig.~\ref{fig:avg_win_rate_refined_palm2} in App.~\ref{app:palm_evaluation}).
In all cases the models fine-tuned on the plain \alpaca-1k-\longest already outperform the baselines with the exception of \lima-1k for \llama-2-13B.
In particular, \lima-1k makes the strongest existing method: however, when we compare it with our \refined-\alpaca-1k-\longest, this last one has a significant advantage over \lima-1k, with an average win rate of 59.9\% across architectures vs the 20.2\% of \lima. %
This shows the effectiveness of the refinement via introspection on the longest examples from \alpaca, even when used by different base models. 
Moreover, we complement the head-to-head comparisons by conducting human-based evaluations, which reflect the preferences of real users. Concretely, we design a user study (see more details in App.~\ref{app:evaluation_details}) to compare the responses generated by our \alpaca-1k-\longest to those of \alpaca-52k, with \llama-2-7B as the base model. Note that we instruct the evaluators not to consider the length of the responses in their judgment. Finally, we collect 425 human preferences, over which \alpaca-1k-\longest obtains 71.0\% win rate, which agrees with the conclusion of the LLMs as judges.

\textbf{AlpacaEval 2.0 evaluation.}
In Table~\ref{tab:alpaca_eval2_small} we report the results on the AlpacaEval 2.0 benchmark of our models and some baselines copied from the public leaderboard.\footnote{\url{https://tatsu-lab.github.io/alpaca_eval/}}
Moreover, we show the architecture, size of IFT and preference datasets, and average response length for each entry.
Among \llama-2-7B models, both \lima-1k and \alpaca-52k fine-tuned models achieve win rate below 3\%, which is outperformed by \alpaca-1k-\longest (3.11\%).
Switching to the instructions refined by introspection (\refined-\alpaca-1k-\longest) almost doubles the win rate, achieving 6.00\%, which even surpasses the original \llama-2-Chat-7B, fine-tuned with 27k instructions and 3M preference pairs. %
Since \citet{jain2023neftune} showed that \neftune, which injects noise on the embedded inputs as augmentation, can improve the performance of IFT, we test it in combination with our dataset: this yields 7.88\% win rate, i.e. the second best \llama-2-7B model appearing on the leaderboard, ahead of  \llama-2-7B-\evol-\neftune \citep{jain2023neftune} and not far from the 8.20\% win rate of Tulu-2-DPO-7B~\citep{ivison2023camels}. %
Interestingly, fine-tuned models with similar average response lengths, exhibit wildly distinct win rates. For example, when we refine \alpaca-1k-\longest via introspection and enable NEFTune in fine-tuning, the win rate rises from 3.16\% to 7.88\%, while the average response lengths are almost the same.
Overall, these results illustrate how using a simple dataset of 1,000 instructions which did not necessitate any manual curation can compete with more expensive and sophisticated alignment schemes relying on SFT with hundreds of thousands of examples and involving RLHF on up to 3M preference pairs. %
Moreover, we observe similar %
behavior with other architectures: %
for Mistral-7B-v0.1 \alpaca-1k-\longest already outperforms the baseline methods, but the refined instructions give the most notable increase (7.13\% to 11.74\%) in win rate.
Similarly, \refined-\alpaca-1k-\longest attains the best results for \llama-2-13B.
Interestingly, unlike for \llama-2-7B, in these cases the improvements given by \neftune are marginal (%
$\leq$ 0.32\%), which highlights the importance of the fine-tuning dataset.
Furthermore, we surprisingly find that \refined-\evol-1k-\longest (5.12\%) underperforms compared to \refined-\alpaca-1k-\longest (6.00\%), which may be attributed to the limited capability of GPT-3.5-Turbo in understanding long-form text (see details of average response lengths in Fig.~\ref{fig:train_data_len_distribution}), such as demonstrations consisting of thousands of tokens.

\begin{table}[t]
    \caption{\textbf{Single-score evaluation results on MT-Bench across different base LLMs and fine-tuning datasets.} MT-Bench assesses the quality of generated answers using GPT-4 as the judge, which scores on a 1-10 scale. %
    }
    \vspace{2mm}
    \centering
    \small \tabcolsep=1.pt
    \begin{tabular}{L{35mm} C{15mm} C{15mm} C{15mm}}
    \toprule
    Datasets        &  \llama-2-7B & \llama-2-13B & Mistral-7B-v0.1 \\
    \midrule
    \alpaca-52k     &  $3.74$      & $5.40$       & $5.35$          \\ 
    \alpagasus-1k   &  $3.63$      & $4.70$       & $6.06$          \\ 
    \lima-1k        &  $3.95$      & $5.18$       & $\textbf{\underline{6.18}}$ \\
    \midrule
    \alpaca-1k-\longest   &  $3.96$      & $5.32$       & $5.80$    \\ 
    \refined-\alpaca-1k-\longest   &  $\underline{4.18}$      & $\textbf{6.09}$       & $6.00$          \\ 
    \rowindent \quad + NEFTune   &  $\textbf{4.28}$      & $\underline{5.98}$       & $\textbf{\underline{6.18}}$          \\ 
    \bottomrule
    \end{tabular}
    \label{tab:mtbench_small}
\end{table}

\begin{figure*}[t]
    \centering
    \small
    \begin{tabular}{c}
    \includegraphics[width=\linewidth]{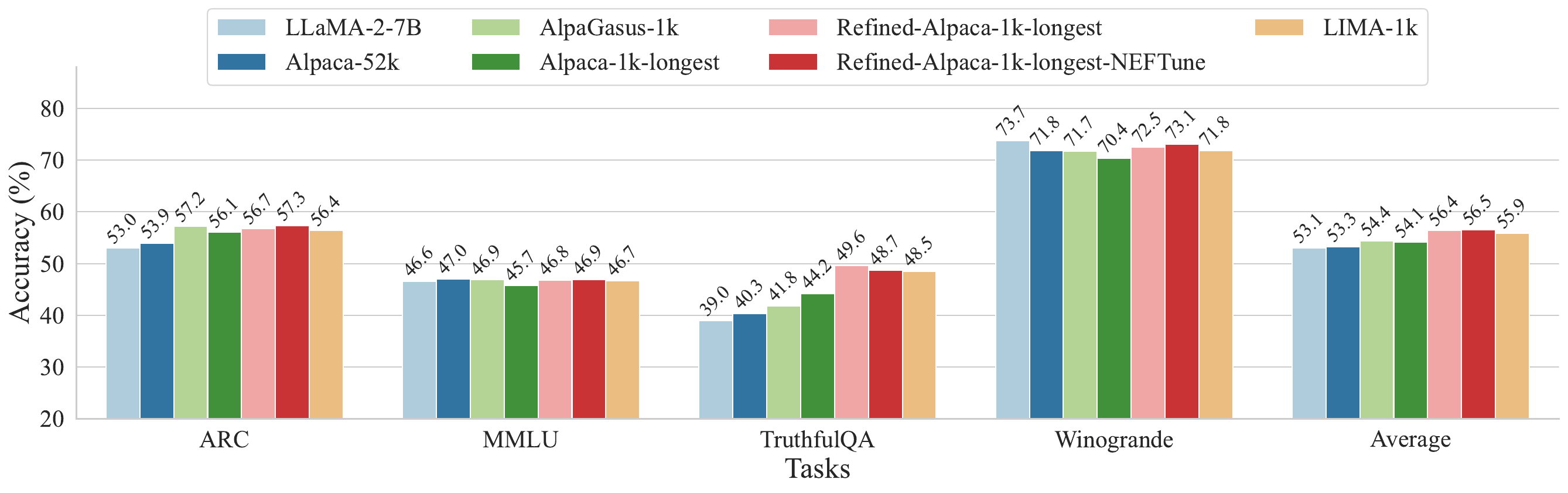}\\

    \end{tabular}
    \vspace{-5mm}
    \caption{\textbf{Open LLM Leaderboard tasks with \llama-2-7B fine-tuned on %
    \alpaca-based datasets and \lima.} %
    The model fine-tuned on Alpaca-1k-longest achieves comparable performance to that of AlpaGasus-1k on average, showing that the performance gain on instruction-following capability does not compromise factuality. %
    Our \refined-\alpaca-1k-\longest, with and without NEFTune, %
    achieve the best results,
    surpassing LIMA-1k on %
    all datasets.
    }
    \label{fig:openllm}
\end{figure*}

\textbf{Changing response length does not affect quality.}
As shown in Table~\ref{tab:alpaca_eval2_small}, the LLMs fine-tuned on (\refined-)1k-\longest lead to longer generation than most competitors.
To test if longer replies are sufficient for higher scores on AlpacaEval 2.0, we increase the maximum number of generated tokens from the default 2048 (used for all baselines as well) to 4096.
This makes the average response length of our best \llama-2-7B model (refined dataset with \neftune) to increase from 1801 to 2478. However, this slightly degrades win rate (-0.05\%).
Similar small variations can be also observed for other models and architectures (see Table~\ref{tab:alpaca_eval2_small}).
Then, length alone does not significantly influence the results on the benchmark.

\textbf{MT-Bench evaluation.}
We show the score-based results on MT-Bench of LLMs fine-tuned on different instruction datasets in Table~\ref{tab:mtbench_small}.
The baselines (\alpaca-52k, \alpagasus-1k, and \lima-1k) achieve scores below 4 when employing \llama-2-7B as the base model. \alpaca-1k-\longest without refinement (3.96) already matches the best of them, and refining the 1,000 raw instructions %
yields a 4.18 MT-Bench score.
Switching to \llama-2-13B and Mistral-7B-v0.1, we show consistent improvements of using \alpaca-1k-\longest and its refined variants compared to baselines.
For all base models, unlike what we observe in AlpacaEval 2.0, applying \neftune does not consistently lead to stronger instruction-following capability, as indicated by MT-Bench scores.

\subsection{Evaluation on factual knowledge benchmarks}

In the following, we study how the models trained on small instruction datasets behave in tasks other than instruction following with an LLM as a judge, and the shortcomings it entails.
For this, we evaluate them on a subset of the Open LLM benchmark: it includes six datasets, from which we exclude HellaSwag because it contains examples also present in the training set of \lima-1k (see discussion in App.~\ref{app:lima_contam}) and GSM80K since all models fail to achieve non trivial performance, which assess several abilities of an LLM including commonsense reasoning, multitask knowledge and truthfulness, at various difficulty levels. %

Fig.~\ref{fig:openllm} reports the results of the models fine-tuned from \llama-2-7B %
on the dataset derived from \alpaca and \lima-1k (the corresponding evaluations for other architectures and \evol-based datasets can be found in App.~\ref{app:mistral_7b_llama2_13b_openllm}%
).
We observe that, on average over the datasets, IFT on \alpaca-52k yields marginal improvement over the base model, while both \alpagasus-1k and 1k-\longest give around a 1\% increase. Significantly better results are achieved by \lima-1k, with 55.9\% vs 53.1\% of the base model.
However, the two models relying on \alpaca-\refined-1k-\longest, without and with \neftune, are the best performing ones with 56.4\% and 56.5\% (without and with \neftune respectively).
This suggests that the IFT dataset might have an effect beyond quality of user interactions. %
In fact, all LLMs are fine-tuned from the same base model, thus we can assume that they have the same factual knowledge, and the different performance is due to how well the alignment phase teaches the model how to follow the right steps to accomplish a given task.
We hypothesize that using longer and more detailed instructions, which force the LLM to better capture the semantics of the task at hand, might positively influence the performance on quantitative (e.g. multiple choice questions answering) tasks as those in Open LLM.

\begin{figure}[!t]
    \centering
    \subfigure[\small Postpone the EOS token (base model: \llama-2-7B)]{
        \includegraphics[width=%
        \columnwidth]{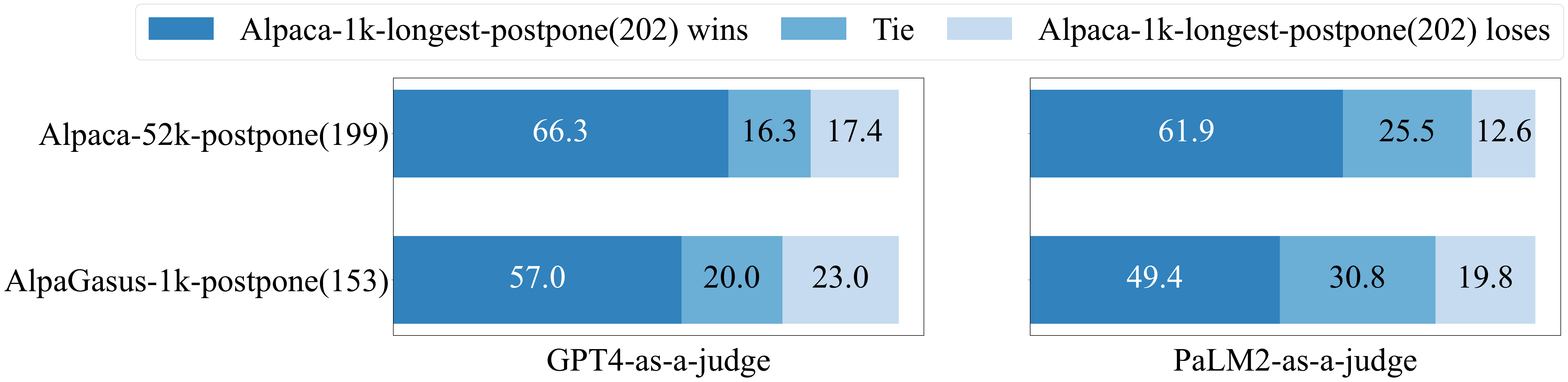}
        \label{fig:llama_2_7b_alpaca_similar_length_multiple_judges}
    }
    \subfigure[\small Prompting strategy (base model: Mistral-7B-v0.1)]{
        \includegraphics[width=%
        \columnwidth]{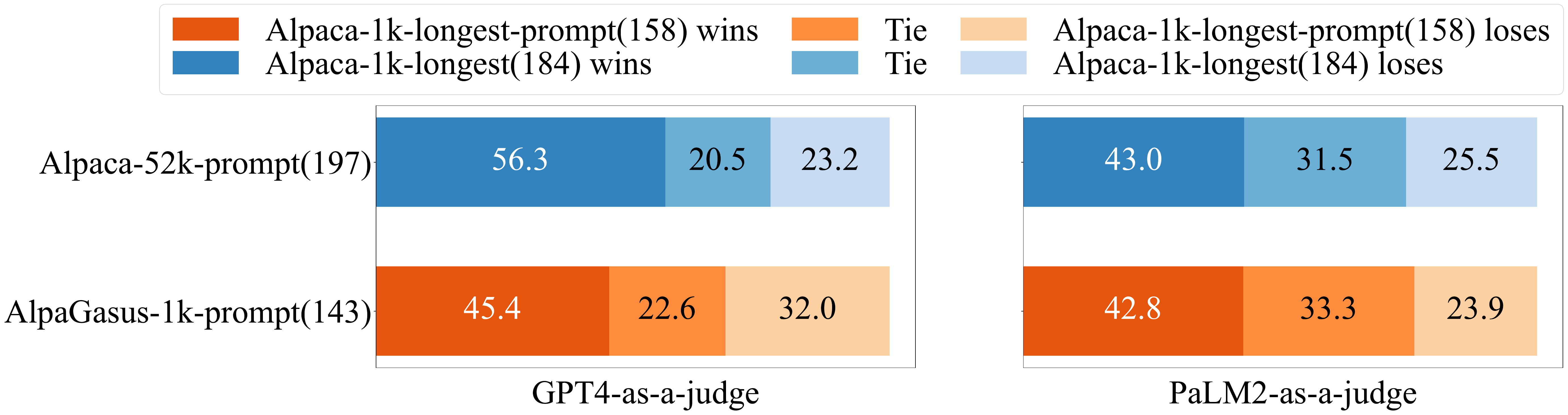}
        \label{fig:mistral_7b_alpaca_similar_length_multiple_judges}
    }
    \subfigure[\small Average number of tokens in responses]{
        \includegraphics[width=%
        \columnwidth, clip, trim=0mm 0mm 7mm 0mm]{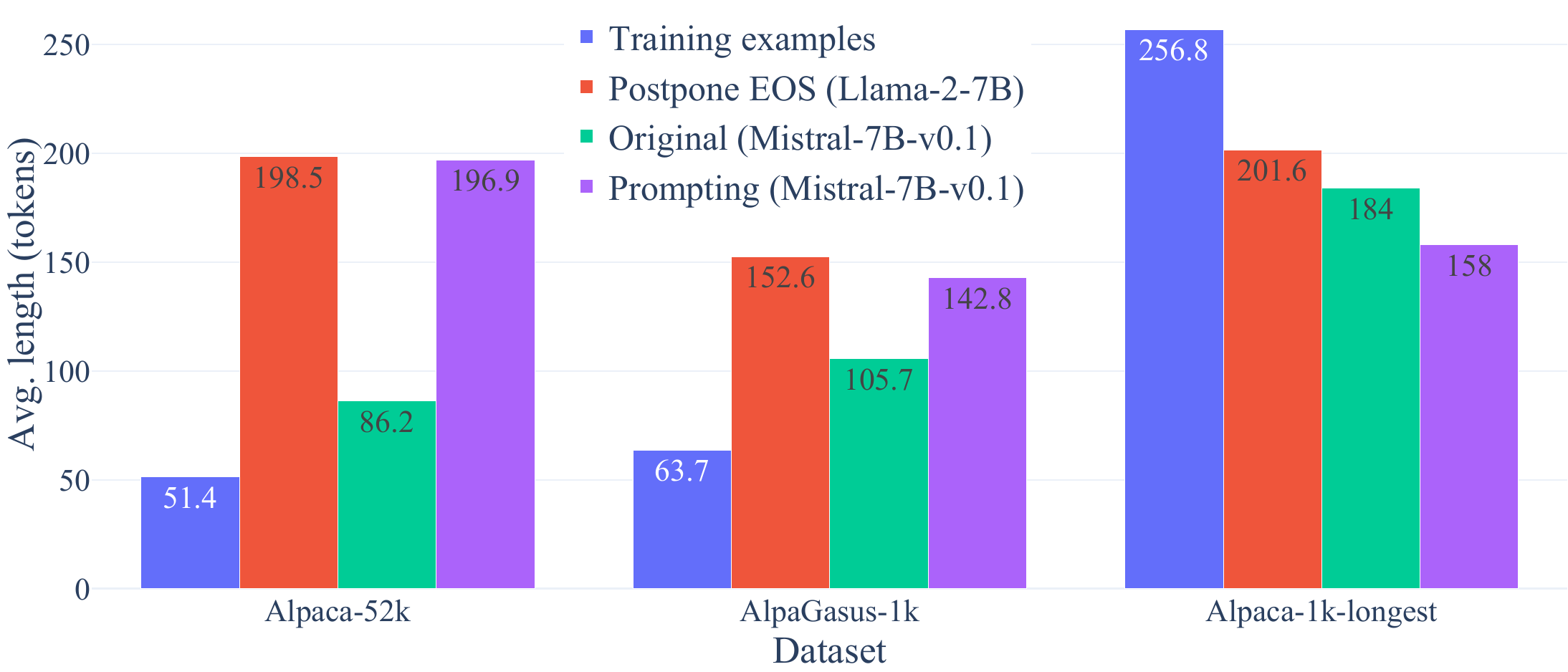}
        \label{fig:num_tokens_dist_similar_length}
    }
    \vspace{-3mm}
    \caption{\textbf{Preference evaluation on generations of similar length.} 
    We control the response length (number of tokens) of different fine-tuned models via (a) postponing the appearance of 
    the end-of-sentence (EOS) token, or (b) designing specialized prompts. The value in parenthesis denotes the average response length of each model.
    We show the resulting average response length of different fine-tuned models (c) (\textit{Original} means %
    using of the original prompt and inference configuration). %
    The generations of our \alpaca-1k-\longest are significantly preferred by GPT-4 and PaLM-2 judges %
    in both length-controlled generation experiments.}
    \label{fig:preference_eval_similar_length}
    
\end{figure}

\section{%
Additional analyses of our models
}

While we uncover the effectiveness of fine-tuning on instructions with long responses, the reason for this success remains elusive. In the following we provide some insights about this phenomenon.

\begin{figure}[!t]
    \centering
    \includegraphics[width=.95\linewidth]{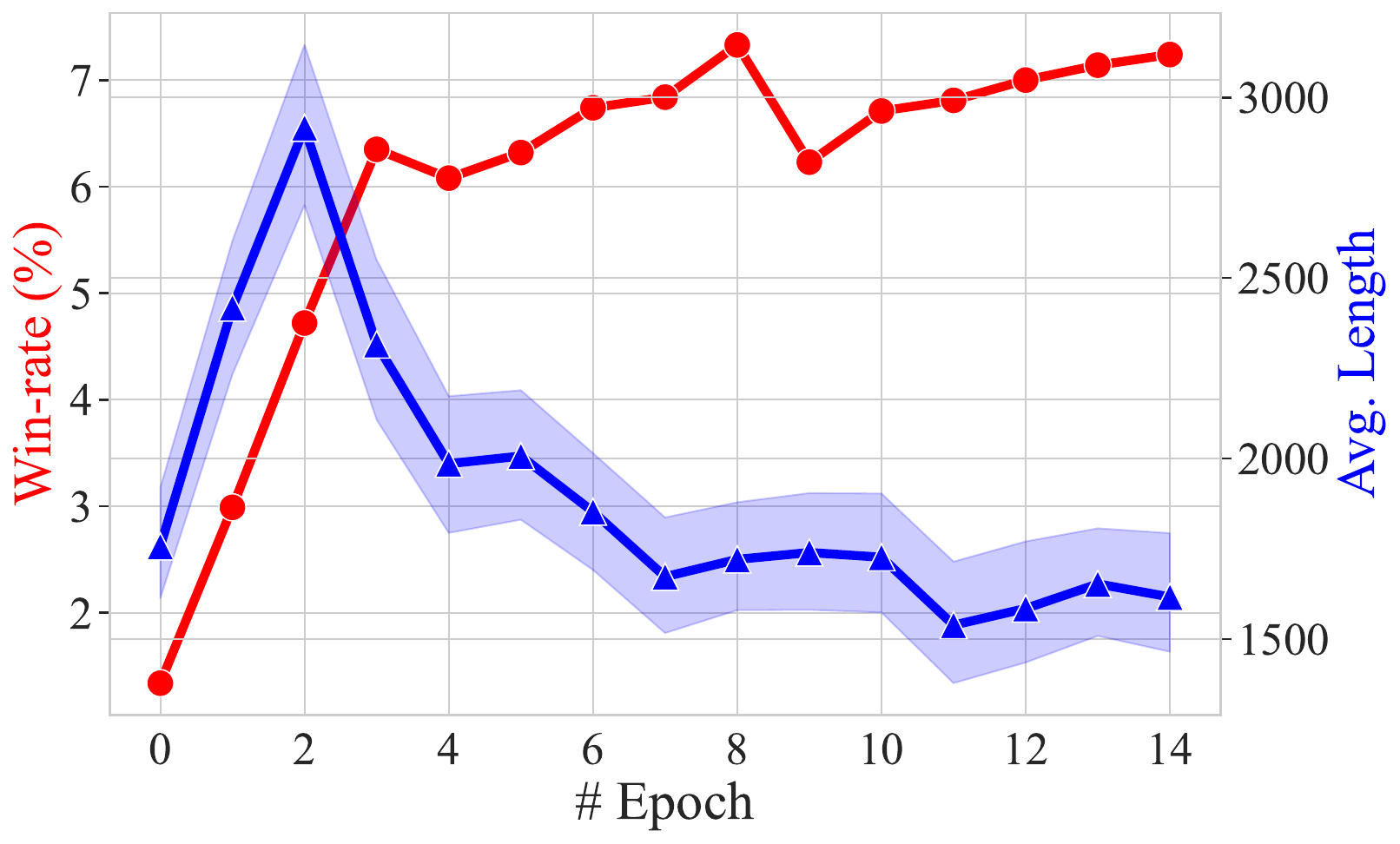}
    \vspace{-2mm}
    \caption{\textbf{Performance of the \llama-2-7B model fine-tuned on refined-Alpaca-1k-longest across different epochs.} A NEFTune noise level of 5 is used in fine-tuning and the win-rates are calculated following the evaluation protocol from AlpacaEval 2.0.}
    \label{fig:llama2_win_rate_length_over_different_epochs}
    \vspace{-10pt}
\end{figure}

\textbf{Comparison on generations of similar length.}
To further support the idea that the length of responses does not explain our models' performance, we artificially increase the response length of the replies from \llama-2-7B models fine-tuned on \alpaca-52k and \alpagasus-1k. 
This extension is achieved by ensuring that the end-of-sentence token does not appear until after the 150th token has been generated.
Fig.~\ref{fig:num_tokens_dist_similar_length} shows that this adjustment makes both baselines output responses of similar length as our \alpaca-1k-\longest. 
However, even in this case, both GPT-4 and \palm judges still significantly prefer our \alpaca-\longest-1k model (Fig.~\ref{fig:llama_2_7b_alpaca_similar_length_multiple_judges}), indicating that artificially increasing the number of generated tokens does not effectively enhance response quality.
Meanwhile, we further test using prompting strategies to control the response length of all models, which might be a more natural choice than directly postponing the appearance of the end-of-sentence token. After extensive exploration with Mistral-7B-v0.1 models (note that none of the prompting strategies we tried are effective with \llama-2-7B models), we could increase the response length of \alpaca-52k and \alpagasus-1k models by asking in the prompt to use $N$ paragraphs, and reduce the response length of our \alpaca-1k-\longest by asking to ``answer in as few words as possible''. When conducting head-to-head comparisons, we ensure that there is minimal variation in average length between the two candidates and have \alpaca-52k-prompt (197 tokens) vs \alpaca-1k-\longest (184 tokens) and \alpagasus-1k-prompt (143 tokens) vs \alpaca-1k-\longest-prompt (158 tokens). The results in Fig.~\ref{fig:mistral_7b_alpaca_similar_length_multiple_judges} indicate that our \alpaca-1k-\longest outperforms both length-controlled counterparts across the two judges, which aligns with the conclusion we obtain in Fig.~\ref{fig:llama_2_7b_alpaca_similar_length_multiple_judges}.

\textbf{Length and win rate are anticorrelated during fine-tuning.}
We track the average length of replies over epochs when fine-tuning for one of our models. As shown in Fig.~\ref{fig:llama2_win_rate_length_over_different_epochs}, except for the early stage of fine-tuning, the response length progressively decreases while the win rate keeps improving.
This indicates that the model does not  simply learn  to output long generations from long training examples but also to produce more refined structures. %

\begin{figure}[h]
    \centering
    \begin{tabular}{c}
    \includegraphics[width=0.46\textwidth]{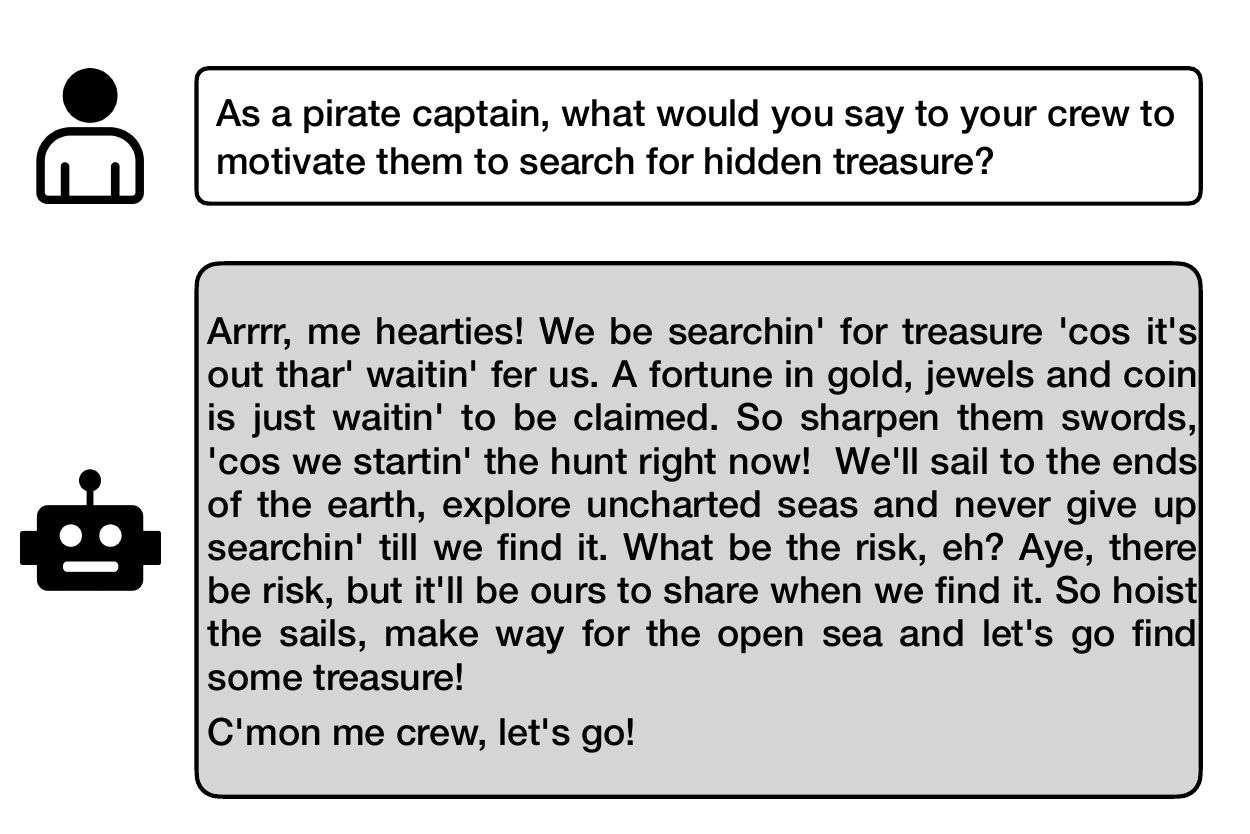}\\
        
    \includegraphics[width=0.46\textwidth]{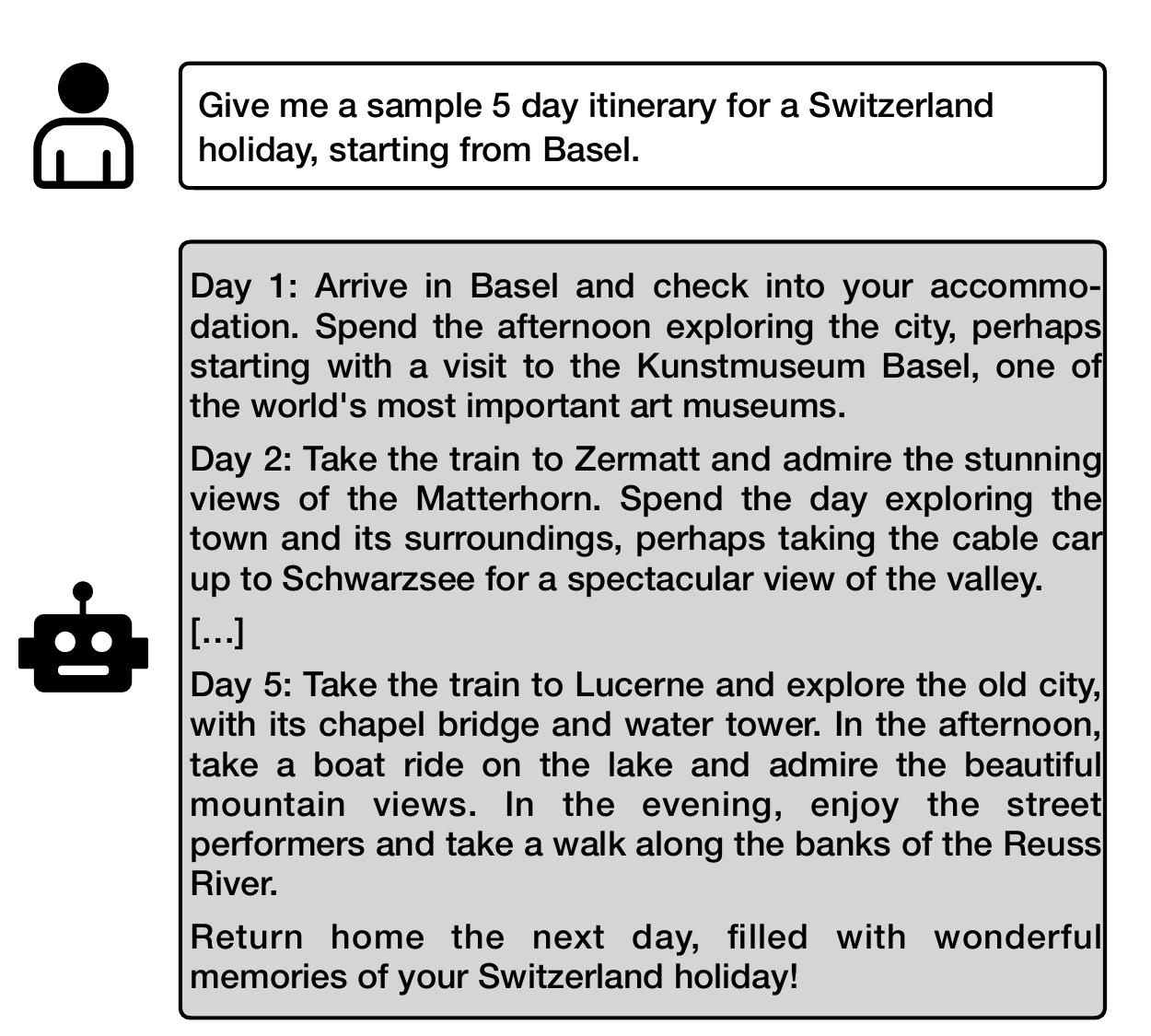}
    \end{tabular}

    \caption{\textbf{Example generations.} Case study to illustrate the instruction-following performance of \llama-2-7B model fine-tuned on \alpaca-1k-\longest.}
    \label{fig:case_study_main_2}
\end{figure}

\textbf{Example generations.}
In Fig.~\ref{fig:case_study_main_2} we provide two examples of completions generated by our \llama-2-7B model fine-tuned on the \alpaca-1k-\longest dataset. We see that the LLM provides organic and detailed responses.
We provide an extended qualitative comparison to other models in App.~\ref{app:case_study}, where one can see that, for example, LIMA can sometimes lead to repetitive outputs while 1k-\longest models tend to have a more engaging tone. 

\textbf{Additional comparisons.}
In App.~\ref{app:diversity}, we show how our approach to data selection works in concert with other sampling techniques that promote diversity. Moreover, we verify that our 1k-\longest models remain effective on text summarization tasks (see results in Table~\ref{tab:long_form_summ}), in which concise answers are preferred. For space reasons, we defer to the appendix the comparison of our \alpaca-1k-\longest to two additional baselines, \alpagasus-9k and the dataset obtained improving \alpaca-52k with reflection-tuning in \citet{li2023reflection}. As shown in App.~\ref{app:alpagasus-9k} and App.~\ref{app:introspection_advantage} respectively, our approach consistently outperforms both baselines.

\section{Discussion
}

\textbf{Quality of the instructions in IFT.}
\citet{chen2023alpagasus} and \citet{zhou2023lima} argue that IFT requires high-quality training examples and use different proxies for quality to create the \alpagasus and \lima datasets.
However, our experiments demonstrate that a simple heuristic for selecting training instructions, such as the length of the response, leads to better-performing models.
It is important to note that length alone is not sufficient. For example, the \lima training examples are on average twice as long as those in \alpaca-1k-\longest. Additionally, we emphasize that length does not necessarily reflect quality, as illustrated by the lower scores given by GPT-3.5-Turbo to the examples in our \alpaca-1k-\longest (Fig.~\ref{fig:scores_distribution}). This suggests that other factors come into play when determining the effectiveness of IFT datasets. As a result, it remains uncertain which specific components in the fine-tuning dataset are crucial for achieving the best model performance.

\textbf{IFT can improve factuality.}
\citet{gudibande2023false} show the possibility of fine-tuning LLMs to imitate the style of ChatGPT. They achieve this by using ChatGPT's responses as an IFT dataset, which can consist of up to 150 million tokens. Remarkably, both human evaluators and LLM-as-a-judge evaluators rate the responses generated by these fine-tuned models nearly as high as those generated by ChatGPT. However,  this fine-tuning approach does not enhance, and in some cases even diminishes, the performance of these models on NLP benchmarks compared to the base model. A similar observation is made by \citet{jha2023limit}, who suggest that \lima-1k (when used to fine-tune the MPT models from \citet{MosaicML2023mpt7B-llm}) does not yield the same level of performance as \alpaca-52k on tasks that do not rely on automated evaluation by an LLM.
In contrast, we demonstrate that IFT can lead to both a stronger preference from various LLMs serving as judges and improved performance on Open LLM tasks. However, it is key to carefully select the instruction dataset for this purpose.
The question of systematically constructing optimal IFT datasets remains an open challenge.

\textbf{Role of length bias of LLMs as judges.}
It is a relevant question whether training on the longest examples simply exploits the bias of LLMs as judges for longer replies rather than leading to inherently better models.
Therefore, we have conducted experiments specifically designed to exclude this scenario, like equating the response length of different LLMs and the human evaluation.
Moreover, we benchmark the ability of the models both with head-to-head comparisons, single-score evaluation, and on factual knowledge datasets.
Since all these analyses show that our models outperform the baselines across datasets, tasks and evaluation methods, we consider them conclusive evidence that our results are not due to a length bias of the LLMs as judges.

\textbf{Conclusions.}
In this work we have shown that using reply length as a heuristic can effectively pre-select instructions for LLMs alignment in SFT. Moreover, a straightforward refinement step is enough to create a dataset of only 1k instruction-response pairs which yields competitive results compared to complex alignment methods like RLHF and DPO. 
Thus, this approach constitutes an inexpensive yet strong baseline for future works on alignment.
Our analysis also challenges the current understanding of high-quality IFT datasets and their impact on fine-tuned model performance in standard NLP benchmarks. 
We emphasize that a major aspect of alignment concerns mitigation of safety risks and ethical use of LLMs.
We have not explored this aspect here, as it demands task-specific approaches.

\section*{Impact statement}
This paper presents work whose goal is to advance the field of Machine Learning. There are many potential societal consequences of our work, none which we feel must be specifically highlighted here.

\section*{Acknowledgements} 
We thank the anonymous reviewers of ICML for insightful comments that have helped to improve the quality of the paper. 
M.A. was supported by the Google Fellowship and Open Phil AI Fellowship.

\bibliographystyle{icml2024}

\newpage
\appendix
\onecolumn
\section{Experimental details}
\label{app:exp_details}

\subsection{IFT datasets} \label{app:datasets}
This section contains a list of instruction fine-tuning datasets that appear in our experiments, along with relevant information.
\begin{itemize}
    \item \textbf{Alpaca}~\citep{alpaca} contains 52k synthetic examples generated by explicitly giving the requirement instruction generation to the text-davinci-003 model. Although the created dataset is intended to be varied, a thorough examination reveals that it is heavily US-centric. It is also discovered that the original version has numerous issues that affect its quality and suitability for training a trustworthy language model. These issues includes hallucinations, merged instructions, empty outputs, impractical instructions like generating images, wrong answers, and non-sensical instructions, etc.
    \item \textbf{AlpaGasus-1k/9k}~\citep{chen2023alpagasus} contains 1k/9k high-quality examples filtered from the original Alpaca-52k dataset. It suggests implementing data selection by means of strong LLMs, such as ChatGPT, to automatically detect and filter out low-quality data. By doing this, they leave out problematic samples, which endanger the effectiveness of refined models.
    \item \textbf{Recycled-Alpaca}~\citep{li2023reflection} comprises of 52k enhanced examples based on Alpaca-52k. Given the initial basic dataset, a high-quality version of each data point is generated using an Oracle model, such as chatGPT. However, a typical issue with using LLMs as judges is the inability to produce different results. To address this potential issue, inspired by Chain-of-Thought prompting,  numerous particular criterias are proposed for the Oracle model to follow, and then strong target LMs respond to those precise requirements with critical responses. The responses to these criteria can then be used as bridges (chains of thought) to create new, satisfied instruction-response combinations.
    \item \textbf{LIMA}~\citep{zhou2023lima} collects a dataset of 1000 prompts and responses for training, with the outputs stylistically aligned but the inputs different. It also provides an open-source test set of 300 prompts and a development set of 50. Curated from multiple sources, LIMA is primarily divided among community Q\&A websites like Stack Exchange, wikiHow, and the Pushshift Reddit Dataset~\citep{baumgartner2020pushshift}, as well as manually created examples. In terms of Q\&A communities, frequently upvoted answers on Reddit are typically hilarious or trolling, requiring more manual effort to align responses that adhere to the proper style. In contrast, answers from Stack Exchange and wikiHow are well-aligned with the behavior of a helpful chat assistant. Human-authored examples are used to boost the diversity of dataset. 
    \item \textbf{Evol-Instruct (WizardLM)}~\citep{xu2023wizardlm} contains 70k training examples with varying complexity and 218 test instances. The training dataset is initially initialized using Alpaca's 52k instruction data. After iteratively completing $M=4$ evolutions, the dataset has 250k instructions. More specifically, for each instruction in each round of evolution, one evolving prompt from total six new prompts (i.e., five from in-depth evolving and one from in-breadth evolving) is selected with equal probability. Then, ChatGPT is used to produce answers for each instruction, yielding $52 \times 4 \times 3 = 624\text{k}$ instruction-response pairs. Finally, the Evol-Instruct dataset is created by picking a subset of 70k instructions. 218 test instructions are collected from diverse sources including online opensource projects, platforms, and forums. This test set is primarily a union of 29 distinct skills identified among real-world human instructions, such as Coding Generation \& Debugging, Reasoning, Math, Writing, Complex Formats, Extensive Disciplines, and so on.
    \item \textbf{Vicuna}~\citep{vicuna2023} divides 80 test instructions into 8 question categories, including Fermi problems, commonsense, roleplay scenarios, coding/math/writing tasks, counterfactual, knowledge, and generic, to evaluate various aspects of a chatbot's performance. Vicuna has been demonstrated to mostly include instructions of low difficulty and complexity~\citep{xu2023wizardlm}. 
    \item \textbf{Self-Instruct}~\citep{wang2022self} has 252 human-authored test instructions with 1 handcrafted output per instruction. Self-Instrction test set is created to better reflect the practical value of instruction-following models. The authors were motivated to curate instructions of different domains ranging from email writing and social media to productivity tools and programming. Authors also deliberately diversify the styles and formats of tasks, such as including instructions of different lengths and considering input/output that takes the form of bullet points, tables, codes, equations, etc.
    \item \textbf{Koala}~\citep{koala_blogpost_2023} consists of 180 real user queries that were posted on the Internet. These user-initiated queries cover a wide range of subjects, typically have a conversational tone, and are probably more indicative of the practical applications of chat-based systems. Queries with a BLEU score of more than 20\% with any example from our training set are filtered away in order to reduce the possibility of test-set leaking. Prompts pertaining to code and languages other than English are also excluded because the crowd workers, who make up the pool of raters, are unable to accurately examine the answers to these questions.

\end{itemize}

\subsection{Training hyperparameters}
\label{app:training_hparams}

This section lists the hyperparameters necessary for reproducing our work. Our experiments are built upon FastChat framework~\citep{zheng2023judging}. In particular, we follow the training configuration as reported in~\citet{alpaca} to fine-tune the base model on full IFT datasets like Alpaca-52k and Evol-Instruct-70k, while we refer to LIMA~\citep{zhou2023lima} and AlpaGasus~\citep{chen2023alpagasus} when fine-tuning the base model on IFT datasets with 1k and 9k training examples, respectively. 
In addition to existing experimental setups in prior work, we adopt the recently proposed NEFTune augmentation for our (\refined-)\alpaca-1k-\longest experiments. We have \texttt{neftune\_noise\_level} set to 5 for \llama-2-7B, while for Mistral-7B-v0.1 and \llama-2-13B it is set to 3. It should be noted that we use 4 $\times$ 40G A100 to finetune \llama-2-7B and 4 $\times$ 80G A100 to finetune Mistral-7B-v0.1 and \llama-2-13B. We present the detailed training hyperparameters in Table~\ref{tab:training_hparams}.

\begin{table*}[htbp]
    \caption{Details of training hyperparameters for all experiments.}
    \vspace{0.3em}
    \centering
    \resizebox{0.92\textwidth}{!}{
        \begin{tabular}{l|ccccccccc}
        \toprule
        Datasets & Data Size & \# GPUs & Epochs & LR & LR Scheduler & Batch Size & Context Win. Len. & WD & Warmup Rate \\
        \midrule
        \multicolumn{9}{l}{\it \textbf{\llama-2-7B}} \\
        \midrule
        \evol-70k & 70k & 4 & 3 & 2e-5 & Cosine & 128 & 512 & 0.0 & 0.3 \\
        \alpaca-52k & 52k & 4 & 3 & 2e-5 & Cosine & 128 & 512 & 0.0 & 0.3 \\
        \alpagasus-9k & 9k & 4 & 3 & 2e-5 & Cosine & 128 & 512 & 0.0 & 0.3 \\
        \alpaca-9k-\longest & 9k & 4 & 3 & 2e-5 & Cosine & 128 & 512 & 0.0 & 0.3 \\
        \alpagasus-1k & 1k & 4 & 15 & 1e-5 & Linear & 128 & 2048 & 0.1 & 0.0 \\
        \lima-1k & 1k & 4 & 15 & 1e-5 & Linear & 128 & 2048 & 0.1 & 0.0 \\
        \alpaca-1k-\longest & 1k & 4 & 15 & 1e-5 & Linear & 128 & 2048 & 0.1 & 0.0 \\
        \evol-\alpagasus-1k & 1k & 4 & 15 & 1e-5 & Linear & 128 & 2048 & 0.1 & 0.0 \\
        \evol-1k-\longest & 1k & 4 & 15 & 1e-5 & Linear & 128 & 2048 & 0.1 & 0.0 \\
        \midrule
        \multicolumn{9}{l}{\it \textbf{Mistral-7B-v0.1}} \\
        \midrule
        \alpaca-52k & 52k & 4 & 3 & 4e-6 & Cosine & 128 & 512 & 0.0 & 0.3 \\
        \alpagasus-1k & 1k & 4 & 15 & 2e-6 & Linear & 128 & 2048 & 0.1 & 0.0 \\
        \lima-1k & 1k & 4 & 15 & 2e-6 & Linear & 128 & 2048 & 0.1 & 0.0 \\
        \alpaca-1k-\longest & 1k & 4 & 15 & 2e-6 & Linear & 128 & 2048 & 0.1 & 0.0 \\
        \midrule
        \multicolumn{9}{l}{\it \textbf{\llama-2-13B}} \\
        \midrule
        \alpaca-52k & 52k & 4 & 5 & 1e-5 & Cosine & 128 & 512 & 0.0 & 0.3 \\
        \alpagasus-1k & 1k & 4 & 15 & 1e-5 & Linear & 128 & 2048 & 0.1 & 0.0 \\
        \lima-1k & 1k & 4 & 15 & 1e-5 & Linear & 128 & 2048 & 0.1 & 0.0 \\
        \alpaca-1k-\longest & 1k & 4 & 15 & 1e-5 & Linear & 128 & 2048 & 0.1 & 0.0 \\
        \bottomrule
        \end{tabular}
    }
    \label{tab:training_hparams}
\end{table*}

\subsection{Evaluation details} \label{app:evaluation_details}

\paragraph{Evaluation metrics for head-to-head comparisons. } Since automated evaluation based on powerful LLMs offers superior scalability, explainability and reproducibility than human evaluation, we apply an LLM that has high human preference agreement as the judge to evaluate the target model (e.g., \llama-2-7B fine-tuned on \alpaca-1k-\longest) and compare it with a baseline model (e.g., GPT-4-Turbo). We append both models' outputs in the input instruction to the LLM judge, followed by a request to the judge which prompts the model to rate the responses with a score between 1 and 10. Since there exists position bias within LLM-based automated evaluation~\citep{zheng2023judging}, we run evaluation on both orders (i.e., placing the response of the target model before/after the baseline model's response) and calculate the win rate (tie is allowed).

\paragraph{LLM-as-a-judge. } %
Given their good agreement with human evaluators shown in LLMBar~\citep{zeng2023evaluating}, we decide to adopt GPT-4 (i.e., GPT-4-0613) and PaLM2 (i.e., text-bison@002) as the LLM judges to appropriately assess the instruction-following performance of instruction-tuned models.

\begin{figure} \centering
\small
\includegraphics[width=0.8\textwidth]{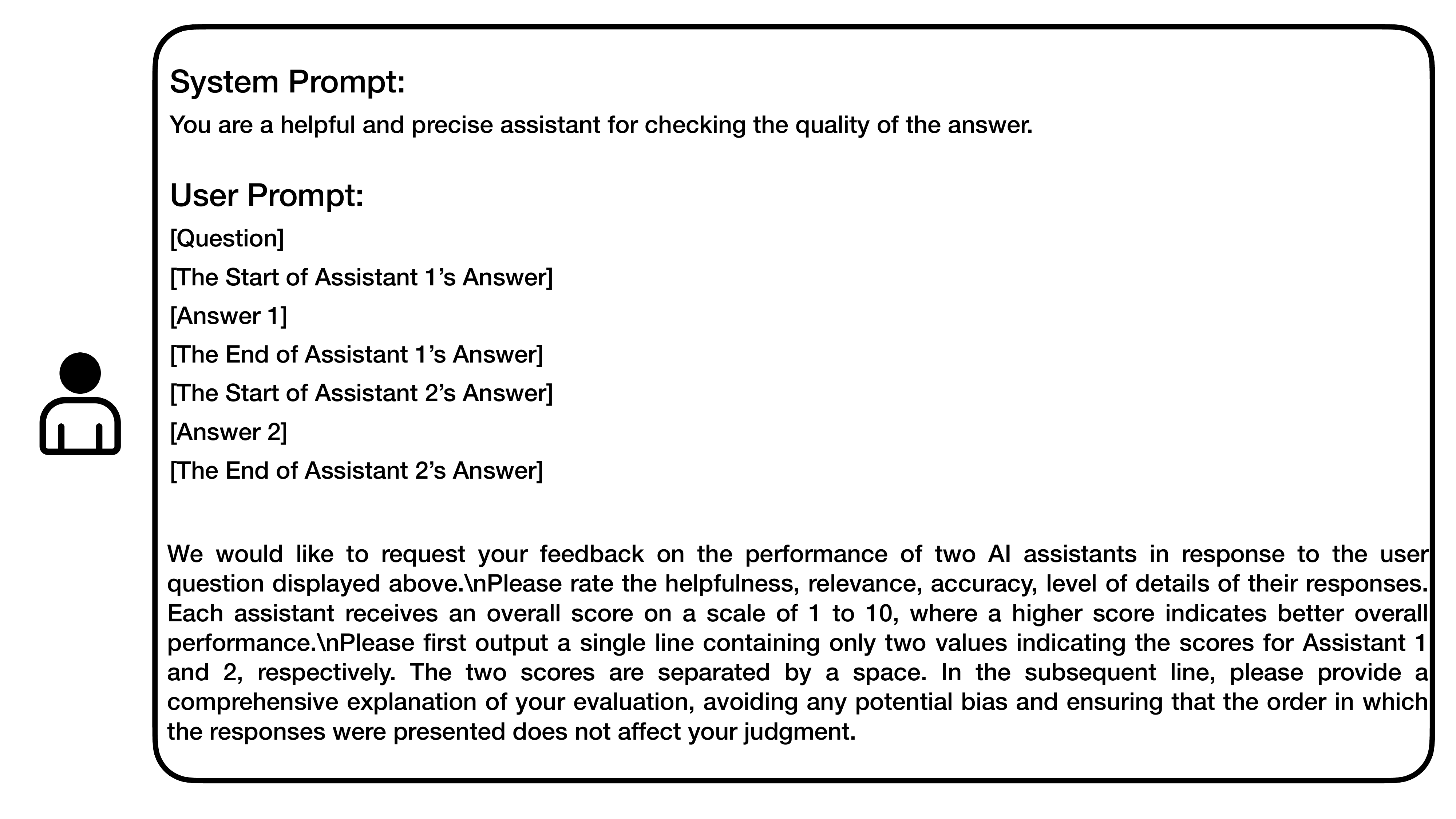}
\vspace{-2mm}
\caption{The prompt template for evaluation using GPT-4 or PaLM2 as the judge.} \label{fig:llm_evaluation_prompt}
\end{figure}

\paragraph{Evaluation prompt for GPT4- and PaLM2-as-a-judge.} We adopt the same evaluation prompt for both GPT4- and PaLM2-as-a-judge as what AlpaGasus~\citep{chen2023alpagasus} uses, which is also the prompt for evaluation used in the original Vicuna work~\citep{vicuna2023}. We provide the detailed form of the prompt in Fig.~\ref{fig:llm_evaluation_prompt}.

\begin{figure}
\centering
\small
\subfigure[\small The rule of the user study]{
    \includegraphics[width=1.\textwidth]{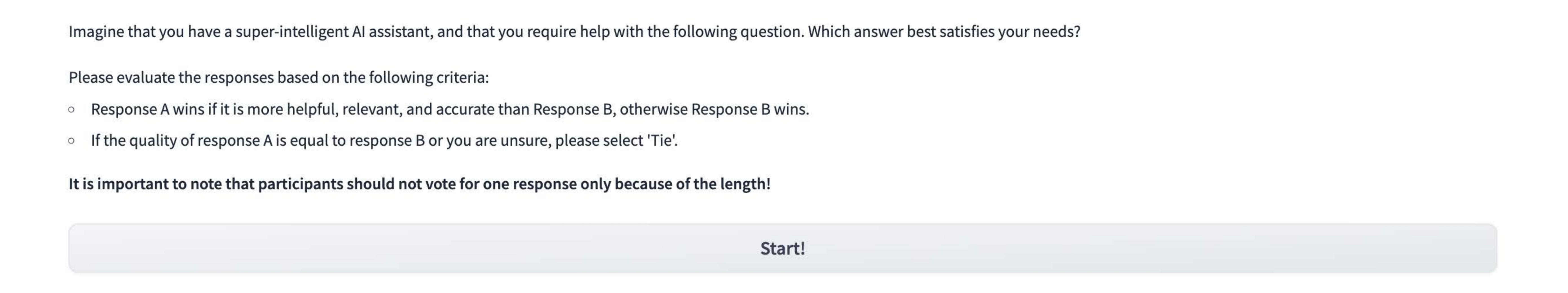}
    \label{fig:user_study_rule}
}

\subfigure[\small An example of the user study]{
    \includegraphics[width=1.\textwidth]{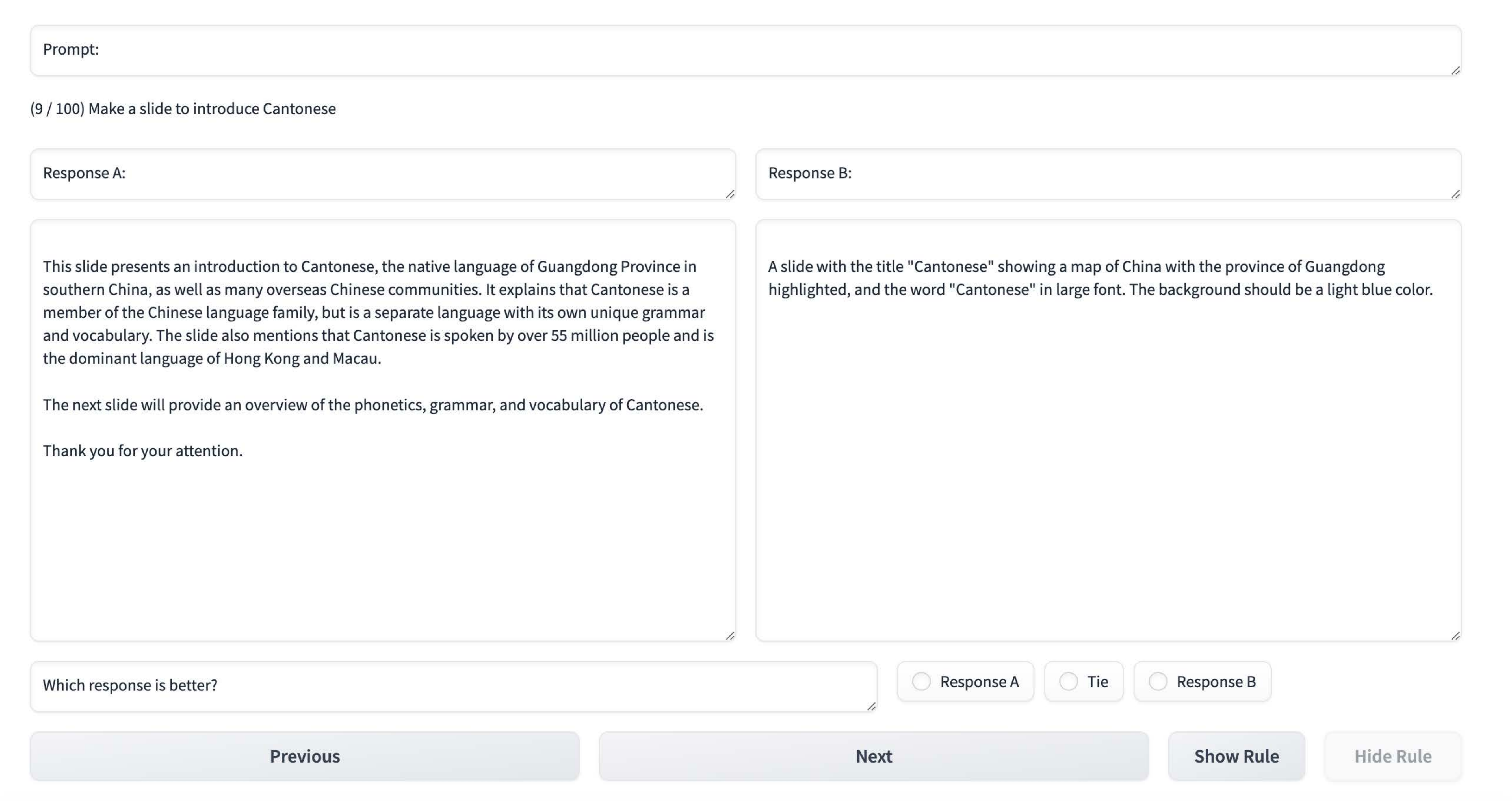}
    \label{fig:user_study_question}
}

\vspace{-2mm}
\caption{\textbf{The user study template of human evaluation.} Our user study consists of 100 questions, uniformly sampled from 5 test sets: LIMA, Vicuna, Koala, WizardLM, and Self-Instruct. We deployed this user study online using Gradio.}
\label{fig:user_study}
\end{figure}

\paragraph{Human evaluation. } Model-based head-to-head evaluation is notorious for its implicit preference for more verbose and engaging answers. Thus we conduct a human-based evaluation, in which the rule emphasizes that human annotators should choose the better answer from two candidates based solely on relevancy and helpfulness, and ignore potential superficial features, such as an engaging tone and response length. In particular, we sample 100 random instructions from evaluation datasets we use in model-based head-to-head comparisons and generate the responses for the baseline model, \llama-2-7B-\alpaca-52k, and the target model, \llama-2-7B-\alpaca-1k-\longest. To improve the efficiency of human annotation and reach more annotators, we designed a demo for the user study (see the full template in Fig.~\ref{fig:user_study}), building with \textit{Gradio}~\cite{abid2019gradio}, an interactive design tool.

\paragraph{AlpacaEval 2.0. } We apply the AlpacaEval 2.0 benchmark in our experiments since it provides transferable comparisons, which is impossible to achieve in head-to-head evaluation. AlpacaEval 2.0 provides 805 test instructions, on which we generate new responses using the target model, and then calculate the score by competing with the baseline model (i.e., GPT-4-Turbo) judged by a designated automatic evaluator.

\paragraph{MT-Bench. } The test dataset of this benchmark~\cite{zheng2023judging} covers 8 common categories of user prompts: coding, math, reasoning, extraction, roleplay, writing, humanities/social science, and STEM. It contains 80 questions, all of which are high-quality and challenging, designed to assess models' abilities to engage in multi-turn conversations and follow instructions. 

\paragraph{Open LLM Leaderboard. } Several multiclass classification datasets %
are used to compute the models ranking: ARC~\citep{clark2018think}, MMLU~\citep{hendrycks2020measuring}, TruthfulQA~\citep{lin2022truthfulqa}, Winogrande~\citep{sakaguchi2021winogrande}, HellaSwag~\citep{zellers2019hellaswag}. The combination of datasets widely measures an LLM's capacity to react to factual queries and reasoning challenges, and we use this benchmark to compare the model's change in factual capabilities before and after instruction fine-tuning. 

\begin{figure}[htbp]
    \centering
    \subfigure[\small Head-to-head comparisons (in \%) with different LLM-judges]{
        \includegraphics[width=0.7\textwidth]{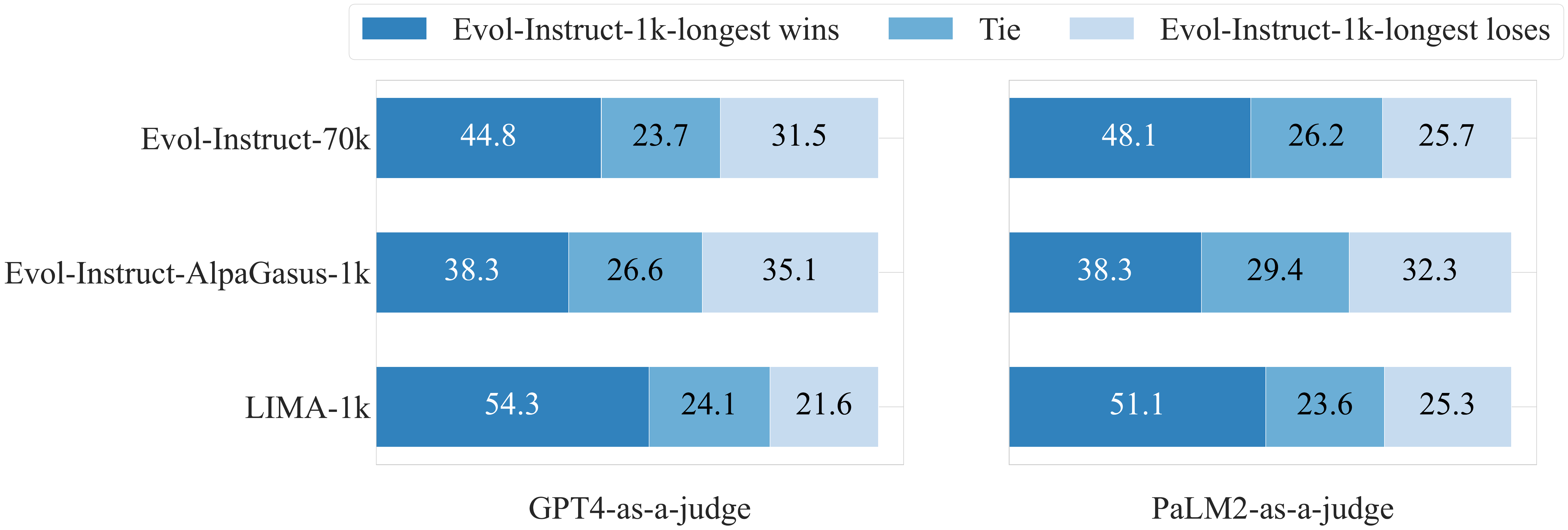}
        \label{fig:llama_2_7b_evol_instruct_multiple_judges}
    }
    \subfigure[\small Avg. number of tokens in responses]{
        \includegraphics[width=.7\textwidth]{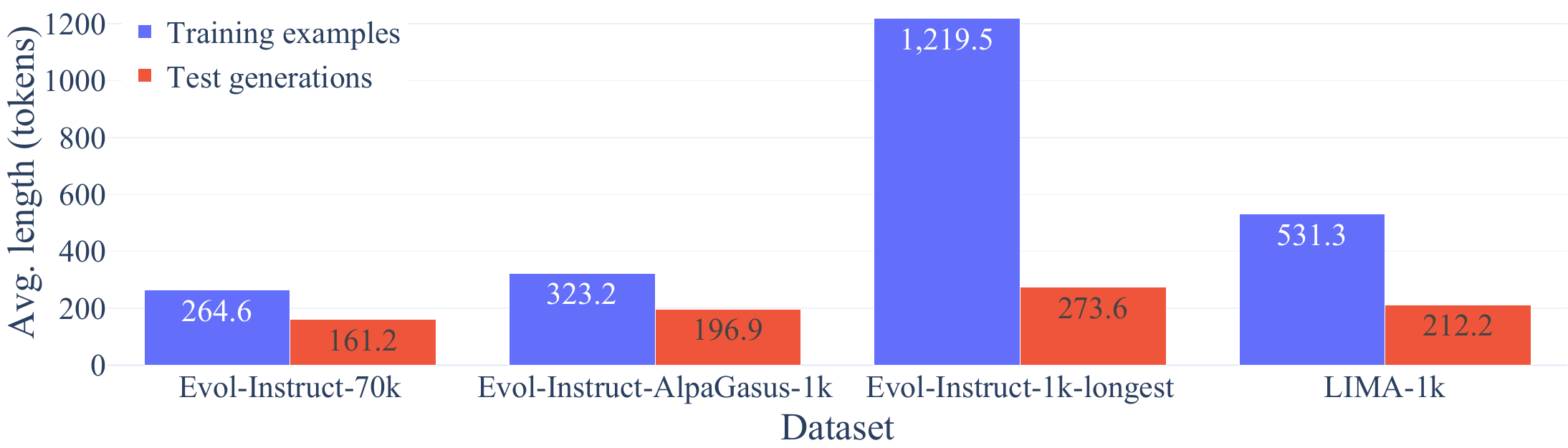}
        \label{fig:len_dist_llama_2_7b_evol_instruct}
    }
    
    \caption{\textbf{Effect of using long instructions from \evol-70.} We fine-tune LLaMA-2-7B models on \evol-70k~\citep{xu2023wizardlm}, \evol-AlpaGasus-1k, LIMA-1k and our \evol-1k-\longest.
    \textbf{(a)} \evol-1k-longest beats three baselines in instruction-following performance according to both GPT-4 and \palm as judges.
    \textbf{(b)} \evol-1k-\longest leads to the largest average response length at test time. Interestingly, the average length of training responses for \evol-1k-\longest is more than twice as long as that of \lima-1k, but the average length of \evol-1k-\longest at test time only increases by 28.9\%. }
    \label{fig:evol_instruct_average_pairwise_performance}
\end{figure}

\begin{figure} \centering
\small
\includegraphics[width=0.60\textwidth]{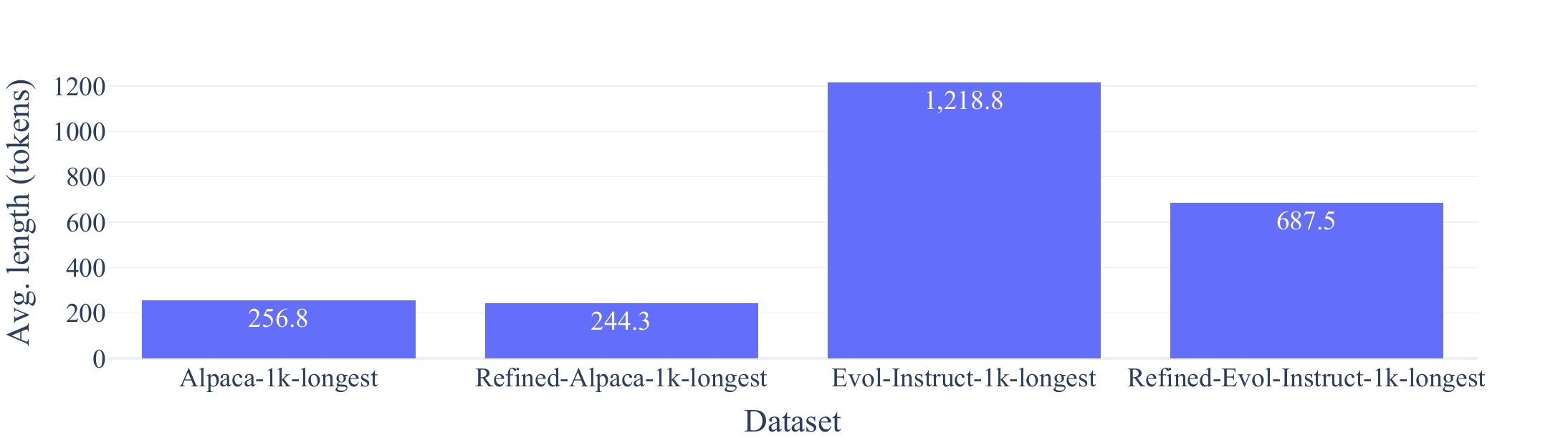}
\caption{\textbf{The average response lengths (as number of tokens) of training examples.} We show the average response lengths for different datasets before and after the instruction refinement step.}
\label{fig:train_data_len_distribution}
\end{figure}

\section{Additional results}

\subsection{Scores of \alpaca-1k-longest according to GPT-3.5-Turbo} \label{app:scores}

\begin{figure} \centering
\small
\includegraphics[width=0.60\textwidth]{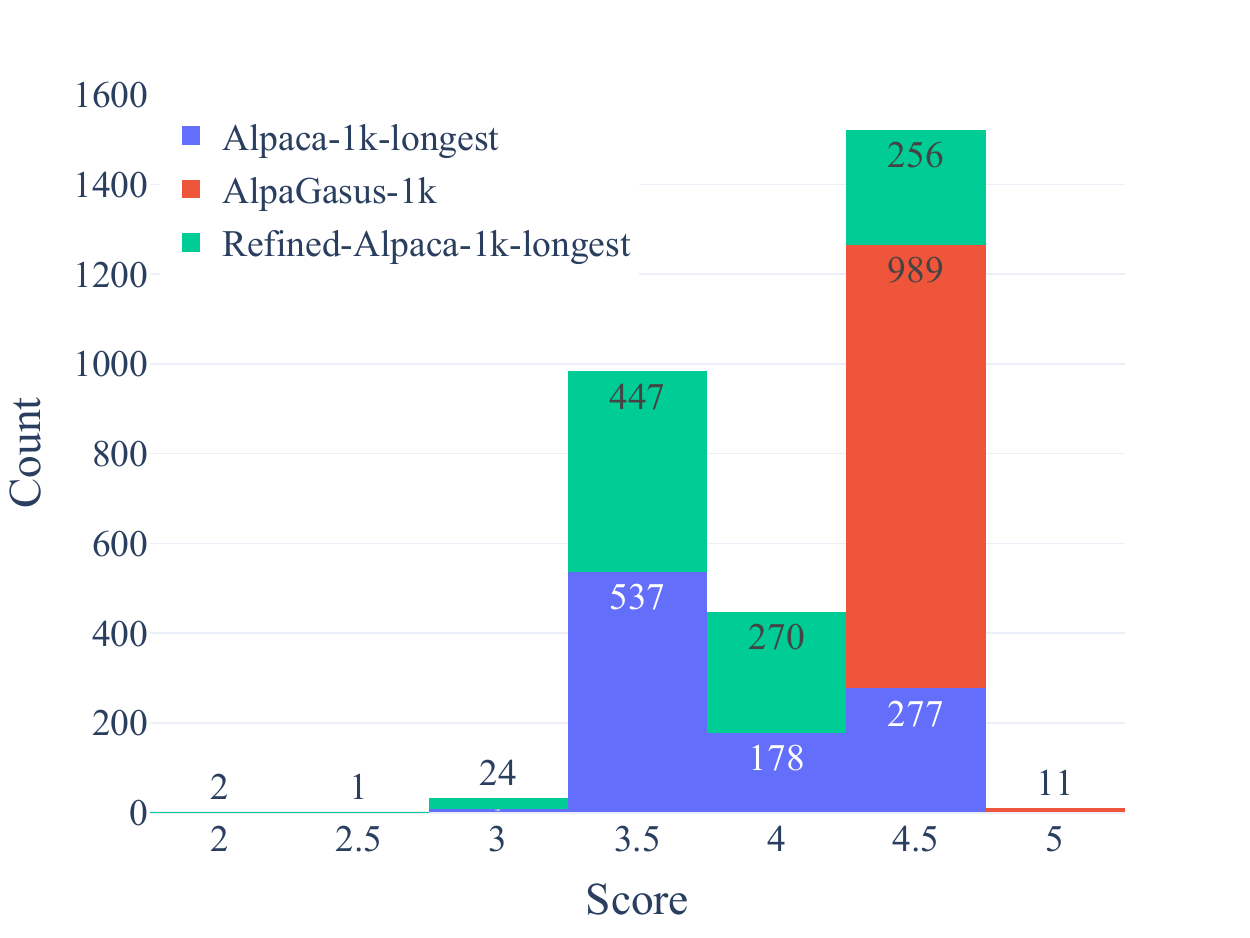}
\caption{\textbf{Quality of training examples.} We show the distribution of the scores, as measured by GPT-3.5-Turbo, of the \alpagasus-1k, \alpaca-1k-\longest, and \refined-\alpaca-1k-\longest datasets (scale of scores is 1-5).} \label{fig:scores_distribution}
\end{figure}

In Fig.~\ref{fig:scores_distribution} we show the score distribution from \citet{chen2023alpagasus} for the 1k longest examples compared to those of \alpagasus-1k (i.e. that highest scoring ones): we see that the overlap between the two datasets is minimal, and most of the longest examples have score of 3.5.
Interestingly, this suggests that GPT-3.5-Turbo prefers longer responses when used as a judge, e.g. in the AlpacaEval 2.0 benchmark, while favors different features when asked to score the quality of the instruction-response pairs in \alpaca. %

\subsection{\palm-as-a-judge details} \label{app:palm_evaluation}

\begin{figure*}[p]
    \centering
    \subfigure[\small Alpaca-1k-longest vs. LIMA-1k]{
        \includegraphics[width=0.31\textwidth]{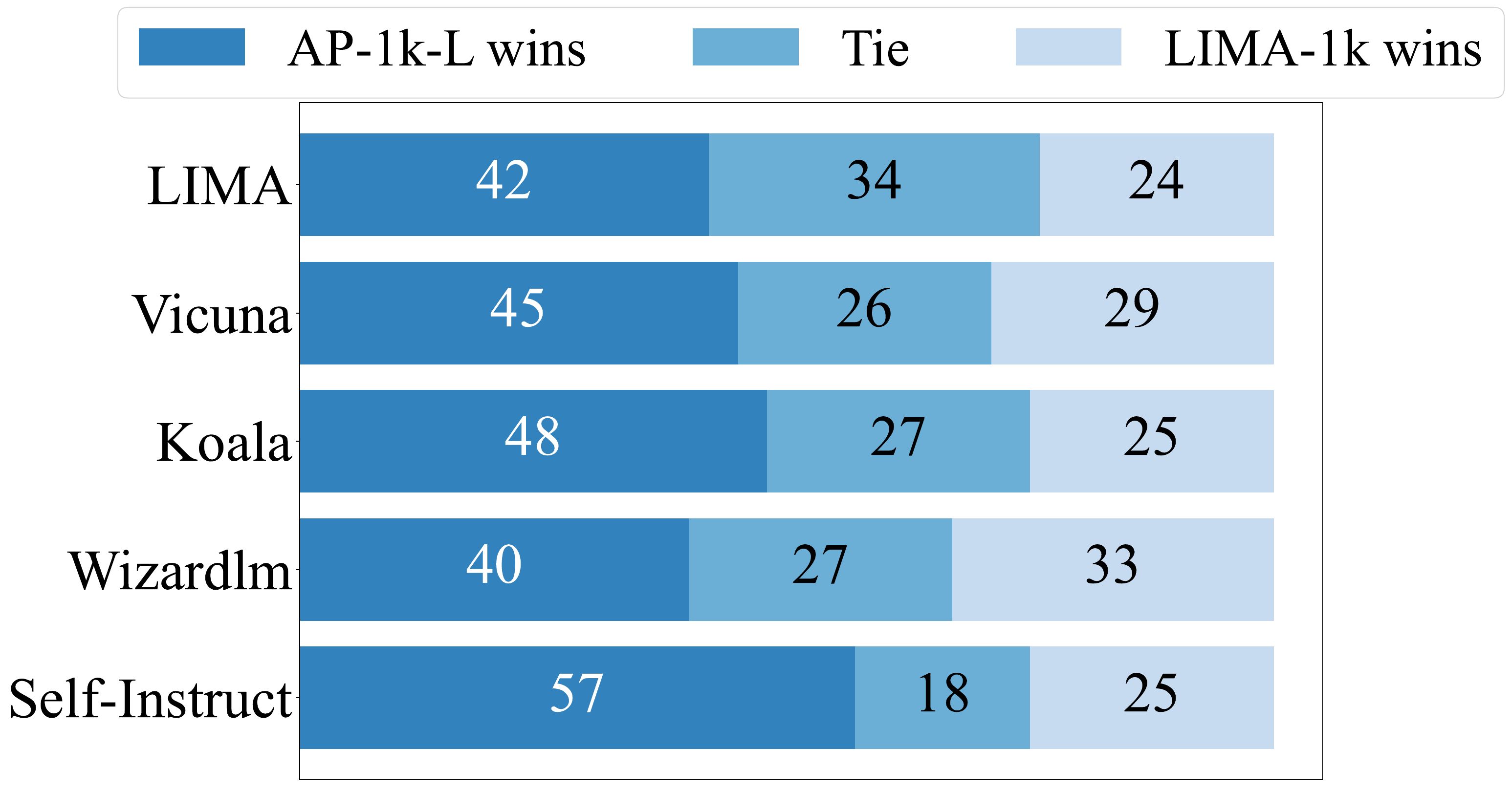}
        \label{fig:llama_2_alpaca_1k_longest_vs_lima_palm2}
    }
    \hfill
    \subfigure[\small Alpaca-1k-longest vs. AlpaGasus-1k]{
        \includegraphics[width=0.31\textwidth]{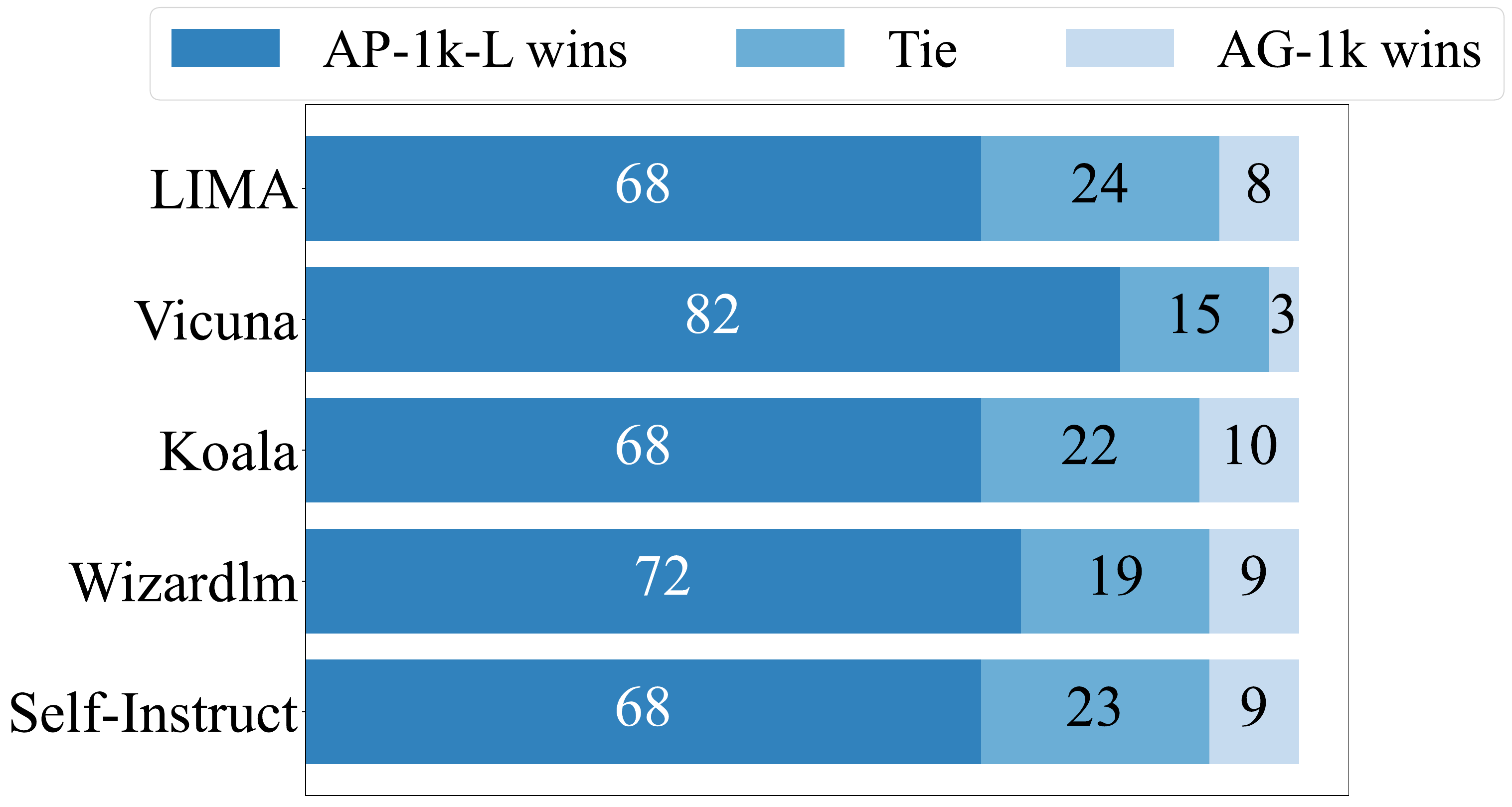}
        \label{fig:llama_2_alpaca_1k_longest_vs_alpagasus_1k_palm2}
    }
    \hfill
    \subfigure[\small Alpaca-1k-longest vs. Alpaca-52k]{
        \includegraphics[width=0.31\textwidth]{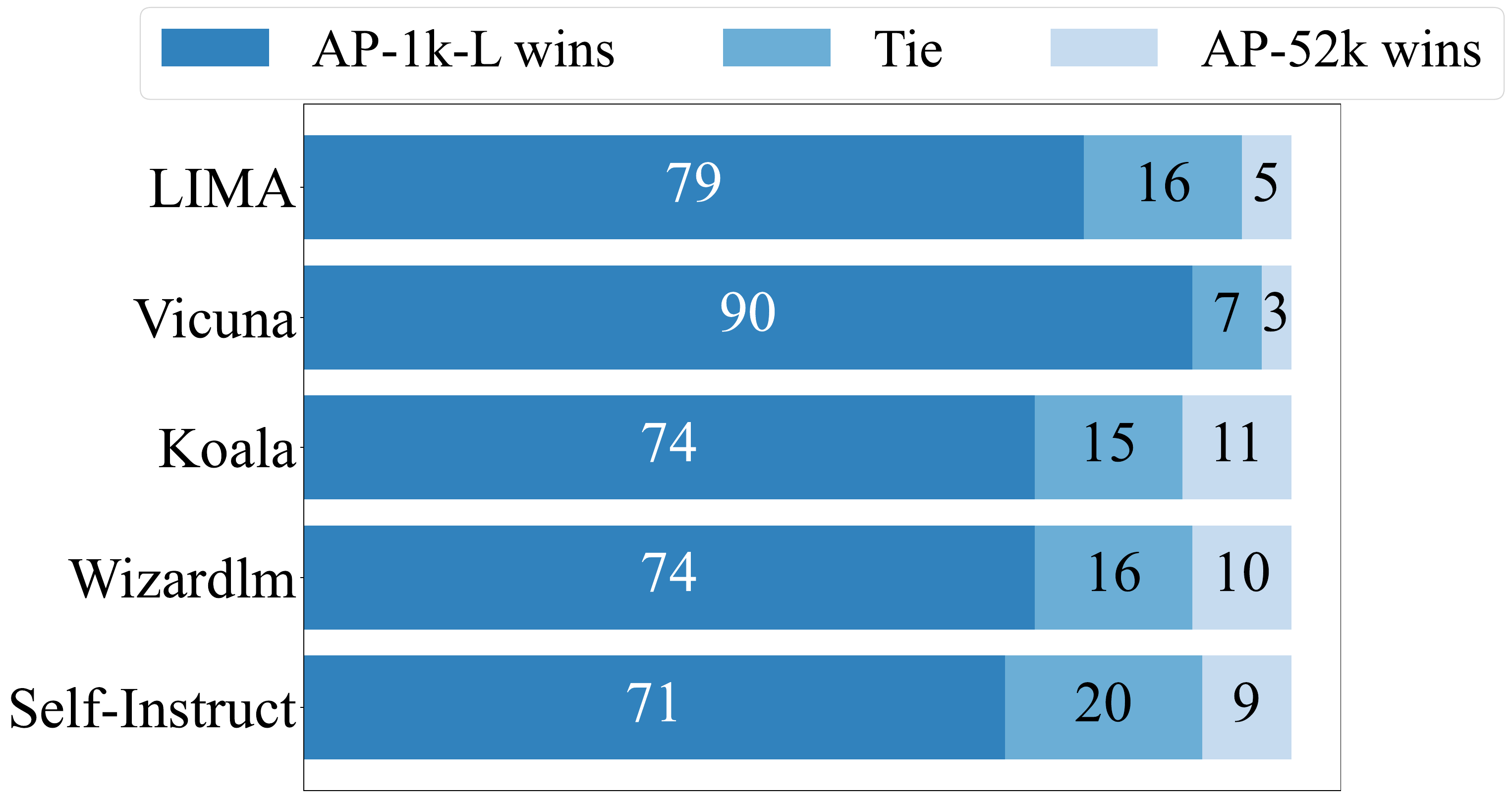}
        \label{fig:llama_2_alpaca_1k_longest_vs_alpaca_52k_palm2}
    }
    \hfill
    \subfigure[\small Evol-Instruct-longest vs. LIMA-1k]{
        \includegraphics[width=0.31\textwidth]{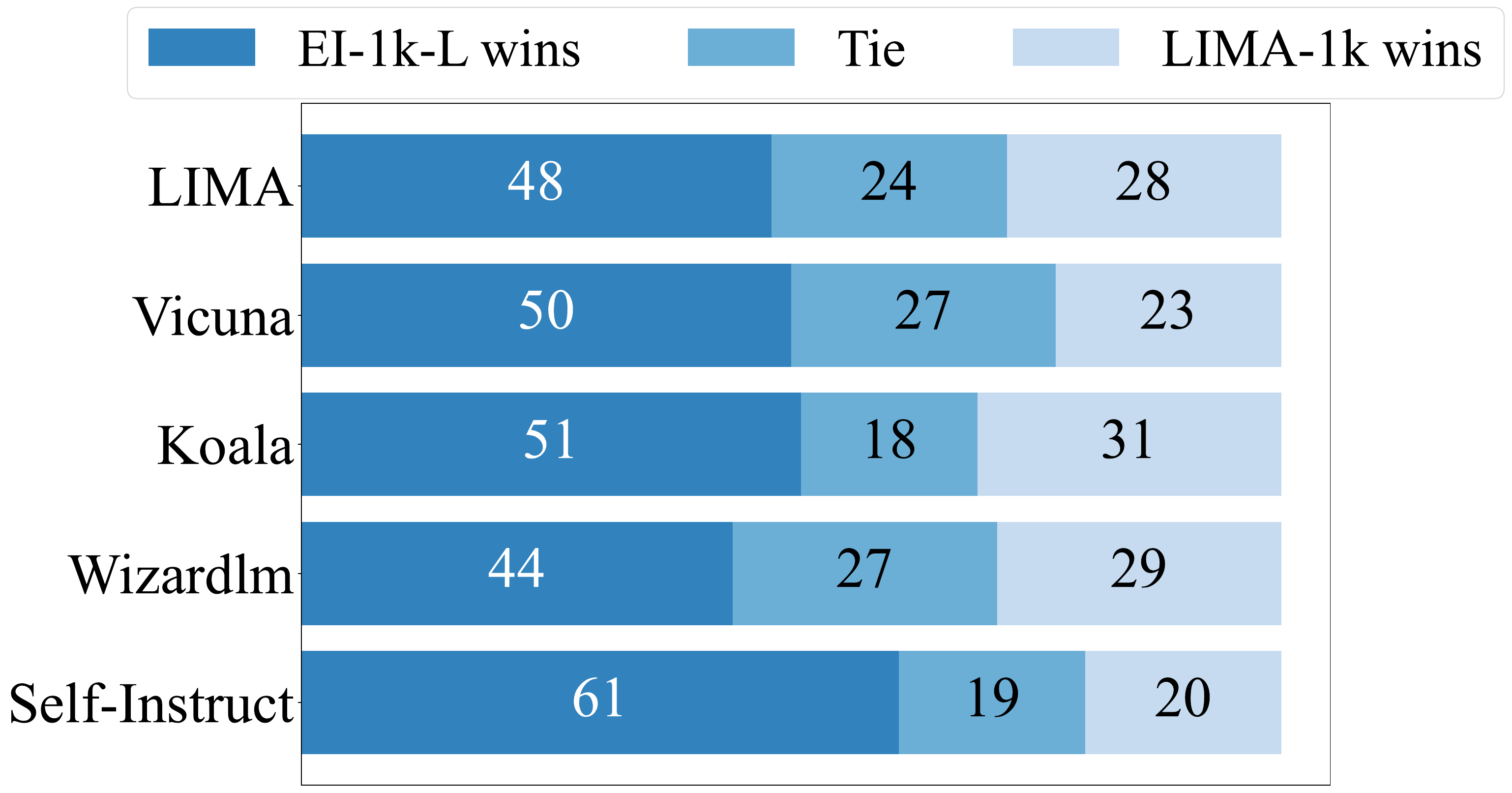}
        \label{fig:llama_2_evol_instruct_1k_longest_vs_lima_palm2}
    }
    \hfill
    \subfigure[\small EI-1k-longest vs. EI-AlpaGasus-1k]{
        \includegraphics[width=0.31\textwidth]{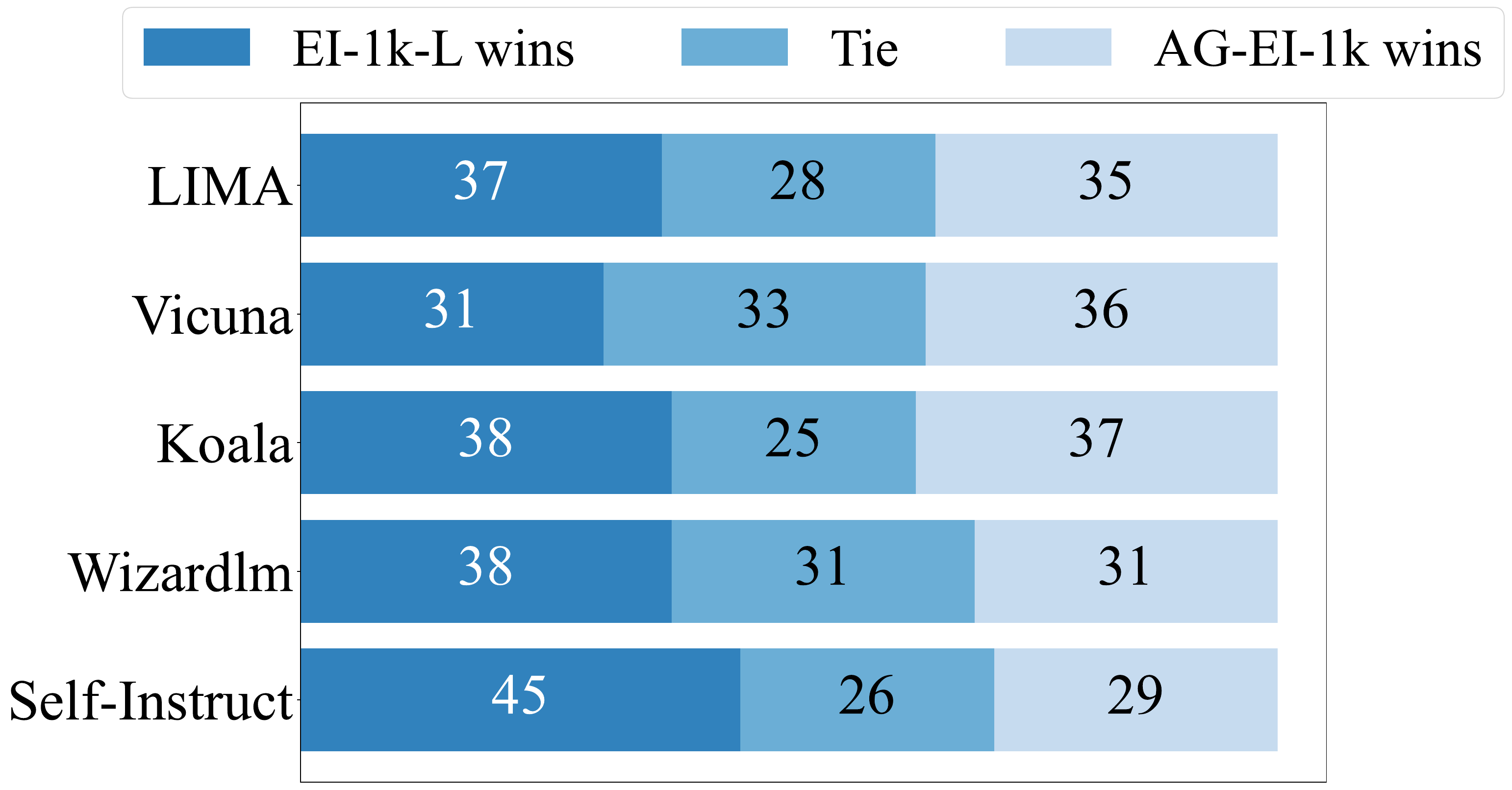}
        \label{fig:llama_2_evol_instruct_1k_longest_vs_alpagasus_evol_instruct_1k_palm2}
    }
    \hfill
    \subfigure[\small EI-1k-longest vs. Evol-Instruct-70k]{
        \includegraphics[width=0.31\textwidth]{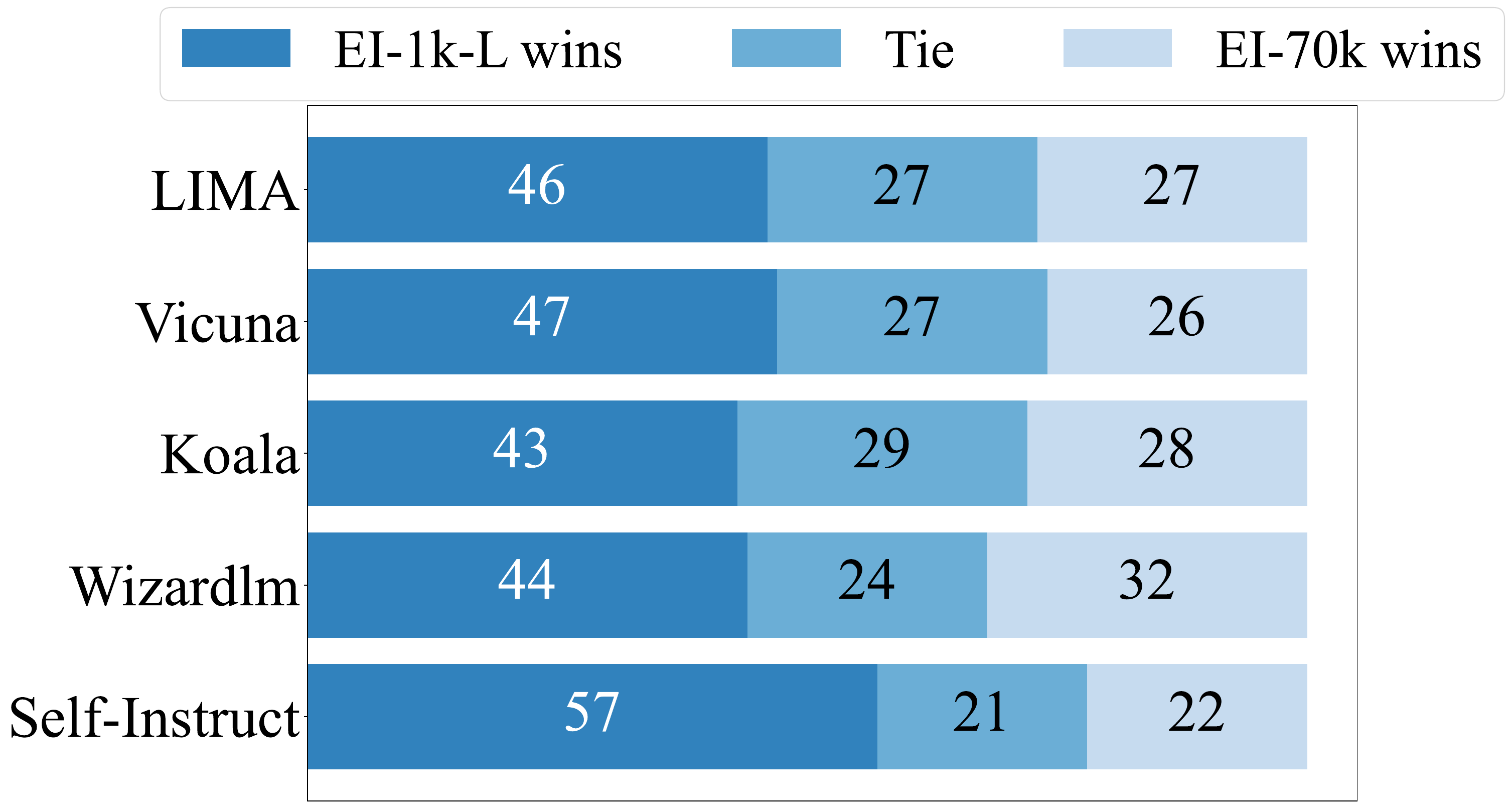}
        \label{fig:llama_2_evol_instruct_1k_longest_vs_evol_instruct_52k_palm2}
    }
    \hfill
    \caption{\textbf{Detailed preference evaluation (in \%, with \palm-as-a-judge).}
    For each pair of LLMs we report the win rate on 5 datasets (LIMA, Vicuna, Koala, WizardLM, Self-Instruct) according to \palm-as-a-judge. 
    \textbf{Top:} we compare fine-tuning on \alpaca-1k-\longest (AP-1k-L) to \alpaca-52k, \alpagasus-1k, and \lima-1k.
    \textbf{Bottom:} we compare fine-tuning on \evol-1k-\longest (EI-1k-L) to \evol-70k, \evol-\alpagasus-1k (i.e. using the method of \citet{chen2023alpagasus} to subsample \evol-70k), and \lima-1k.
    Our datasets of long responses consistently lead to higher preferences (higher win rate) on average than the existing methods. }
    \label{fig:preference_eval_on_alpaca_evol_palm2}
    
\end{figure*}

\begin{figure*}[p]
    \centering
    \subfigure{
        \includegraphics[width=\textwidth]{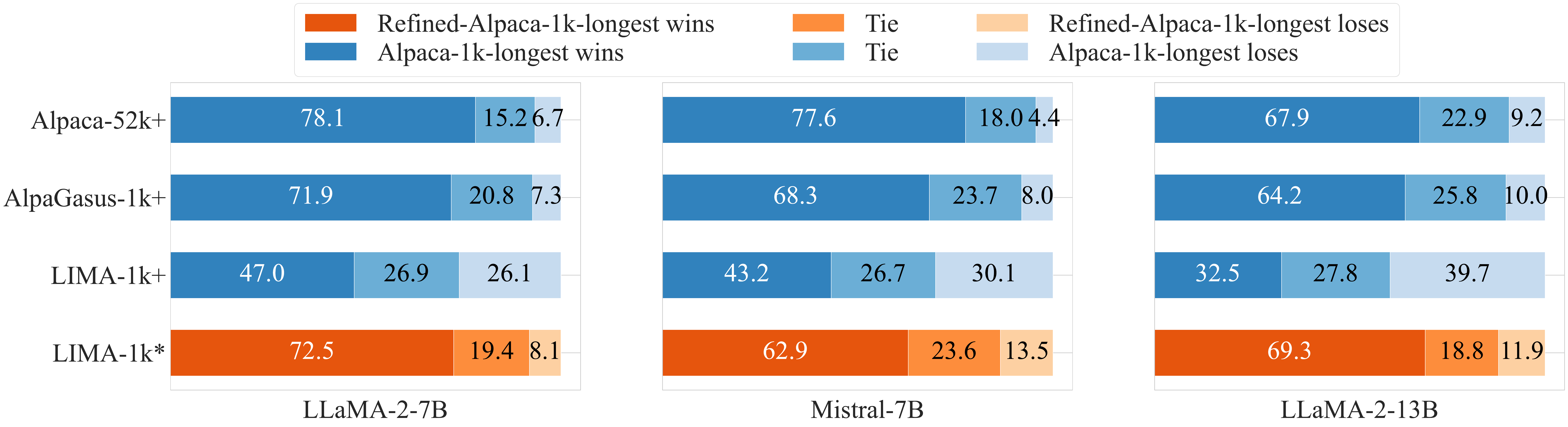}
        
    }

    \vspace{-2mm}
    \caption{\textbf{Refinement via introspection improves instruction-following performance across architectures (\palm-as-a-judge).} We report the average preference performance (\%) across 5 evaluation sets.
    We show win rate of models with different architectures fine-tuned on \alpaca-1k-\longest against \alpaca-52k, \alpagasus-1k and \lima-1k in blue (+ symbol). Additionally we illustrate the improvement brought by our \refined-\alpaca-1k-\longest over \lima-1k, the strongest baseline, in red (* symbol).
    }
    \label{fig:avg_win_rate_refined_palm2}
\end{figure*}

We present detailed preference evaluation results using PaLM2-as-a-judge on an array of \llama-2-7B-based models in Fig.~\ref{fig:preference_eval_on_alpaca_evol_palm2}. Moreover, we show the improvement given by the refined dataset in Fig.~\ref{fig:avg_win_rate_refined_palm2}. In both cases the observations are consistent with what obtained with GPT-4 as judge (see Fig.~\ref{fig:preference_eval_alpaca_evol} and Fig.~\ref{fig:avg_win_rate_refined} respectively).

\subsection{Preference evaluation on Mistral-7B-v0.1 and LLaMA-2-13B}
\label{app:mistral_7b_llama2_13b}

\begin{figure}[p]
    \centering
    \subfigure[\small Head-to-head comparisons with different LLM-judges]{
        \includegraphics[width=0.47\textwidth]{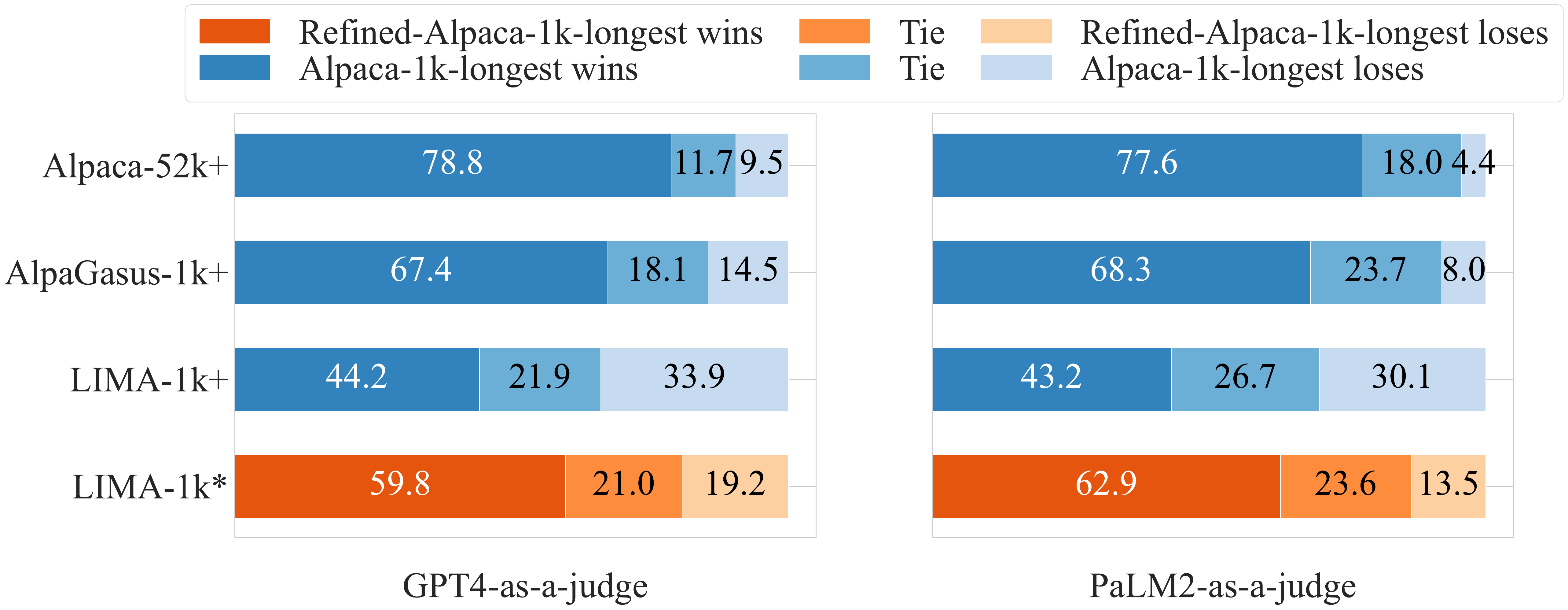}
        \label{fig:mistral_7b_alpaca_avg_win_rate_multiple_judges}
    }
    \subfigure[Avg. number of tokens in responses]{
        \includegraphics[width=0.46\textwidth]{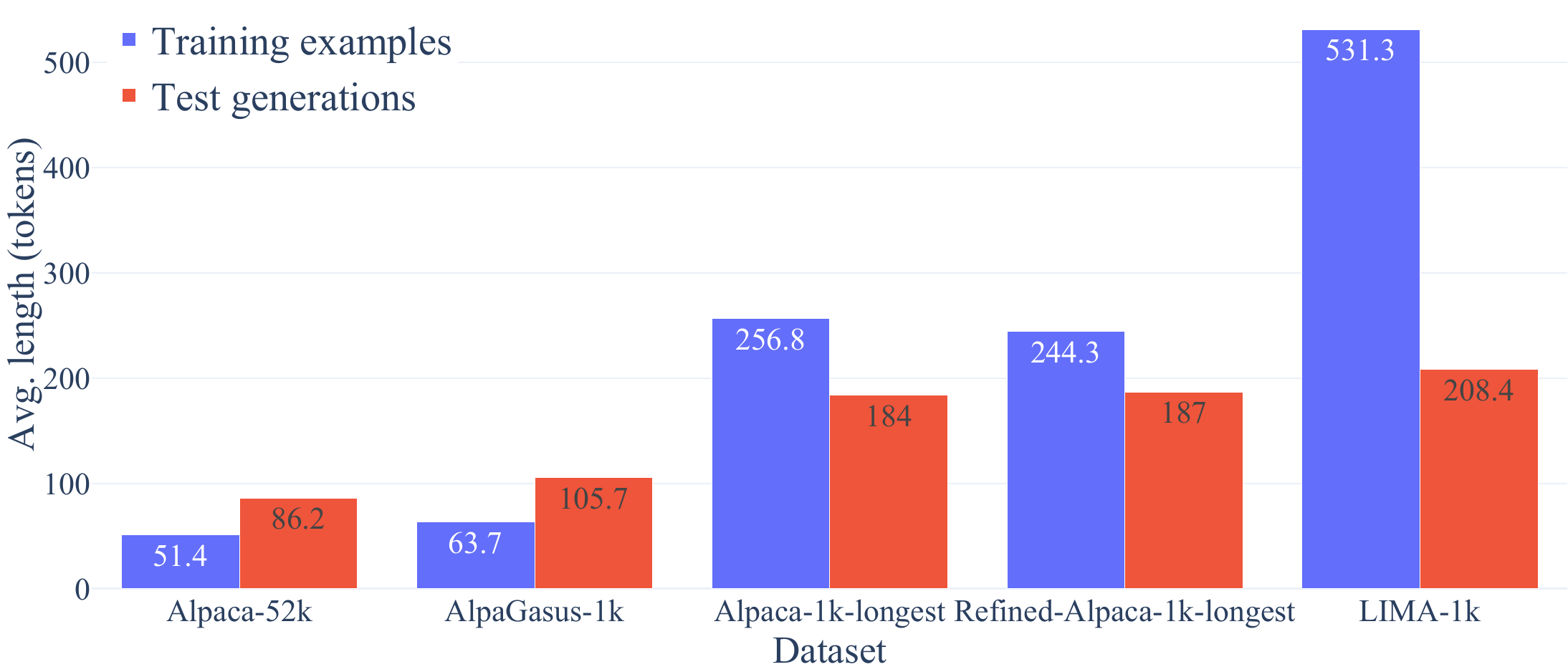}
        \label{fig:len_distribution_mistral_7b}
    }
    \caption{\textbf{Selecting the longest responses leads to a strong IFT dataset (Mistral-7B-v0.1).} We fine-tune Mistral-7B-v0.1 models on Alpaca-52k~\citep{alpaca}, AlpaGasus-1k~\citep{chen2023alpagasus}, LIMA-1k~\citep{zhou2023lima} and our \alpaca-1k-\longest. We show win rate of models with different architectures fine-tuned on \alpaca-1k-\longest against \alpaca-52k, \alpagasus-1k and \lima-1k in blue (+ symbol). Additionally we illustrate the improvement brought by our \refined-\alpaca-1k-\longest over \lima-1k, the strongest baseline, in red (* symbol).
    \textbf{(a)} Alpaca-1k-longest beats three baselines in instruction-following performance according to both GPT-4 and \palm as judges. And \refined-\alpaca-1k-\longest further enhance the instruction fine-tuning performance as demonstrated by larger win-rates given by both LLM-judges.
    \textbf{(b)} \alpaca-1k-\longest leads to an average response length at test time higher than \alpaca-52k and \alpagasus-1k, but smaller than \lima-1k, which demonstrates the LLM-judges' preference on our models is induced by better response quality instead of length bias.
    }
    \label{fig:avg_win_rate_mistral_models}
\end{figure}

\begin{figure}[p]
    \centering
    \subfigure[\small Head-to-head comparisons with different LLM-judges]{
        \includegraphics[width=0.47\textwidth]{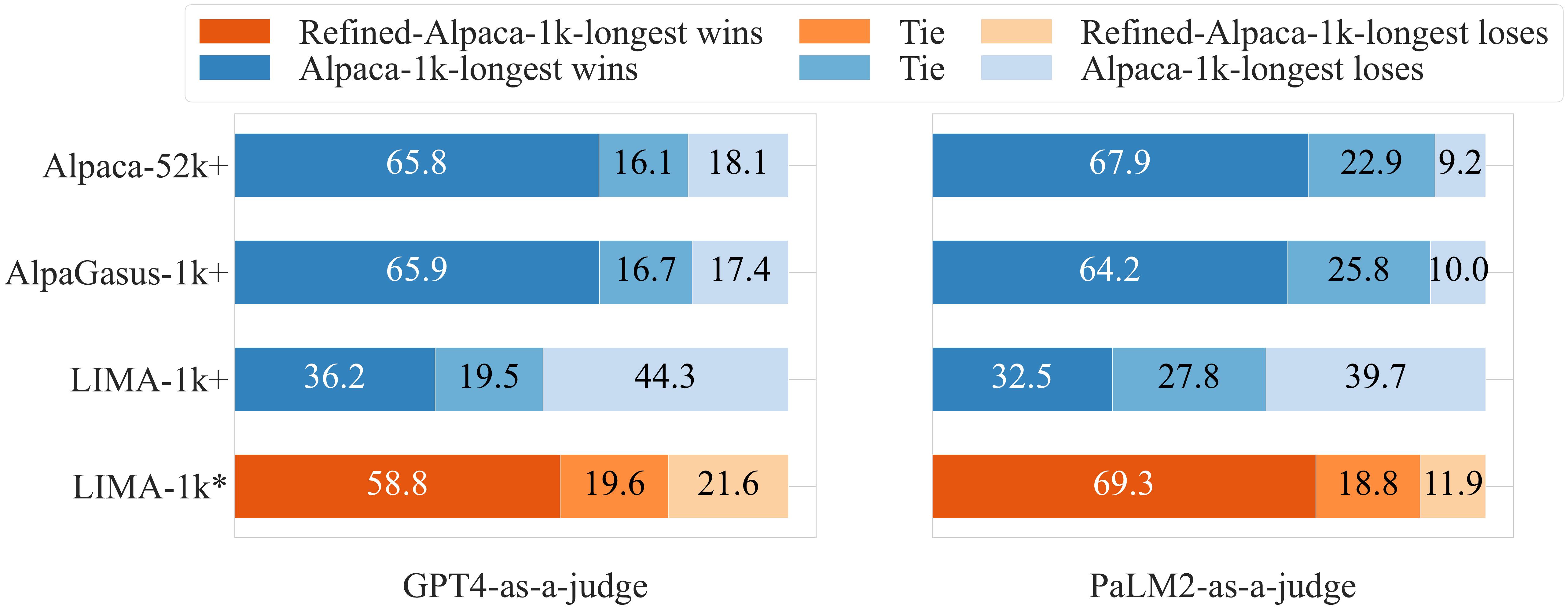}
        \label{fig:llama2_13b_alpaca_avg_win_rate_multiple_judges}
    }
    \subfigure[\small Avg. number of tokens in responses]{
        \includegraphics[width=0.46\textwidth]{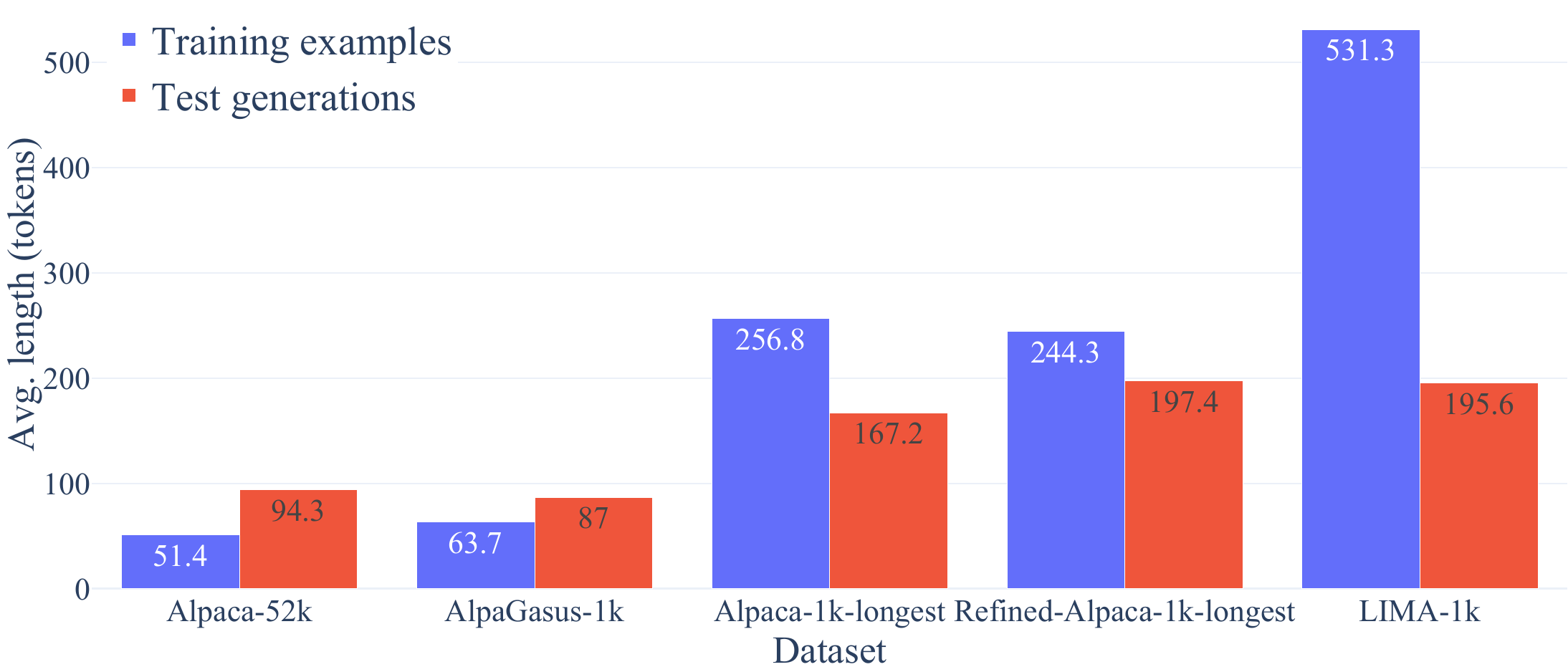}
        \label{fig:len_distribution_llama_2_13b}
    }
    \caption{\textbf{Selecting the longest responses leads to a strong IFT dataset (\llama-2-13B).} We fine-tune \llama-2-13B models on Alpaca-52k~\citep{alpaca}, AlpaGasus-1k~\citep{chen2023alpagasus}, LIMA-1k~\citep{zhou2023lima} and our \alpaca-1k-\longest. We show win rate of models with different architectures fine-tuned on \alpaca-1k-\longest against \alpaca-52k, \alpagasus-1k and \lima-1k in blue (+ symbol). Additionally we illustrate the improvement brought by our \refined-\alpaca-1k-\longest over \lima-1k, the strongest baseline, in red (* symbol).
    \textbf{(a)} Alpaca-1k-longest beats \alpaca-52k and \alpagasus-1k in instruction-following performance according to both GPT-4 and \palm as judges, but underperforms \lima-1k. However, \refined-\alpaca-1k-\longest significantly enhance the instruction fine-tuning performance of the model, surpassing \lima-1k. 
    \textbf{(b)} \alpaca-1k-\longest leads to an average response length at test time higher than \alpaca-52k and \alpagasus-1k, but smaller than \lima-1k. And the average response length of \refined-\alpaca-1k\longest at test time is comparable to that of \lima-1k, which demonstrates the LLM-judges' preference on \refined-\alpaca-1k-\longest is induced by better response quality instead of length bias.}
    \label{fig:avg_win_rate_llama_2_13b_models}
\end{figure}

This section contains the average preference evaluation results on Mistral-7B-v0.1 model and \llama-2-13B model over 5 evaluation sets (i.e., \lima, Vicuna, Koala, WizardLM, and Self-Instruct) as shown in Fig.~\ref{fig:avg_win_rate_mistral_models} and Fig.~\ref{fig:avg_win_rate_llama_2_13b_models}. 

\subsection{Open LLM results on Mistral-7B-v0.1, LLaMA-2-13B, \evol-70k}
\label{app:mistral_7b_llama2_13b_openllm}

\begin{figure*}[h!]
    \centering
    \begin{tabular}{c}

    Base model: \textbf{Mistral-7B-v0.1}\\
    \includegraphics[width=0.95\linewidth]{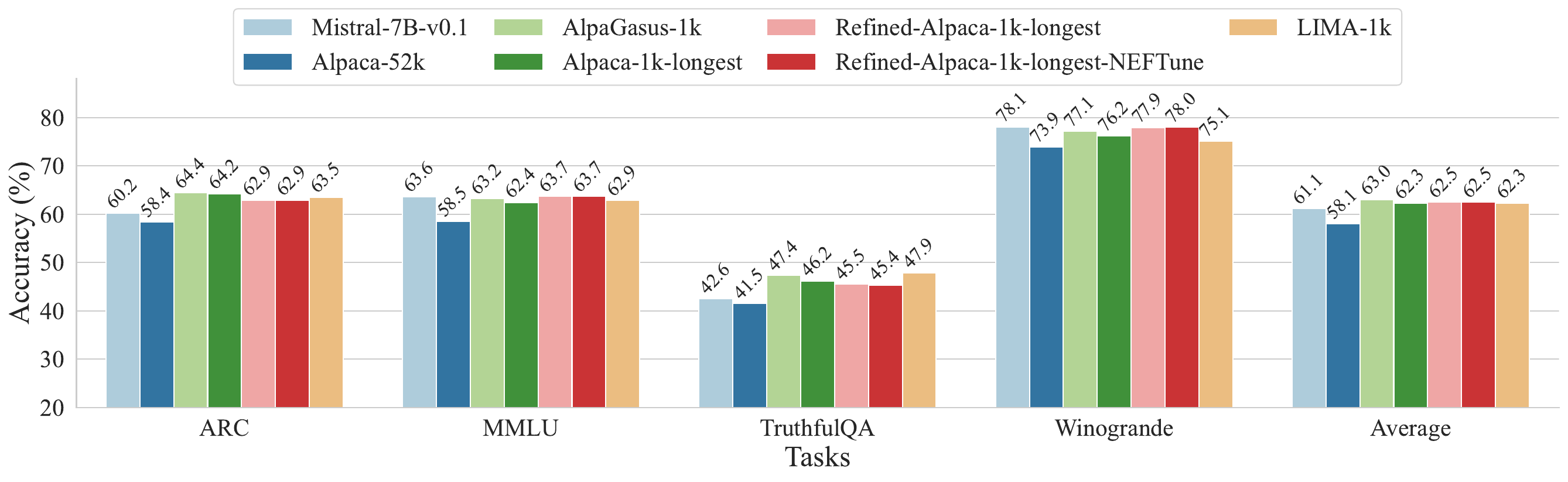}\\
    \midrule
    
    Base model: \textbf{\llama-2-13B}\\
    \includegraphics[width=0.95\linewidth]{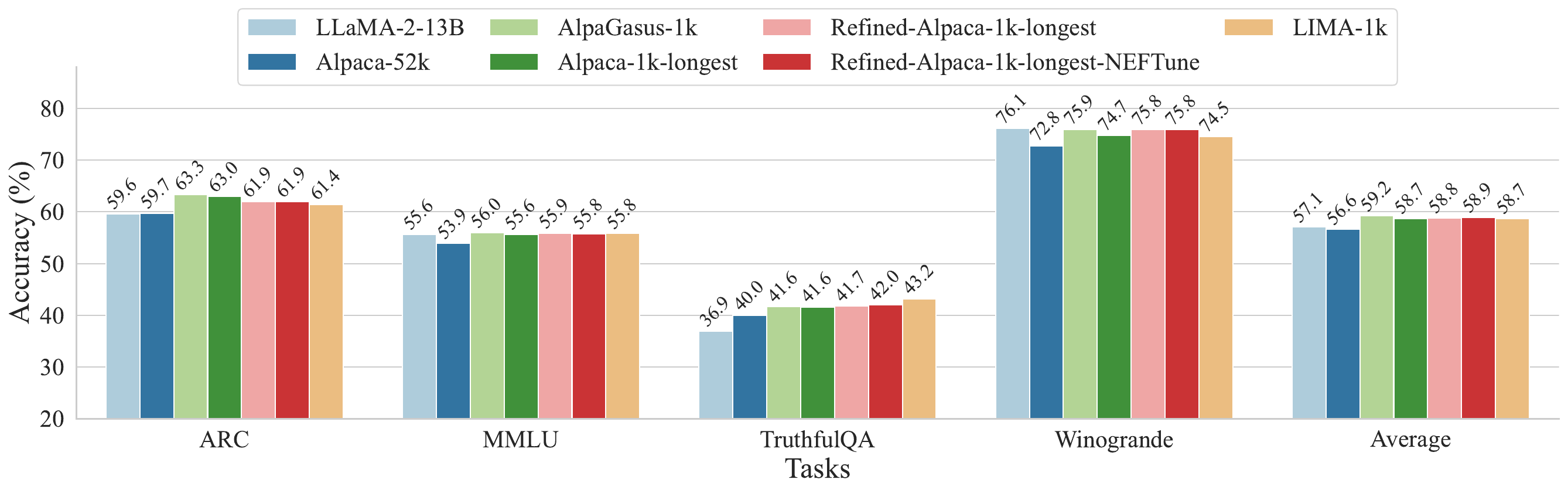}

    \end{tabular}
    
    \caption{\textbf{Open LLM Leaderboard tasks with Mistral-7B-v0.1 and \llama-2-13B fine-tuned on \alpaca-based datasets and \lima.}
    The model fine-tuned on \alpaca-1k-\longest achieves comparable performance to that of \lima-1k and significantly outperforms both base models and \alpaca-52k on average, showing that the performance gain on instruction-following capability does not compromise factuality. %
    }
    \label{fig:openllm_results}
\end{figure*}

\begin{figure*}[h!]
    \centering
    \includegraphics[width=\linewidth]{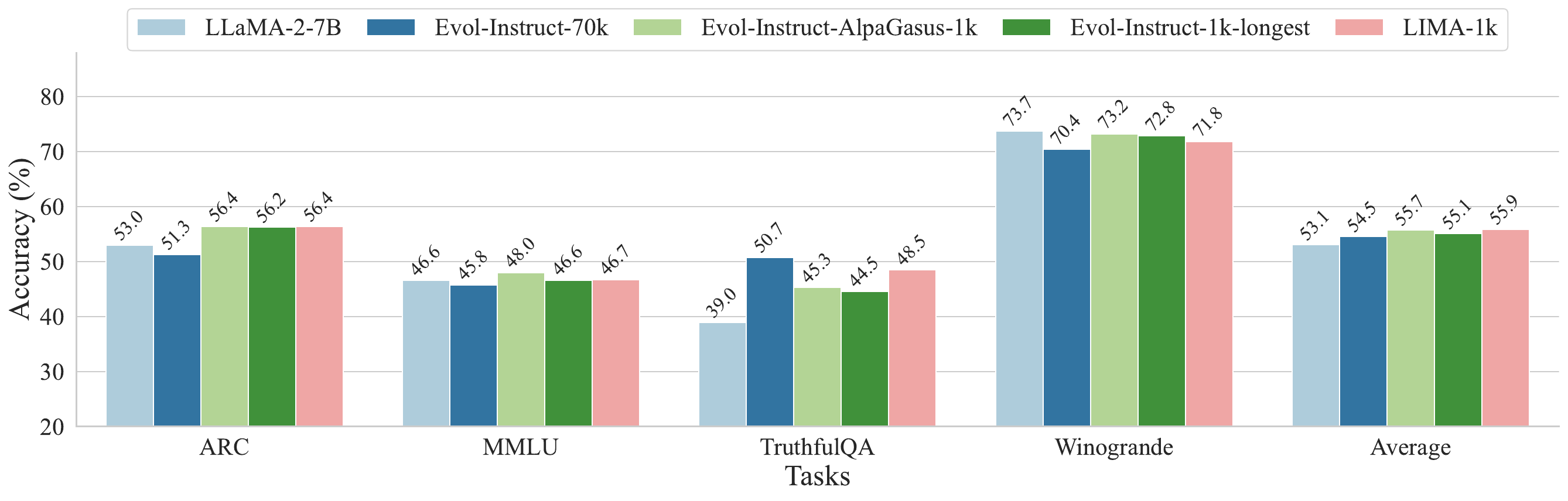}
    \vspace{-2em}
    \caption{\textbf{Open LLM Leaderboard tasks with \llama-2-7B fine-tuned on \evol-based datasets and \lima.} The model fine-tuned on \evol-1k-longest surpasses \llama-2-7B and \evol-70k on average, showing that the performance gain on instruction-following capability does not compromise factuality.
    }
    \label{fig:openllm_llama2_7b_evol_instruct}
\end{figure*}

This section contains the evaluation results of Mistral-7B-v0.1 model and \llama-2-13B model on (Fig.~\ref{fig:openllm_results}) and of \llama-2-7B fine-tuned on \evol-based datasets (Fig.~\ref{fig:openllm_llama2_7b_evol_instruct}) over the Open LLM benchmark.

\subsection{Empirical proof of the intuition behind utilizing longer examples for IFT}
\label{app:empirical_proof_intuition_longer_context}

\begin{figure}[t]
    \centering
    \subfigure[\small Base dataset: \alpaca-52k]{
        \includegraphics[width=0.47\textwidth]{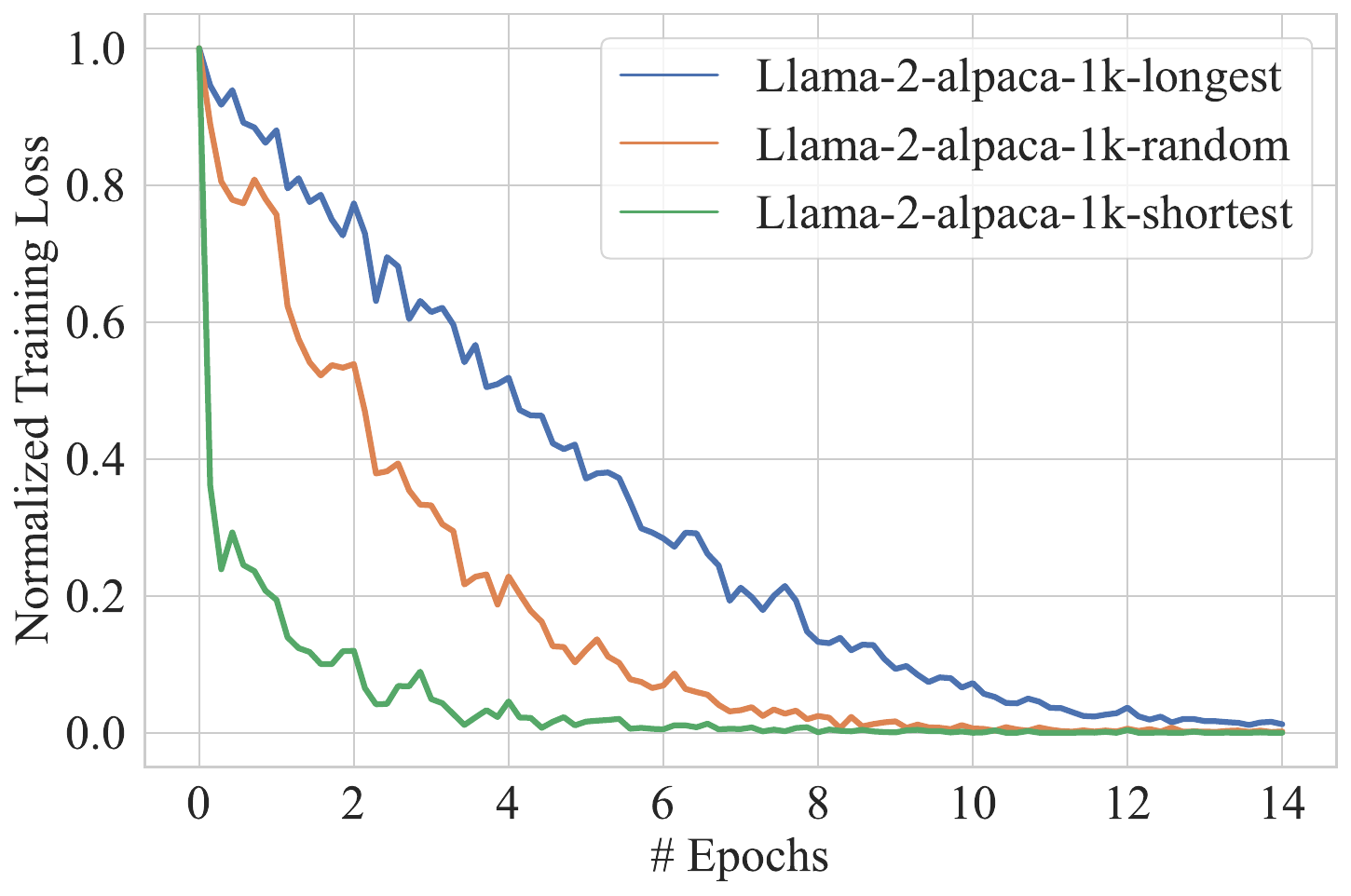}
        \label{fig:alpaca_training_loss_curve}
    }
    \subfigure[\small Base dataset: \evol]{
        \includegraphics[width=0.46\textwidth]{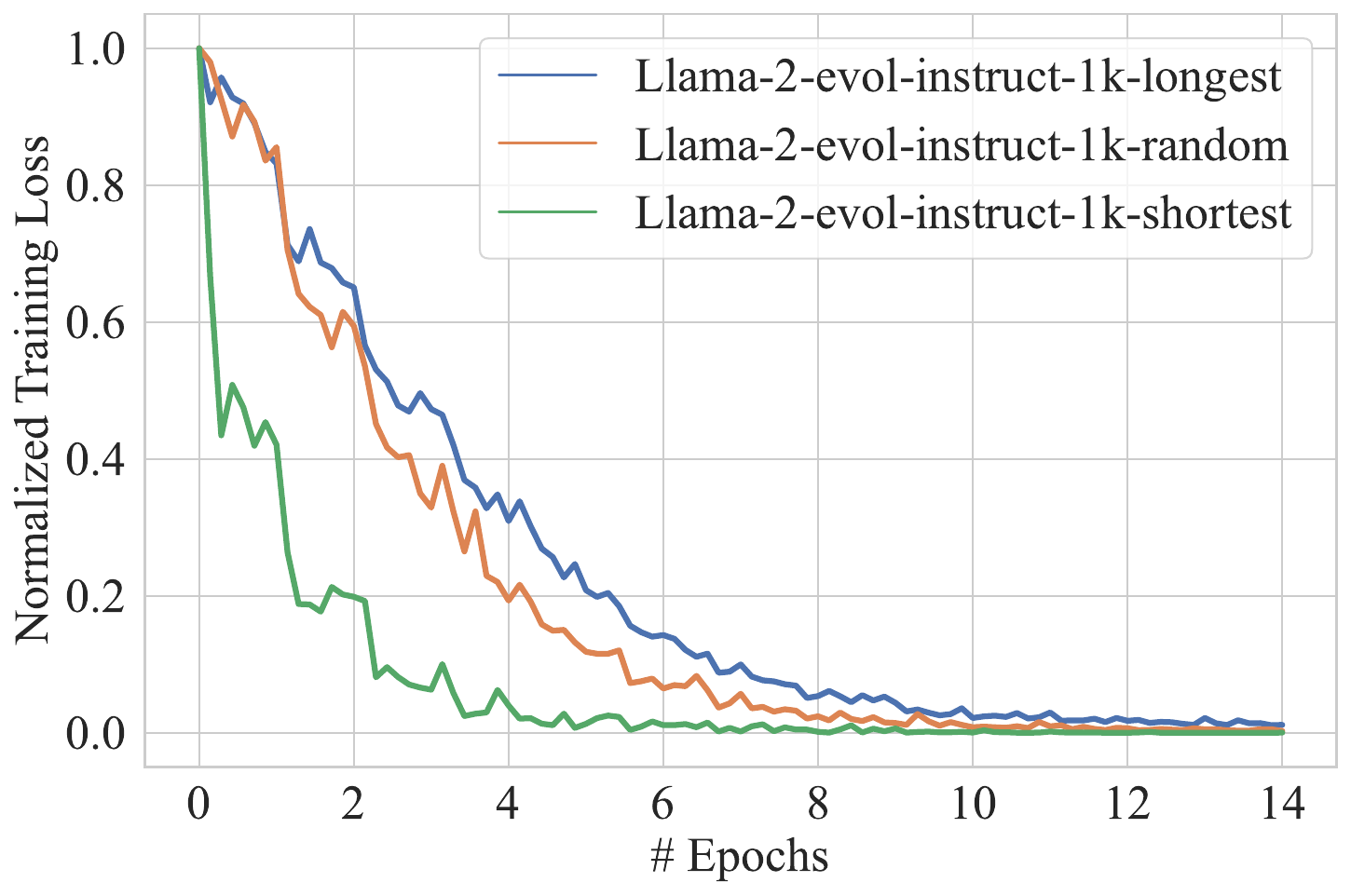}
        \label{fig:evol_instruct_training_loss_curve}
    }
    \caption{\textbf{(Normalized) Training loss curve of Llama2-7B models instruction fine-tuned on different subsets of the (a) \alpaca-52k and (b) \evol datasets.} We utilize normalized training loss because the models' initial training loss values were varied. We conduct normalization by dividing with the initial training loss value. Instruction fine-tuning on longer examples makes the loss converge more slowly than the others during fine-tuning of Llama-2-7B models. }
    \label{fig:training_loss_curve}
\end{figure}

To support the claim that longer responses are harder to fit, we compare the progress of the training loss over epochs of 1k-\longest to 1k-shortest and 1k-random, i.e., the subsets containing the 1k shortest or 1k arbitrarily long responses, respectively. For both Alpaca and Evol-Instruct, we fine-tune Llama-2-7B on each split: Fig.~\ref{fig:alpaca_training_loss_curve} and Fig.~\ref{fig:evol_instruct_training_loss_curve} show that the loss, normalized by its initial value to make it comparable across training sets, converges more slowly when using 1k-longest compared to other subsets. This confirms our hypothesis that longer instructions are harder to fit and thus provide more supervision signals and lead to better generalization. Moreover, Table~\ref{tab:h2h_comp_shortest_random} 
 and Table~\ref{tab:mtbench_shortest_random} show the resulting evaluation results using PaLM2 as the judge and MT-Bench, respectively, both of which align with our intuition.

We verify the intuition that fine-tuning using longer examples makes the model more effective in capturing long-distance semantic connections on three long-form summarization tasks: GovReport~\cite{huang2021efficient}, SummScreen~\cite{chen2022summscreen}, and QMSum~\cite{zhong2021qmsum}, from the SCROLLS~\cite{shaham2022scrolls} benchmark. We measure the quality of the generated summary using the ROUGE-1, ROUGE-2, and ROUGE-L scores. Detailed results are shown in Table~\ref{tab:long_form_summ}. Our 1k-\longest model always achieves the best (i.e., in SummScreen) or the 2nd best (i.e., in GovReport and QMSum) performance. It validates that fine-tuning using long examples maintains strong summarization capability, while showing superior performance on long-form tasks.

\begin{table*}[htbp]
    \caption{\textbf{Evaluation results on three long-form summarization tasks. } We use the ROUGE-1, ROUGE-2, and ROUGE-L scores (the higher score is better) to measure the summarization quality. Our 1k-\longest model always achieves the best or the 2nd best performance, indicating its superior performance on long-form summarization tasks. }
    \setlength{\tabcolsep}{2pt}
    \def\arraystretch{1.15}
    \begin{center}
    \large
    \resizebox{1.0\linewidth}{!}{
        \begin{tabular}{lcccccccccccc}
        \toprule
         & \multicolumn{4}{c}{GovReport} & \multicolumn{4}{c}{SummScreen} & \multicolumn{4}{c}{QMSum}  \\
        \cmidrule(lr){2-5} \cmidrule(lr){6-9} \cmidrule(lr){10-13} %
        Models   &   ROUGE-1   &   ROUGE-2   &   ROUGE-L   &   Avg.   &   ROUGE-1   &   ROUGE-2   &   ROUGE-L   &   Avg.   &   ROUGE-1   &   ROUGE-2   &   ROUGE-L   &   Avg.   \\
        \midrule
        \llama-2-7B (base)  & $4.58$ & $1.66$ & $3.22$ & $3.15$ & $13.18$ & $2.06$ & $9.77$ & $8.34$ & $19.16$ & $5.37$ & $14.89$ & $13.14$ \\
        \alpaca-52k         & $20.91$ & $8.60$ & $13.05$ & $14.19$ & $23.80$ & $3.37$ & $14.28$ & $13.82$ & $26.94$ & $\textbf{6.28}$ & $\textbf{18.35}$ & $\textbf{17.19}$ \\
        \alpagasus-1k       & $\textbf{23.02}$ & $\textbf{8.92}$ & $\textbf{13.68}$ & $\textbf{15.21}$ & $\underline{26.40}$ & $\underline{3.58}$ & $\underline{15.06}$ & $\underline{15.01}$ & $\underline{27.09}$ & $5.74$ & $18.10$ & $16.98$ \\
        \alpaca-1k-\longest & $\underline{22.20}$ & $\underline{8.85}$ & $\underline{13.35}$ & $\underline{14.80}$ & $\textbf{26.56}$ & $\textbf{3.81}$ & $\textbf{15.45}$ & $\textbf{15.27}$ & $\textbf{27.40}$ & $\underline{5.80}$ & $\underline{18.16}$ & $\underline{17.12}$ \\
        \bottomrule
        \end{tabular}
    }
    \end{center}
    \label{tab:long_form_summ}
\end{table*}

\begin{table}[t]
    \caption{\textbf{Head-to-head comparisons (win rates in \%) to 1k-shortest and 1k-random across different IFT datasets.}}
    \vspace{2mm}
    \centering
    \small \tabcolsep=2pt
    \begin{tabular}{L{30mm} C{21mm} C{12mm} C{21mm}}
        \toprule
         Model        & 1k-\longest wins & Tie & 1k-longest loses \\
        \midrule
        \addlinespace[2mm]
        
        \multicolumn{4}{l}{\textbf{Base dataset: Alpaca-52k}} \\
        \toprule
        1k-shortest   & $\textbf{97.0}$ &  $2.6$  &  $0.4$  \\
        1k-random     & $\textbf{72.1}$ &  $18.0$ &  $9.9$  \\
        \midrule
        \addlinespace[2mm]
        
        \multicolumn{4}{l}{\textbf{Base dataset: Evol-Instruct-70k}} \\
        \toprule
        1k-shortest   & $\textbf{93.4}$ &  $4.9$  &  $1.7$  \\
        1k-random     & $\textbf{39.7}$ &  $29.1$ &  $31.2$  \\
        \midrule
        \addlinespace[2mm]
        
        \multicolumn{4}{l}{\textbf{Base dataset: Open-Hermes-1M}} \\
        \toprule
        1k-shortest   & $\textbf{95.9}$ &  $3.5$  &  $0.6$  \\
        1k-random     & $\textbf{84.3}$ &  $10.4$ &  $5.3$  \\
        \bottomrule
    \end{tabular}
    \label{tab:h2h_comp_shortest_random}
\end{table}

\begin{table}[t]
    \caption{\textbf{Results on MT-Bench for Llama-2-7B fine-tuned on different IFT datasets.}
    }
    \vspace{2mm}
    \centering
    \small \tabcolsep=2pt
    \begin{tabular}{L{20mm} C{21mm} C{30mm} C{30mm}}
        \toprule
         Model & \alpaca-52k  & \evol-70k  & Open-Hermes-1M \\
         \midrule
         1k-shortest & $1.78$ & $2.34$ & $1.46$ \\
         1k-random   & $3.74$ & $4.03$ & $4.06$ \\
         1k-\longest & $\textbf{3.96}$ & $\textbf{4.27}$ & $\textbf{4.18}$ \\
        
        \bottomrule
    \end{tabular}
    \label{tab:mtbench_shortest_random}
\end{table}

\subsection{Diversity of training examples}
\label{app:diversity}

To examine the effect of data diversity, we test our approach on the Open-Hermes-2.5~\cite{OpenHermes2.5} dataset, which includes data (around 1M instructions) from different sources. Table~\ref{tab:diversity} shows that selecting the 1k-longest with stratified sampling, i.e., preserving diverse data sources, results in slightly better performance on the MT-Bench than uniform sampling. This shows the importance of preserving the diversity of the original dataset. However, while this is simple to do for Open-Hermes-2.5 (where the source of data is available), it is not straightforward to control on other datasets, such as Alpaca without manual inspections. Finally, the same observation holds with Mistral-7B-v0.1 as the base model, where our 1k-longest with stratified sampling achieves results very close to the entire dataset, which is 1000x larger. 

\begin{table}[t]
    \caption{\textbf{Performance on MT-Bench for models fine-tuned on Open-Hermes-2.5.}
    }
    \vspace{2mm}
    \centering
    \small \tabcolsep=2pt
    \begin{tabular}{L{60mm} C{25mm}}
        \toprule
         Model & MT-Bench Score \\
        \midrule
        \addlinespace[2mm]
        
        \multicolumn{2}{l}{\textbf{Base model: Llama-2-7B}} \\
        \toprule
        Open-Hermes-1k-\longest (uniform sampling)    & $4.18$  \\
        Open-Hermes-1k-\longest (stratified sampling) & $4.31$  \\
        \midrule
        \addlinespace[2mm]
        
        \multicolumn{2}{l}{\textbf{Base model: Mistral-7B-v0.1}} \\
        \toprule
        Open-Hermes-1k-\longest (uniform sampling)    & $6.83$  \\
        Open-Hermes-1k-\longest (stratified sampling) & $6.97$  \\
        Open-Hermes-1M                                & $7.22$  \\
        \bottomrule
    \end{tabular}
    \label{tab:diversity}
\end{table}

\section{Comparison to additional baselines} 

\subsection{\alpagasus-9k}
\label{app:alpagasus-9k}

\begin{figure}[htbp]
    \centering
    \subfigure[\small Alpaca-9k-longest vs. AlpaGasus-9k]{
        \includegraphics[width=0.46\textwidth]{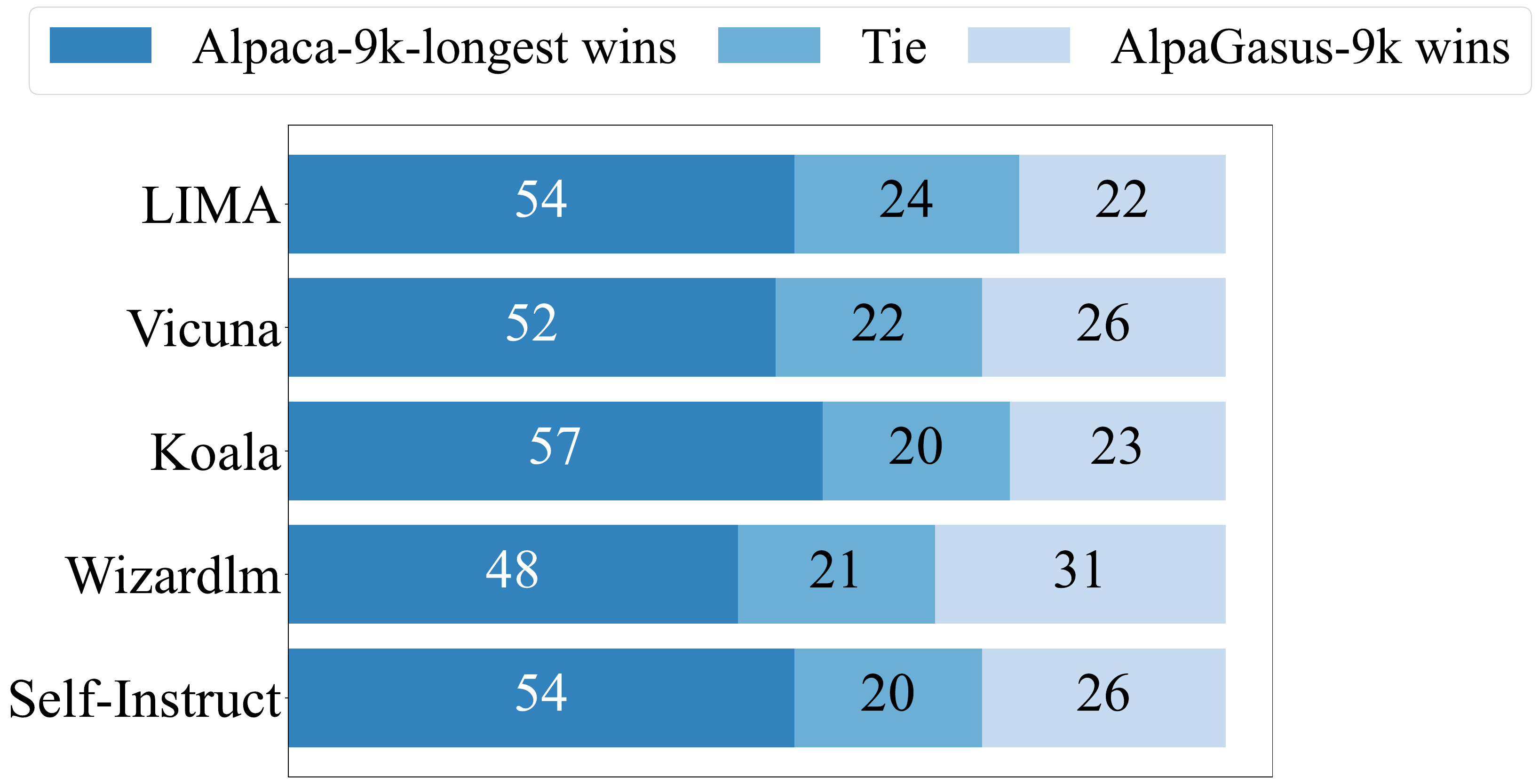}
        \label{fig:llama_2_alpaca_9k_longest_vs_alpagasus_9k_gpt_4}
    }
    \hfill
    \subfigure[\small Alpaca-1k-longest vs. AlpaGasus-9k]{
        \includegraphics[width=0.46\textwidth]{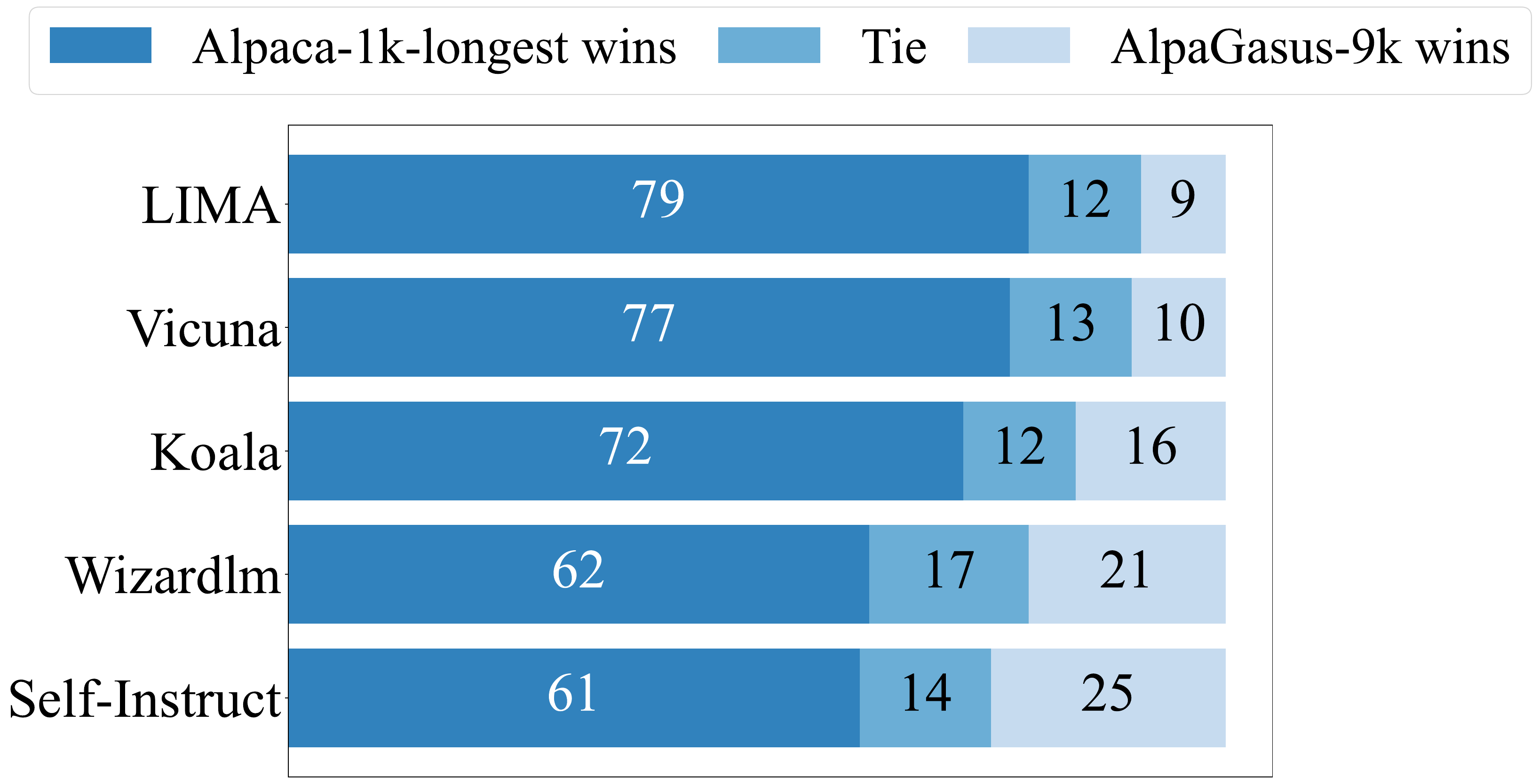}
        \label{fig:llama_2_alpaca_1k_longest_vs_alpagasus_9k_gpt_4}
    }
    
    \caption{\textbf{Preference evaluation (\%) using GPT4-as-a-judge} on LLaMA-2-7B models fine-tuned on \alpagasus-9k, \alpaca-9k-\longest, and \alpaca-1k-\longest.}
    \label{fig:alpagasus_9k_gpt_4}
\end{figure}

In this section, we validate the advantage of length heuristics by comparing \alpaca-9k-\longest with \alpagasus-9k, which is the best filtered subset from Alpaca-52k in the AlpaGasus paper~\citep{chen2023alpagasus}. The detailed experimental results are shown in Fig.~\ref{fig:llama_2_alpaca_9k_longest_vs_alpagasus_9k_gpt_4}, where \alpaca-9k-\longest consistently outperforms \alpagasus-9k in 5 evaluation sets. We further show comparisons between \alpaca-1k-\longest and \alpagasus-9k in Fig.~\ref{fig:llama_2_alpaca_1k_longest_vs_alpagasus_9k_gpt_4}, which also supports our main claim: length is a strong criterion for constructing instruction fine-tuning dataset. Details of experimental setup can be seen in Table~\ref{tab:training_hparams}.

\subsection{Reflection-tuning}
\label{app:introspection_advantage}

In this section, we show the advantage of proposed introspection technique by comparing it with reflection-tuning~\citep{li2023reflection} on \llama-2-7B and \llama-2-13B models. We present experimental results on the Open LLM benchmark and AlpacaEval 2.0 in Table~\ref{tab:reflection_tuning}.

\begin{table*}[htbp]
    \caption{Comparison between our introspection strategy and that of in the Reflection-Tuning~\citep{li2023reflection}. * denotes that results are copied from the paper. Note that the performance of \llama-2-7B-Recycled-Alpaca-52k on the Winogrande task is evaluated using the open-source model checkpoint provided by Reflection-Tuning. }
    \vspace{0.3em}
    \centering
    \resizebox{0.95\textwidth}{!}{
        \begin{tabular}{l|ccccccccc}
        \toprule
             Models                     & \# SFT data       & ARC  & HellaSwag & MMLU & TruthfulQA & Winogrande & Average & AlpacaEval 2.0 & Avg. Length \\
             \midrule
             \llama-2-7B & 0 & $52.99$ & $78.64$ & $46.56$ & $38.97$ & $73.72$ & $58.18$ & / & / \\
             \llama-2-7B-Alpaca-52k & 52k & $53.92$ & $78.82$ & $47.05$ & $40.32$ & $71.82$ & $58.39$ & $2.74$ & $586$ \\
             \llama-2-7B-Recycled-Alpaca-52k* & 52k & $53.92$ & $77.68$ & $\textbf{47.55}$ & $45.55$ & $71.82$ & $59.30$ & $5.93$ & $1470$ \\
             \llama-2-7B-Refined-Alpaca-1k-L & 1k & $\textbf{56.74}$ & $\textbf{80.23}$ & $46.82$ & $\textbf{49.59}$ & $\textbf{72.45}$ & $\textbf{61.17}$ & $\textbf{6.00}$ & $1732$ \\
             \midrule
             \llama-2-13B & 0 & $59.64$ & $82.15$ & $55.63$ & $36.92$ & $76.09$ & $62.09$ & / & / \\
             \llama-2-13B-Alpaca-52k & 52k & $59.73$ & $83.08$ & $53.87$ & $39.98$ & $72.77$ & $61.24$ & $3.90$ & $556$ \\
             \llama-2-13B-Recycled-Alpaca-52k* & 52k & $58.70$ & $80.80$ & $53.11$ & $\textbf{43.12}$ & ? & ? & ? & ?\\
             \llama-2-13B-Refined-Alpaca-1k-L & 1k & $\textbf{61.95}$ & $\textbf{83.88}$ & $\textbf{55.86}$ & $41.74$ & $\textbf{75.85}$ & $\textbf{63.86}$ & $\textbf{8.44}$ & $1646$ \\
        \bottomrule
        \end{tabular}
    }
    \label{tab:reflection_tuning}
\end{table*}

\section{Case study}
\label{app:case_study}

\begin{figure*}[htbp]
    \centering
    \subfigure[\small Example \#1: generate an itinerary in Switzerland]{
        \includegraphics[width=0.8\textwidth]{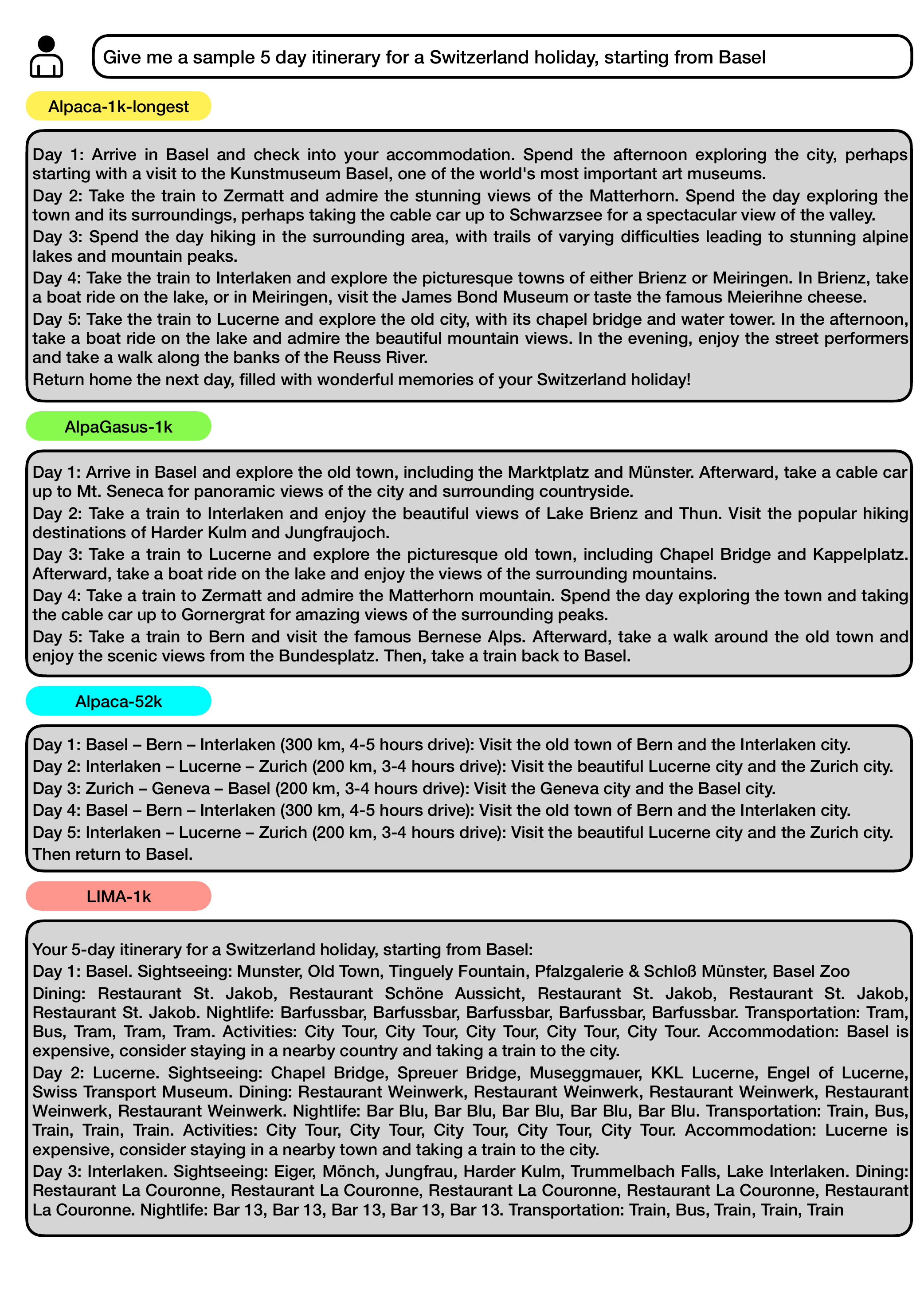}
    }
    \caption{Case study on the LLaMA-2-7B model fine-tuned on different IFT datasets.}
    \label{fig:llama2_7b_case_study_1}
    
\end{figure*}

\begin{figure*}[htbp]
    \centering
    \subfigure[\small Example \#2: give an inspiring speech as a pirate captain]{
        \includegraphics[width=0.46\textwidth]{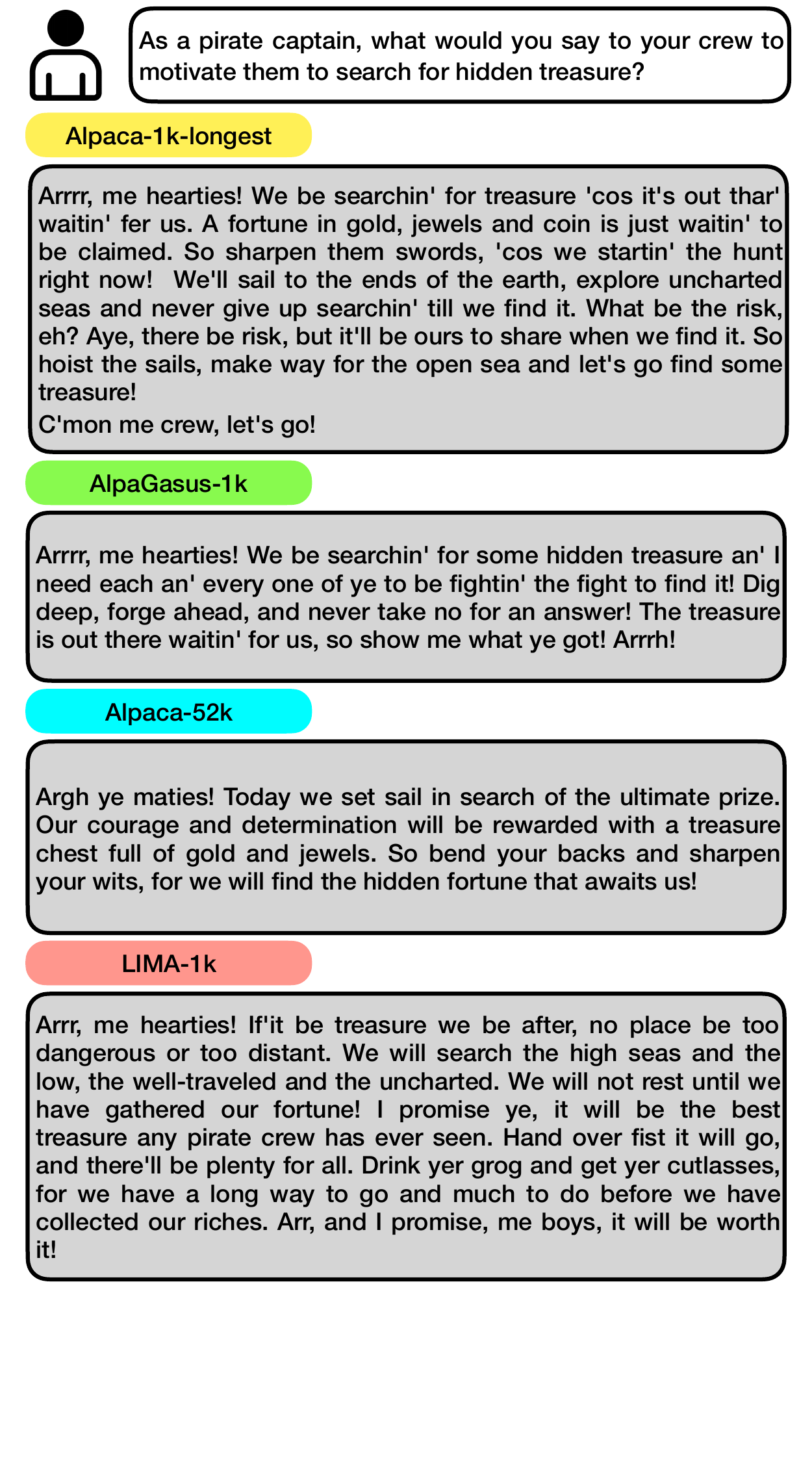}
    }
    \subfigure[\small Example \#3: write a code snippet to validate an email address]{
        \includegraphics[width=0.46\textwidth]{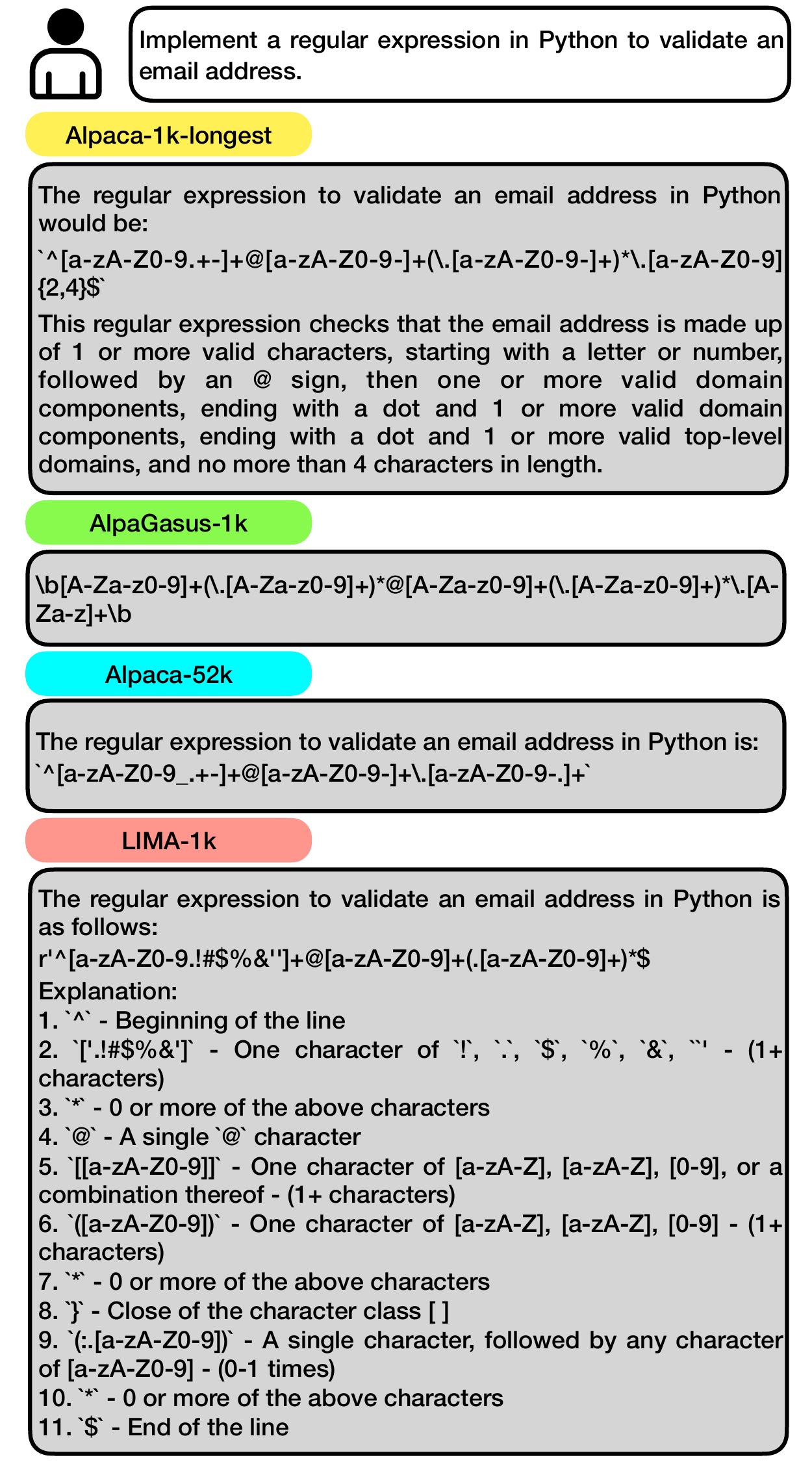}
    }
    \caption{Case study on the LLaMA-2-7B model fine-tuned on different IFT datasets.}
    \label{fig:llama2_7b_case_study_2_3}
    
\end{figure*}

\begin{figure*}[htbp]
    \centering
    \subfigure[\small Example \#4: reasoning]{
        \includegraphics[width=0.46\textwidth]{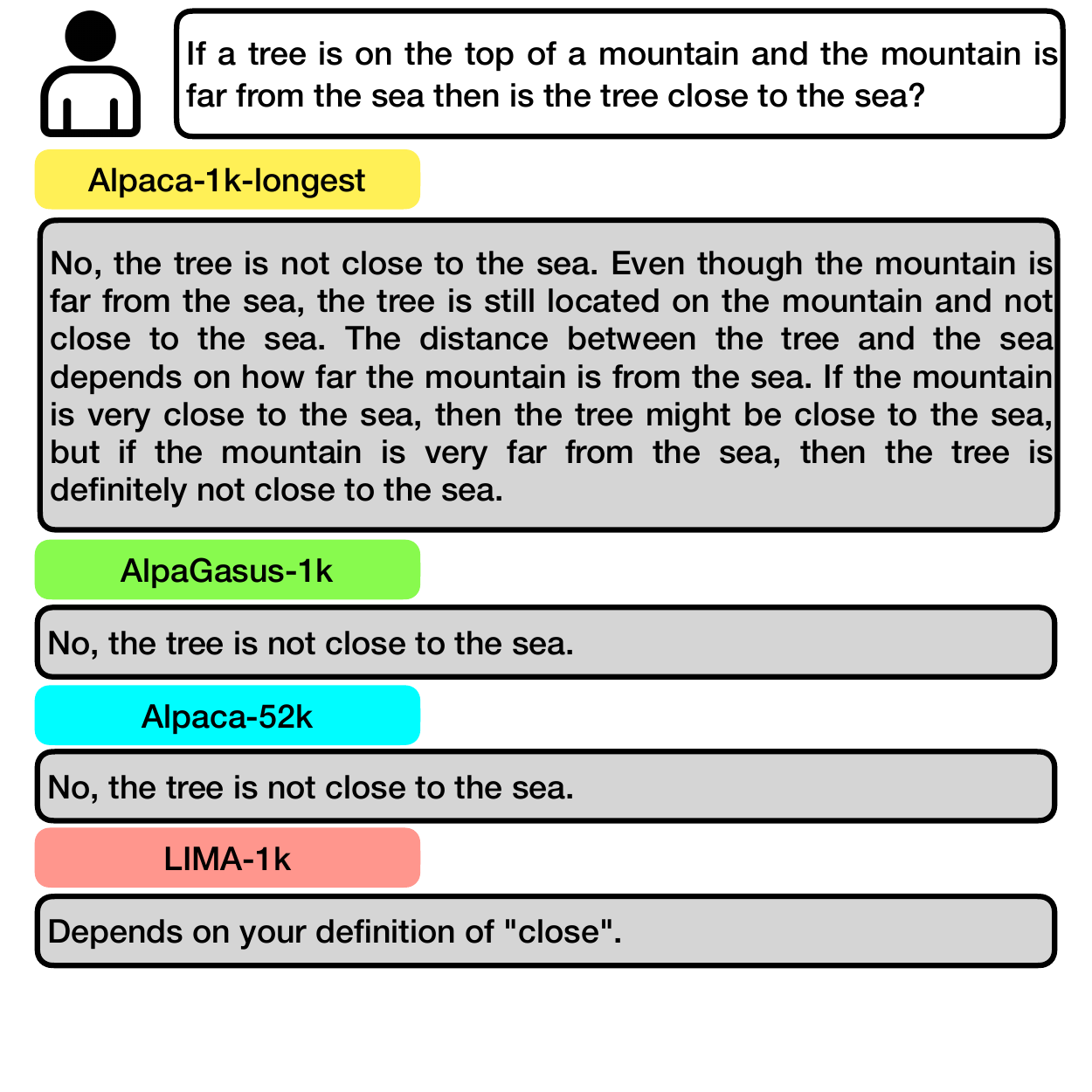}
    }
    \subfigure[\small Example \#5: reasoning]{
        \includegraphics[width=0.46\textwidth]{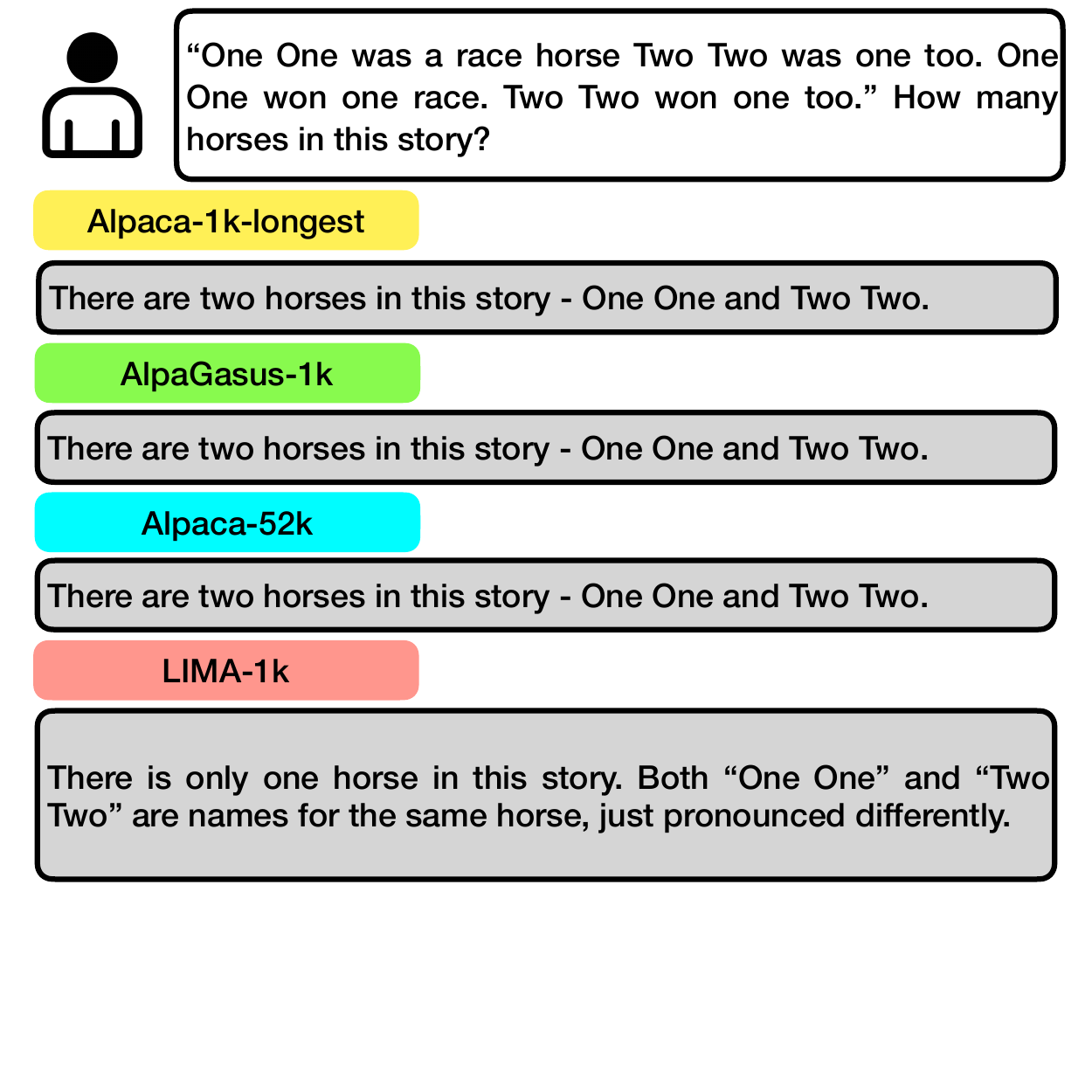}
    }
    \subfigure[\small Example \#6: calculate average profits]{
        \includegraphics[width=0.46\textwidth]{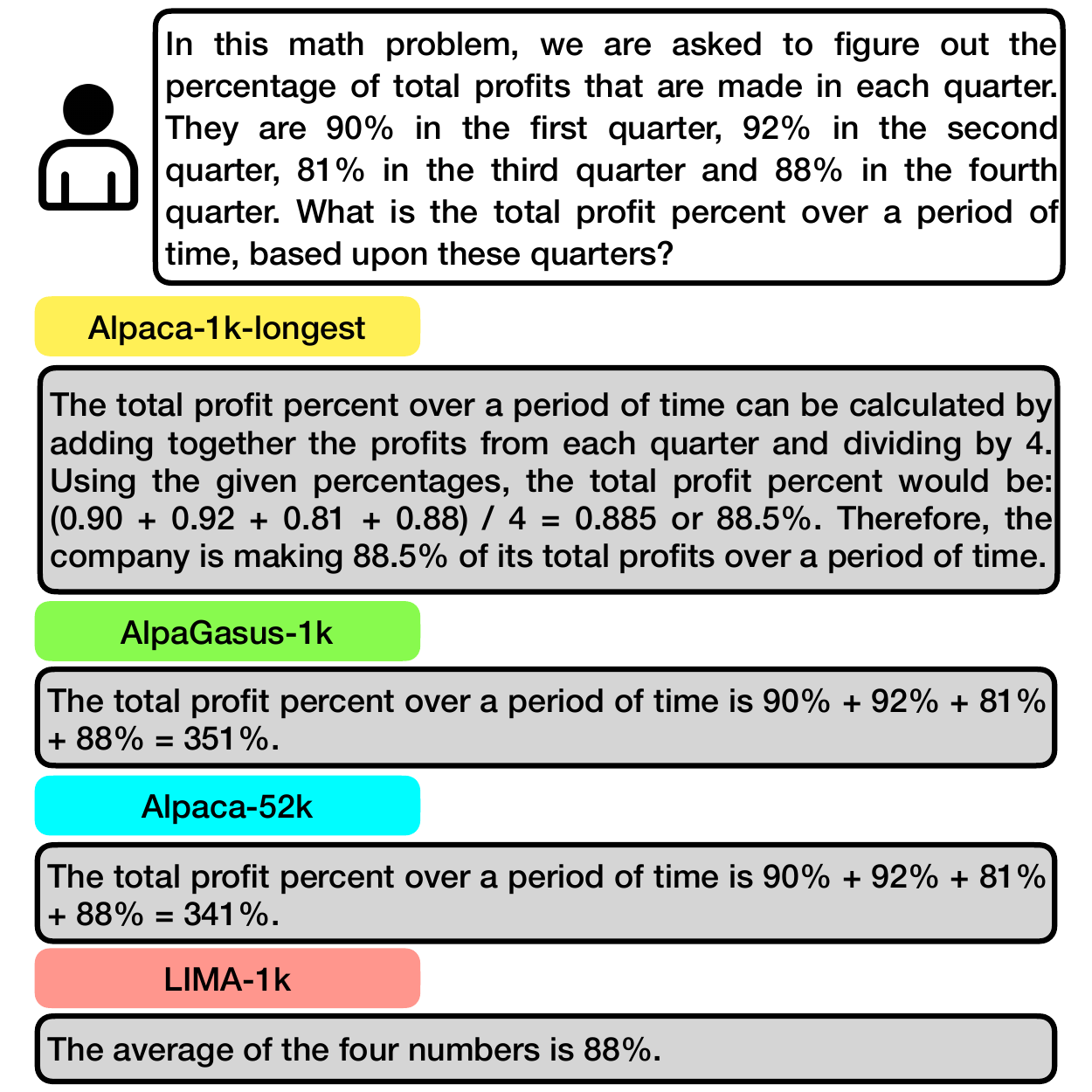}
    }
    \subfigure[\small Example \#7: what language does Argentina people speak]{
        \includegraphics[width=0.46\textwidth]{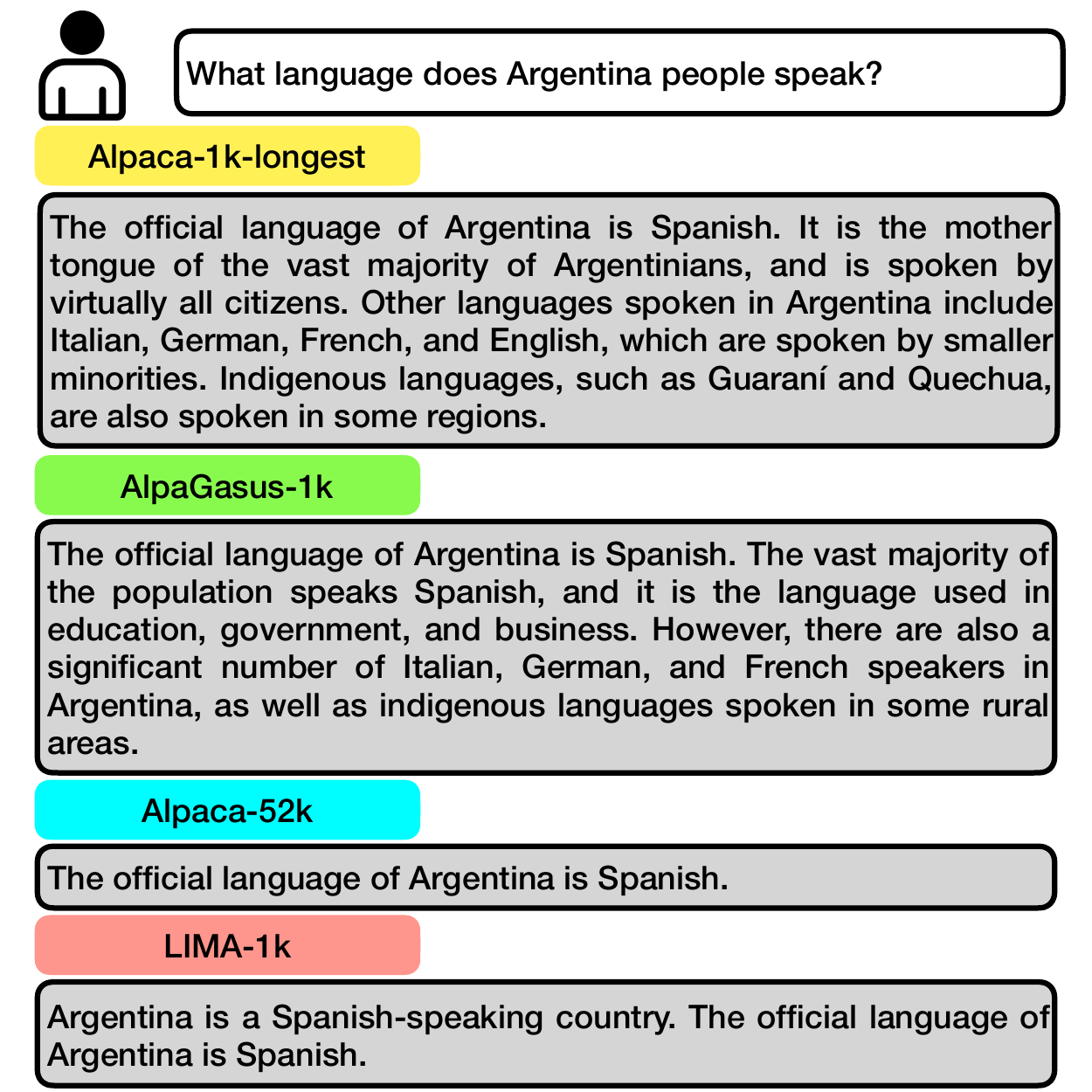}
    }
    \caption{Case study on the Mistral-7B-v0.1 model fine-tuned on different IFT datasets.}
    \label{fig:mistral_7b_case_study_4_5_6_7}
    
\end{figure*}

\begin{figure*}[htbp]
    \centering
    \subfigure[\small Example \#8: generate a birthday wish message]{
        \includegraphics[width=0.46\textwidth]{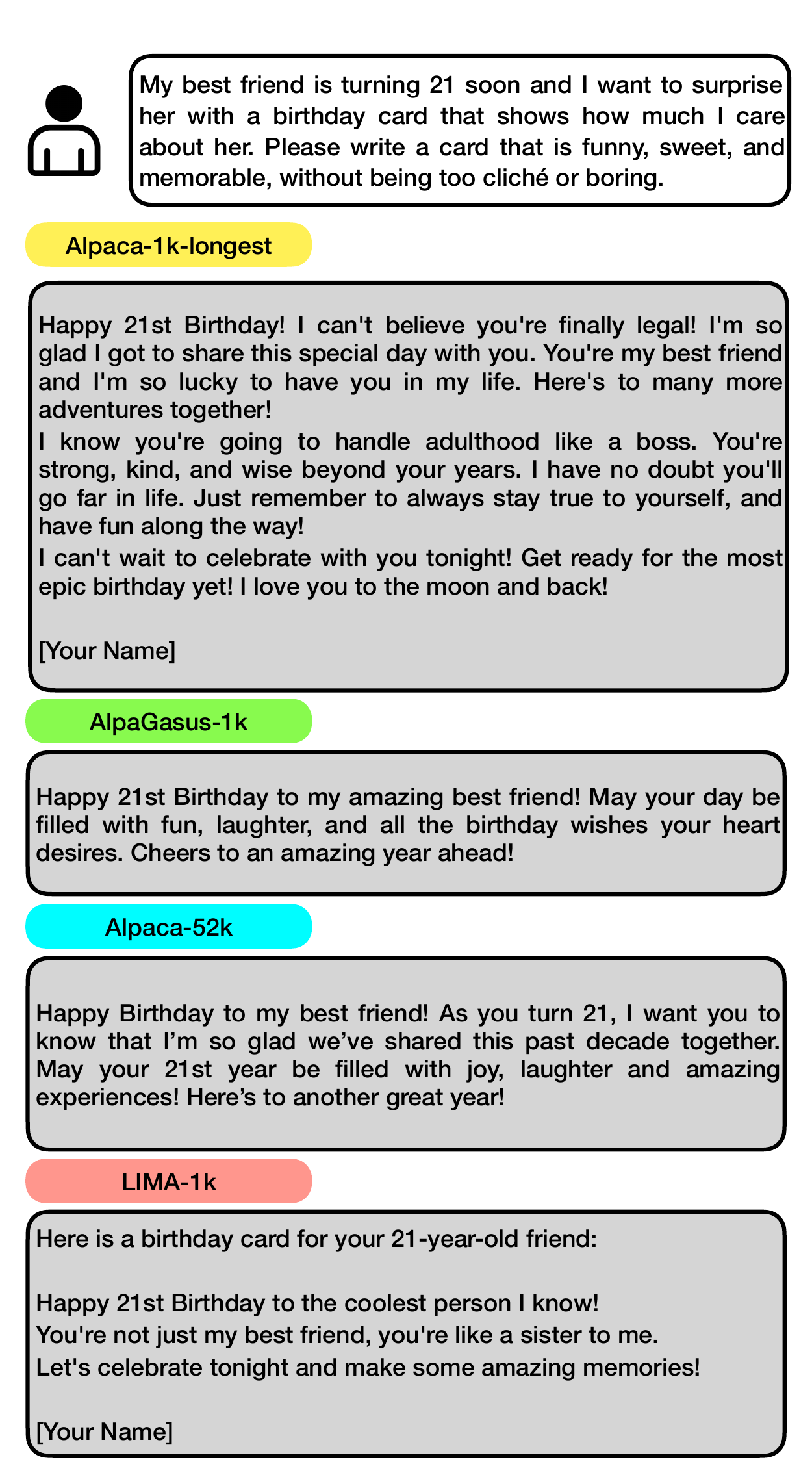}
    }
    \subfigure[\small Example \#9: answer a biology question]{
        \includegraphics[width=0.46\textwidth]{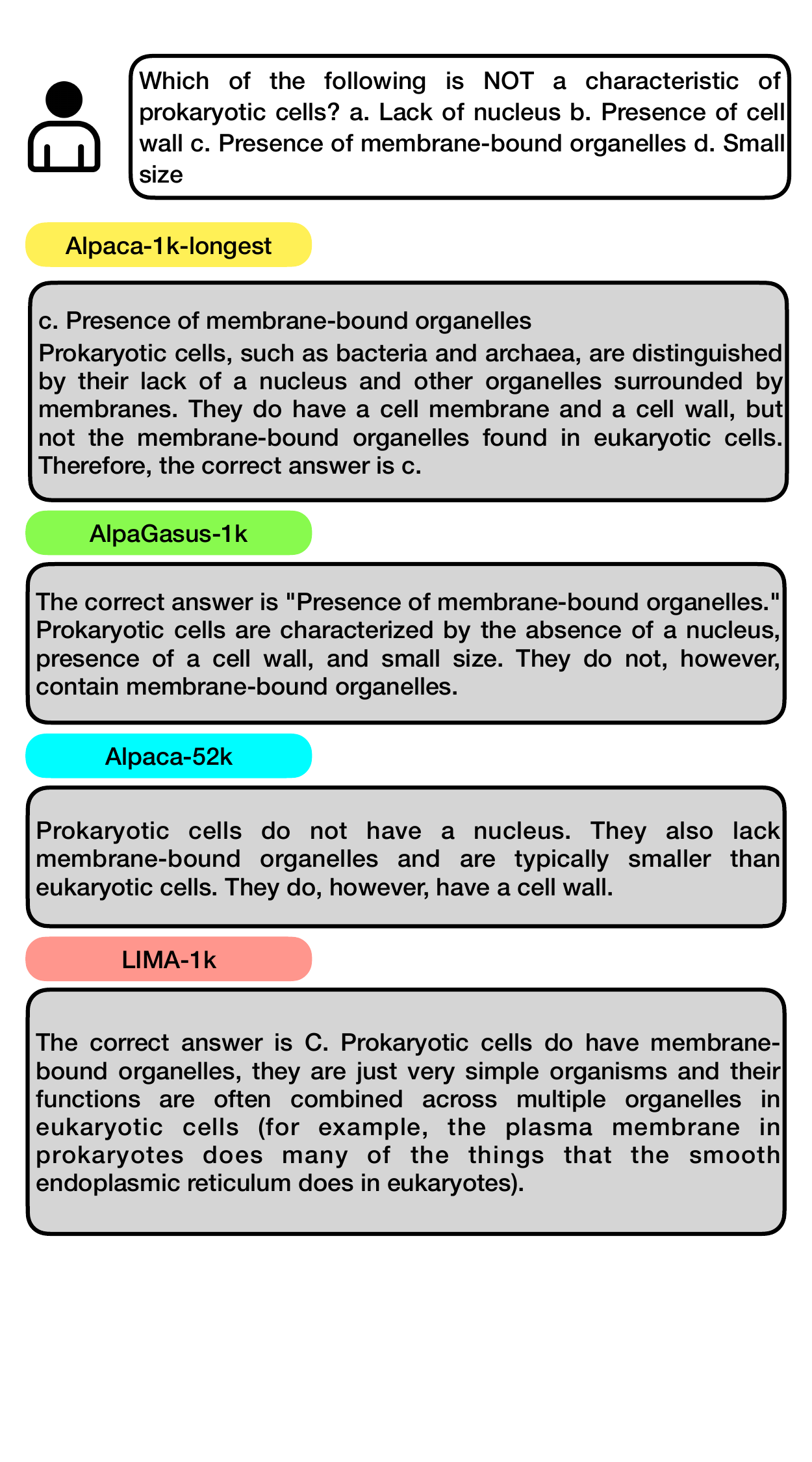}
    }
    \caption{Case study on the LLaMA-2-13B model fine-tuned on different IFT datasets.}
    \label{fig:llama2_13b_case_study_8_9}
    
\end{figure*}

\begin{figure*}[htbp]
    \centering
    \subfigure[\small Example \#10: human society development]{
        \includegraphics[width=0.8\textwidth]{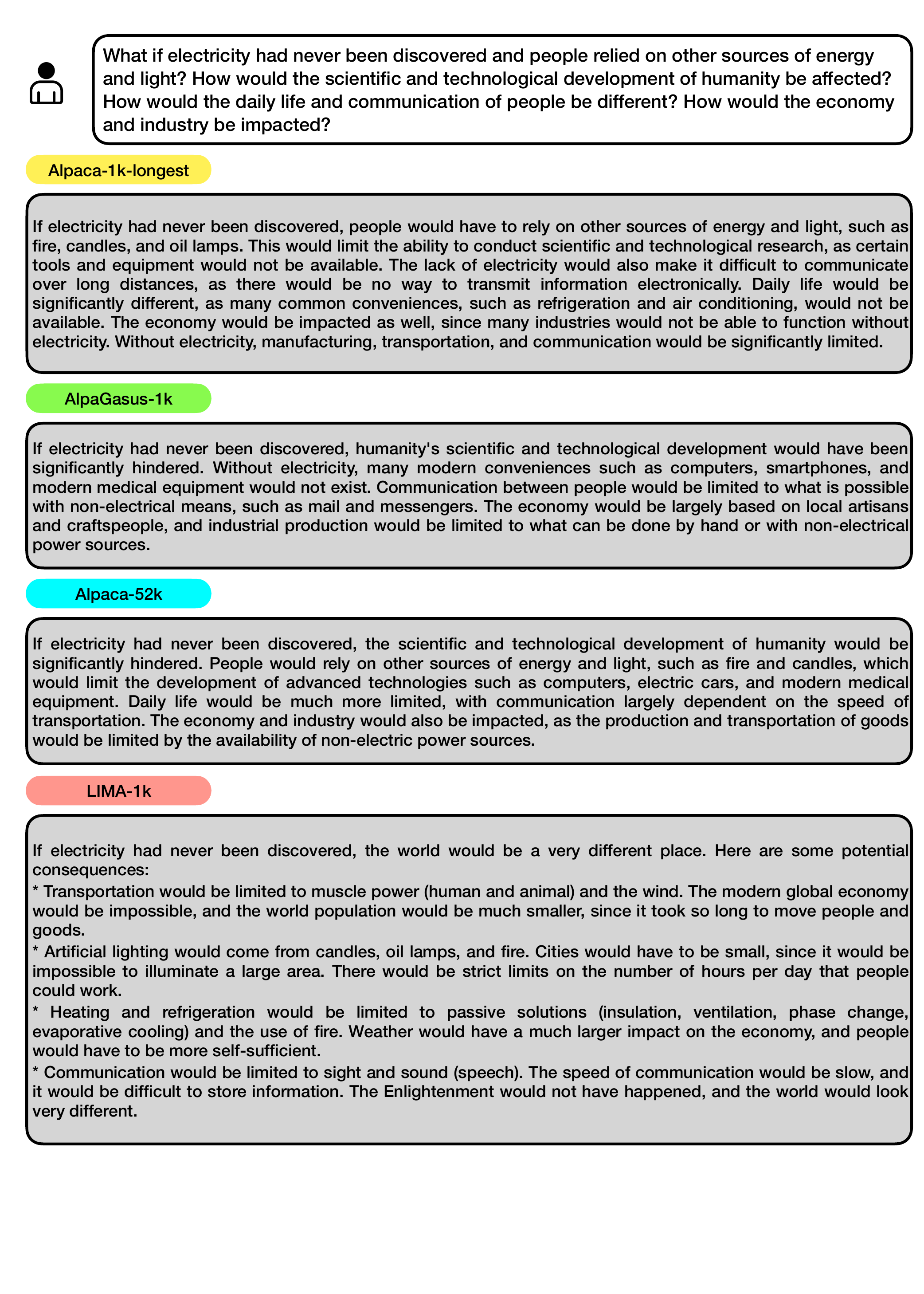}
    }
    \caption{Case study on the LLaMA-2-13B model fine-tuned on different IFT datasets.}
    \label{fig:llama2_13b_case_study_10}
    
\end{figure*}

This section consists of ten test instructions and corresponding responses of \llama-2-7B (Fig.~\ref{fig:llama2_7b_case_study_1} and Fig.~\ref{fig:llama2_7b_case_study_2_3}), Mistral-7B-v0.1 (Fig.~\ref{fig:mistral_7b_case_study_4_5_6_7}), and \llama-2-13B (Fig.~\ref{fig:llama2_13b_case_study_8_9} and Fig.~\ref{fig:llama2_13b_case_study_10}) models fine-tuned on \alpaca-1k-\longest, \alpagasus-1k, \alpaca-52k, and \lima-1k datasets. Details of training hyperparameters are shown in Table~\ref{tab:training_hparams}. We add detailed comments for qualitative analysis on responses generated by \llama-2-7B in Section~\ref{app:qualitative_analysis_llama2_7b}. We omit detailed analysis for Mistral-7B-v0.1 and \llama-2-13B since we make similar observations as for \llama-2-7B.

\subsection{Detailed comments on \llama-2-7B examples}
\label{app:qualitative_analysis_llama2_7b}

\paragraph{Example \#1: generate an itinerary in Switzerland. }

\begin{itemize}
    \item \alpaca-1k-\longest provides a well-structured and detailed itinerary for a 5-day trip to Switzerland, starting from Basel. It includes a variety of activities, such as visiting museums, hiking, exploring towns, and enjoying local cuisine. It also suggests different modes of transportation, such as trains and cable cars, which are common in Switzerland. Its answer is relevant, accurate, and helpful. However it mentions a ``famous Meierihne cheese'', which does not exist at all. We believe this hallucination happens because of the knowledge capabilities of the base model. 
    \item \alpagasus-1k also provides a well-structured response and includes a variety of activities, it is slightly less detailed than \alpaca-1k-\longest's response. For example, in Interlaken, \alpagasus-1k suggests visiting popular hiking destinations but did not provide any information about what one might see or do there. However, \alpagasus-1k does a good job of suggesting a variety of activities and destinations, making the itinerary interesting and diverse.
    \item \alpaca-52k's answer is less detailed and less helpful. The assistant suggested visiting the same cities on multiple days, which is not efficient or practical for a 5-day trip. The assistant also did not provide specific activities or places to visit in each city, which makes the answer less useful for someone planning a trip.
    \item \lima-1k's answer is cut off and does not cover the full 5 days. It also repeats the same dining and nightlife options for each day, which is not very helpful or realistic.
\end{itemize}

\paragraph{Example \#2: give an inspiring speech as a pirate captain. }

\begin{itemize}
    \item \alpaca-1k-\longest provides excellent responses to this question. It uses appropriate pirate language and provides motivating speeches that would encourage a pirate crew to search for hidden treasure. The response is relevant, accurate, and detailed, providing a vivid picture of the adventure and potential rewards.
    \item \alpagasus-1k's response is shorter and less detailed, but still motivational and in line with the question.
    \item \alpaca-52k's response is also motivational and uses appropriate language, but is less detailed and less vivid in its description of the journey and the treasure.
    \item \lima-1k also provides excellent responses to this question. It uses appropriate pirate language and provides motivating speeches that would encourage a pirate crew to search for hidden treasure. The response is relevant, accurate, and detailed, providing a vivid picture of the adventure and potential rewards.
\end{itemize}

\paragraph{Example \#3: write a code snippet to validate an email address. }

\begin{itemize}
    \item \alpaca-1k-\longest provides a correct regular expression for validating an email address in Python and also explained what each part of the expression does. The explanation was clear and concise, making it easy to understand how the regular expression works.
    \item \alpagasus-1k also provides a correct regular expression for validating an email address. However, there is no explanation or context provided, which might make it difficult for someone unfamiliar with regular expressions to understand.
    \item \alpaca-52k's answer is also correct and accurate, but lacks the detailed explanation
    \item \lima-1k's regular expression is incorrect and does not match the standard email format. The explanation provided by \lima-1k is also incorrect and confusing, as it does not correctly explain what each part of the regular expression does.
\end{itemize}

\section{Data contamintation on LIMA-1k}
\label{app:lima_contam}

\begin{figure*}[t]
    \centering
    \includegraphics[width=0.6\linewidth]{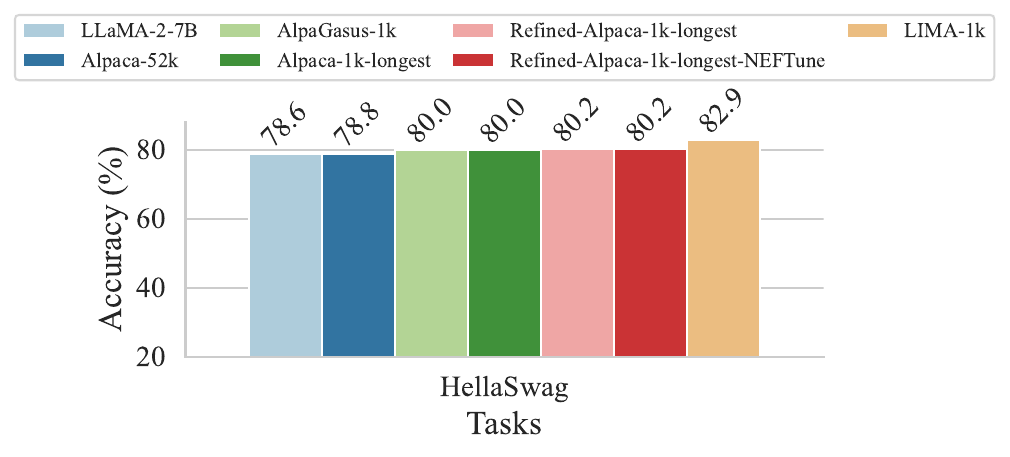}
    \vspace{-2mm}
    \caption{\small The performance of a diverse array of instruction fine-tuned models on the HellaSwag task.
    The very high accuracy of the models fine-tuned on \lima-1k might be explained by data contamination (see discussion in App.~\ref{app:lima_contam}).
    }
    \label{fig:openllm_llama2_7b_hellaswag}
\end{figure*}

With over 240k how-to articles covering a wide range of topics, wikiHow is an online publication in the style of a wiki, where articles are frequently regarded as high-quality content. \lima~\citep{zhou2023lima} contains 200 wikiHow examples. The article's title serves as a prompt (e.g., ``How to Cook an Omelet?'') and the body text as an answer. HellaSwag~\citep{zellers2019hellaswag} from Open LLM leaderboard also includes wikiHow articles to enhance the content diversity. By cross validating the evaluation set of the HellaSwag task and the training set of \lima, we find that the style and format of 200 wikiHow examples in \lima are highly similar to that of in HellaSwag evaluation set. Also, surprisingly, we notice that multiple examples (e.g., ``How to get a free room upgrade in las vegas?'', ``How to teach a child to use scissors?'', ``How to handle poking wires on braces?'', etc.) appear in both datasets, which is a strong signal of data contamination. The performance of \lima-1k model on the HellaSwag task is also suspiciously higher than the other baselines as shown in Fig.~\ref{fig:openllm_llama2_7b_hellaswag}.

\end{document}